\documentclass[10pt,journal,compsoc]{IEEEtran}
\ifCLASSOPTIONcompsoc
  \usepackage[nocompress]{cite}
\else
  \usepackage{cite}
\fi
\ifCLASSINFOpdf
\else
\fi

\usepackage{cite}
\usepackage{amsmath,amssymb,amsfonts}
\usepackage{algorithmic}
\usepackage{graphicx}
\usepackage{textcomp}
\usepackage{xcolor}
\usepackage{rotating}
\usepackage{multirow}
\usepackage{subfigure}

\usepackage{dsfont}

\usepackage{diagbox}    
\usepackage{url}

\newcommand{\ourmethod}{\textrm{AE--SAD}}    

\newcommand{\nop}[1]{}

\begin{document}
\title{Reconstruction Error-based Anomaly Detection with Few Outlying Examples}

\author{Fabrizio~Angiulli,
        Fabio~Fassetti,
        and~Luca~Ferragina
\IEEEcompsocitemizethanks{\IEEEcompsocthanksitem F. Angiulli, F. Fassetti and L. Ferragina are with DIMES Dept., University of Calabria, 87036 Rende (CS), Italy.\protect\\
E-mail: \{fabrizio.angiulli,fabio.fassetti,luca.ferragina\}@unical.it}\thanks{Manuscript received January D1, 2023; revised Month D2, 2023.}
}

\IEEEtitleabstractindextext{
\begin{abstract}
Reconstruction error-based neural architectures constitute a classical deep learning approach to anomaly detection which has shown great performances. 
It consists in training an Autoencoder to reconstruct a set of examples deemed to represent the normality and then to point out as anomalies those data that show a sufficiently large reconstruction error. Unfortunately, these architectures often become able to well reconstruct also the anomalies in the data.    
This phenomenon is more evident 
when there are anomalies in the training set. In particular when these anomalies are labeled, a setting called semi-supervised, the best way to train Autoencoders is to ignore anomalies and minimize the reconstruction error on normal data.

The goal of this work is to investigate approaches to allow reconstruction error-based architectures to instruct the model to put known anomalies outside of the domain description of the normal data. Specifically, our strategy exploits a limited number of anomalous examples to increase the contrast between the reconstruction error associated with normal examples and those associated with both known and unknown anomalies, thus enhancing anomaly detection performances.

The experiments show that this new procedure achieves better performances than the standard Autoencoder approach and the main deep learning techniques for semi-supervised anomaly detection. 
\end{abstract}

}

\maketitle

\IEEEdisplaynontitleabstractindextext
\IEEEpeerreviewmaketitle
\section{Introduction}
\label{sec:intro}

\IEEEPARstart{A}{nomaly} detection is a fundamental and widely applicable data mining and machine learning task, whose aim is to isolate samples in a dataset that are suspected of being generated by a distribution different from the rest of the data.

Depending on the composition of training and test sets, anomaly detection settings can be classified as unsupervised, semi-supervised, and supervised.
In the supervised setting training data are labeled as normal and abnormal and and the goal is to build a classifier. The difference with standard classification problems there is posed by the fact that abnormal data form a rare class. In the semi-supervised setting, the training set is composed by both labelled and unlabelled data. A special case of this setting is the one-class classification when we have a training set composed only by normal class items. In the unsupervised setting the goal is to detect outliers in an input dataset by assigning a score or anomaly degree to each object.

For each of these scenarios many techniques have been proposed, and recently deep learning based ones are witnessing great attention due to their effectiveness, among them, those employing Autoencoders are largely used.
Roughly speaking, Autoencoders are neural networks that aim at reconstructing the input data after a dimensionality reduction step and anomalies are data with high \textit{reconstruction error}.
In \cite{ChandolaBK09,Aggarwal2013,RuffReview,Pang2021} detailed descriptions of such settings and related techniques are provided.

In this work we present a deep learning based technique for semi-supervised anomaly detection particularly suitable in the context in which a few known anomalies are available.
The idea at its basis is to output a reconstructed version of the input data through an innovative strategy to enlarge the difference between the \textit{reconstruction error} of anomalies and normal items by instructing the model to put anomalies outside of the domain description of the normal data.

Specifically we propose the $\ourmethod$ algorithm (for Semi-supervised Anomaly Detection through Auto-Encoders),
based on a novel training procedure and a new loss function that exploits the information of the labelled anomalies in order to allow Autoencoders to learn to poorly reconstruct them. 
We note that due to the strategy pursued by standard Autoencoders,
that is based on a loss aiming at minimizing the training data reconstruction error such as the mean squared error function, if they are not provided just with normal data during the training
they would learn to correctly reconstruct also the anomalous examples and this may cause a worsening of their generalization capabilities.
Thus, the best strategy to exploit Autoencoders for semi-supervised anomaly detection,
that is when anomalies are available, 
consists simply in disregarding anomalous examples.
In this way the knowledge about the anomalies would not be exploited.
Conversely, by means of the here proposed loss, the Autoencoder is able to take advantage of the information about the known anomalies and use it to increase the contrast between the reconstruction error associated with normal examples and those associated with both known and unknown anomalies, thus enhancing anomaly detection performances.

The experiments show that this new procedure achieves better performances than the standard Autoencoder approach and the main deep learning techniques for both unsupervised and semi-supervised anomaly detection. Moreover it shows better generalization on anomalies generated with a distribution different from the one of the anomalies in the training set and robustness to normal data pollution.

The main contributions of this work are listed next:
\begin{itemize}
    \item We introduce $\ourmethod$, a novel approach to train Autoencoders that applies to the semi-supervised anomaly detection setting and exploits the presence of labeled anomalies in the training set;
    \item We show that $\ourmethod$ is able to obtain excellent performances even with \textit{few anomalous examples} in the training set;
    \item We perform a sensitivity analysis confirming that the selection of the right values for the hyperparameters of $\ourmethod$ is not critic and the method reaches good results in a relatively small number of epochs;
    \item We apply $\ourmethod$ on tabular and image datasets, and on both types of data it outperforms the state-of-the-art competitors;
    \item Finally, we show that our method behaves better than competitors also in two particularly relevant scenarios, that are the one in which \textit{anomalies in the test set belong to classes different from the ones in the training set} and the one in which \textit{the training set is contaminated by some mislabeled anomalies}.
\end{itemize}

The paper is organized as follows. In Section \ref{sec:rel} the main related works are described. In Section \ref{sec:meth} the method $\ourmethod$ is described and a comparative example with standard Autoencoders is provided in order to highlight the motivations behind our approach. In Section \ref{sec:exp} the experimental campaign is described and the results are presented. Finally, Section \ref{sec:concl} concludes the paper.

\section{Related Works}
\label{sec:rel}

Several classical data mining and machine learning approaches have been proposed to detect anomalies, namely, 
statistical-based \cite{DaviesG1993,BarnettL1994},
distance and density-based  \cite{KN00,BKNS00,AP02,AngiulliF09,Angiulli20,JTH01},
reverse nearest neighbor-based \cite{HautamakiKF04,RadovanovicNI15,Angiulli17},
isolation-based \cite{LiuTZ12},
angle-based \cite{KriegelSZ08},
SVM-based \cite{ScholkopfPSSW01,TaxD04},
and many others \cite{ChandolaBK09,Aggarwal2013}.

In last years, the main focus has been on deep learning-based methods \cite{Goodfellow-et-al-2016,chalapathy2019deep,RuffReview,Pang2021}. 
Traditional deep learning methods for anomaly detection belong to the family of \textit{reconstruction error}-based methods employing Autoencoders.
Reconstruction error-based anomaly detection \cite{HawkinsHGR02,recprob} consists in training an Autoencoder to reconstruct a set of examples and then to detect as anomalies those inputs that show a sufficiently large reconstruction error.
This approach is justified by the fact that, since the reconstruction process includes a dimensionality reduction step
(the \textit{encoder}) followed by a step mapping back representations in the
compressed space (also called the \textit{latent space})
to examples in the original space (the \textit{decoder}),
regularities should be better compressed and, hopefully,
better reconstructed \cite{HawkinsHGR02}.

Besides reconstruction error methods, recently some approaches \cite{anogan,fanogan,ganomaly,gaal} based on Generative Adversarial Networks have been introduced. Basically, they consist in a training procedure where a \emph{generator} network learns to produce artificial anomalies as realistic as possible, and a \emph{discriminator} network assigns an anomaly score to each item.

In this work we focus on the family of reconstruction error-based methods. In this context a well known weakness of Autoencoders is that they suffer from the fact that after the training they become able to well reconstruct also anomalies, thus worsening the effectiveness of the reconstruction error as anomaly score. In order to reduce the impact of this phenomenon, methods that isolate anomalies by taking into account the mapping of the data into the latent space together with their reconstruction error have been introduced \cite{vaeout,LatentOut,LatentOutISMIS}.
Specifically, these methods are tailored for the unsupervised and One-Class settings, but they do not straightly apply to the semi-supervised one, in that they cannot exploit labeled examples.

In last years, some approaches \cite{Deep-SAD,a3,devnet,munawar2017limiting,ding2022catching} based on deep learning have been proposed to specifically address the task of \textit{Semi-Supervised Anomaly Detection}.

In \cite{a3} is introduced $A^3$, a neural architecture that combines three different networks that work together to identify anomalies: a \emph{target network} that address the task of feature extractor, an \emph{anomaly network} that generates synthetic anomalies and an \emph{alarm network} that discriminate between normal and anomalous data. In particular the anomaly network has a generative task and is used to balance the training set with artificial anomalous examples.

In \cite{Deep-SVDD} is presented Deep-SVDD a method that tries to mimic, by means of deep learning architectures, the strategy pursued by different traditional one-class classification approaches, like SVDD and OC-SVM, consisting in transforming normal data so to enclose them in a minimum volume hyper-sphere. Differently form traditional reconstruction error based-approaches, these approaches can be trained to leave labeled anomalous outside the hyper-sphere representing the region of normality. 

Although the original method was specifically designed for One-Class classification, in \cite{Deep-SAD} it has been modified in order to address the semi-supervised task, following the paradigm introduced in \cite{gornitz2013toward}. This is done by means of a loss function that minimizes the distance of normal points from a fixed center and maximizes the distance of anomalous ones from it; this approach is called Deep-SAD.

In \cite{devnet} is introduced DevNet which represents one of the seminal Deep Learning approaches specific for Semi-Supervised Anomaly Detection. 
Next, the authors of \cite{a3} and \cite{Deep-SAD} showed that their methods outperform DevNet.


\cite{ding2022catching} considers the 
peculiar scenario 
in which certain anomalous classes (called \emph{gray swans}) are available in the training set, while certain others are not (\emph{black swans}). Authors introduce an approach, called DRA, that employs a multi-head neural architecture that aims at learning a representation of the anomalies by combining information coming from seen anomalies, pseudo-anomalies and latent residual anomalies.

In \cite{munawar2017limiting} the concept of \emph{Negative Learning} is introduced in the context of Anomaly Detection. It is based on the idea of inducing AE to reconstruct anomalous items worst than normal ones. In this paper this paradigm is implemented by dividing the training of the AE in two phases: a \emph{one-class} training in which the standard MSE loss is minimized only on normal items of the training set, and a \emph{negative} training in which only anomalies in the training set are considered and the objective is to maximize the MSE.
In the following of the paper we will refer to this method as Neg-AE.
 
Some unsupervised methods have been designed to include a semi-supervised strategy exploiting self-generated pseudo-anomalies \cite{astrid2021learning,zaheer2022generative}.

In \cite{astrid2021learning} it is introduced a methodology that worsen the reconstruction of anomalies by training an AE to reconstruct normal data in the standard way and generating \emph{pseudo-anomalies} modifying normal data. Once that a pseudo-anomaly is created, the Autoencoder is forced to reconstruct it as similar as possible to the normal item from which it has been generated. The idea is that, after the training phase, the AE will treat real anomalies by reconstructing them as the normal class, thus producing an higher reconstruction error.

In \cite{zaheer2022generative} Generative Cooperative Learning (GCL) is presented. It is a framework specifically built for the unsupervised anomaly detection task, thus it is assumed that labels are not available for the training. GCL is composed by an AE and a binary classifier that are trained in a cooperative fashion by means of an iterative process. In a fixed iteration each module is trained independently and produces pseudo-labels that are passed to the other module.
The two modules exchange information about the potential anomalies with each other improving their capabilities to isolate those anomalies.
In particular, the AE employs the pseudo-anomalies obtained from the classifier by applying a Negative Learning approach to learn to poorly reconstruct them.
A similar approach is adopted in \cite{Deep-UAD}. Here an AE and a fully connected network enclosing normal items into an hypersphere, are trained in a alternate cooperative manner. In this case the sharing of information is implemented by means of the anomaly score of each module that is used as weight for the training of the other one.

The idea of exploiting labelled abnormal data has been also applied to the Abnormal Event Detection in video task \cite{gornitz2013toward,georgescu2021background,acsintoae2022ubnormal}. The architecture introduced in \cite{georgescu2021background} and also considered in \cite{acsintoae2022ubnormal}, for example, is composed by an object detection module that isolates data to train one \emph{appearance} AE and two motion (one forward and one backward) AEs.
The three Autoencoders learn to reconstruct the objects extracted from the input video. At the same time, the AEs are prevented from learning to reconstruct examples from a pool of out-of-distribution training samples passing through the encoders and adversarial decoder
branches. 

\section{Method}
\label{sec:meth}
In this section the technique $\ourmethod$ is presented. It is designed for 
anomaly detection and, in particular, it is focused on \textit{semi-supervised} anomaly detection,
namely on a setting where the training set mainly contains
samples of one \textit{normal} class and very few examples of \textit{anomalous} data.
The aim is to exploit the information coming from this set to train a system able to detect anomalies in a test set.
Specifically, let $X=\left\{\mathbf{x}_1,\dots,\mathbf{x}_n\right\}$ be the training set
and for each $i \in [1\dots n]$, let $y_i\in\{0,1\}$ be the label of $\mathbf{x}_i$, with $y_i=0$ if $\mathbf{x}_i$ belongs to the normal class and $y_i=1$ if $\mathbf{x}_i$ is anomalous; w.l.o.g. we assume that $X \subseteq [0,1]^d$ which is always possible to obtain by normalizing the data.

The proposed technique is based on an \textit{Autoencoder} (AE), a special type of neural network whose aim is to reconstruct in output the data $\mathbf{x}$ received in input and, thus, its loss can be computed as the mean squared error between $\mathbf{x}$ and $\hat{\mathbf{x}}$:
\begin{equation}
    \mathcal{L}(\mathbf{x})=\left|\left|\mathbf{x}-\hat{\mathbf{x}}\right|\right|_2^2.
        \label{eq:base-loss}
\end{equation}
In the unsupervised scenario, \textit{Autoencoder-based anomaly detection} consists in training an AE
to reconstruct a set of data and then to detect as anomalies
those samples showing largest \emph{reconstruction error}, which is defined as the value assumed by Equation \eqref{eq:base-loss} after the training phase and plays the role of anomaly score.
This approach is justified by the observation that, since the reconstruction process includes a dimensionality reduction step (\textit{encoding}) followed by a step mapping back representations in the latent space to examples in the original space (\textit{decoding}), regularities should be better compressed and, hopefully, better reconstructed.

Unfortunately, deep non-linear architectures are able to perform high dimensionality reduction while keeping reconstruction error low and often they generalize so well that they can also
well reconstruct anomalies.

To alleviate this problem and, thus, to improve the performance of the AE exploiting the presence of anomalous examples in the training set, we propose the following novel formulation of the loss:
\begin{equation}
    \label{eq:inverted-loss}
    \mathcal{L}_F(\mathbf{x})=
    \left(1-y\right)\cdot\left|\left|\mathbf{x}-\hat{\mathbf{x}}\right|\right|^2+
    \lambda\cdot y\cdot\left|\left|F\left(\mathbf{x}\right)-\hat{\mathbf{x}}\right|\right|^2,
\end{equation}
where $F:[0,1]^d\rightarrow[0,1]^d$, and $\lambda$ is an hyperparameter that controls the weight of the anomalies, in relation to the normal items, during the training.

When $\mathbf{x}$ is normal the contribution it brings to the loss is $\left|\left|\mathbf{x}-\hat{\mathbf{x}}\right|\right|^2$ which means that the reconstruction $\hat{\mathbf{x}}$ is forced to be similar to $\mathbf{x}$ as in the standard approach. Conversely, if $\mathbf{x}$ is an anomaly, the contribution brought to the loss is $\left|\left|F\left(\mathbf{x}\right)-\hat{\mathbf{x}}\right|\right|^2$ which means that in this case $\hat{\mathbf{x}}$ is forced to be similar to $F(\mathbf{x})$. 
Hence, the idea is that, by exploiting Equation \eqref{eq:inverted-loss}
to evaluate the loss during the training process, normal data $\mathbf{x}$ are likely to be mapped to $\mathbf{\hat{x}}$ which is as similar as possible to $\mathbf{x}$ and anomalous data $\mathbf{x}$ are likely to be mapped to $F(\mathbf{x})$ which is substantially different from $\mathbf{x}$.
Basically, the Autoencoder is trained to reconstruct in output the anomalies in the worst possible way and at the same time to maintain a good reconstruction of the normal items.

Moreover, differently from the standard approach, the proposed technique
does not employ the same function both for training the system and for computing the anomaly score.
Indeed, Equation \eqref{eq:inverted-loss}
is exploited to compute the loss during the training process and Equation \eqref{eq:base-loss} is exploited to compute the anomaly score since it accounts for the reconstruction error and then it is likely to evaluate anomalies as data incorrectly reconstructed being $F(\mathbf{x})$ likely to be substantially different from $\mathbf{x}$.

In this work, unless otherwise stated,  we use as $F(\mathbf{x})$ the function
$F(\mathbf{x}) = \mathbf{1}-\mathbf{x}$, which corresponds, in the domain of the images, to the negative image of $\mathbf{x}$.

\begin{figure*}[t]
\centering
\subfigure[$\ourmethod$.]{
\centering
\begin{minipage}[b]{0.48\textwidth}
\includegraphics[width=1\textwidth]{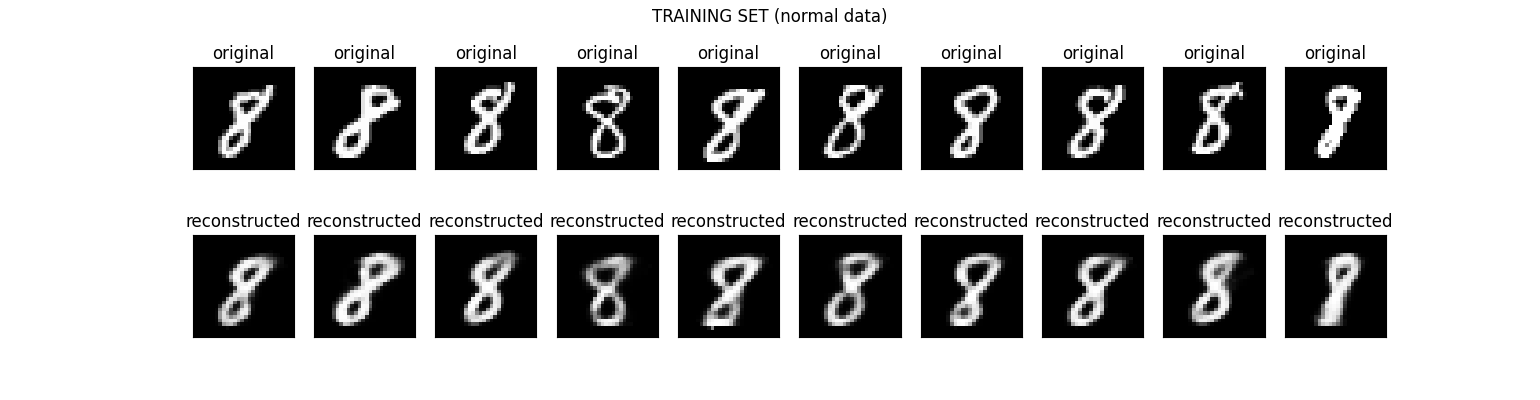} \\
\includegraphics[width=1\textwidth]{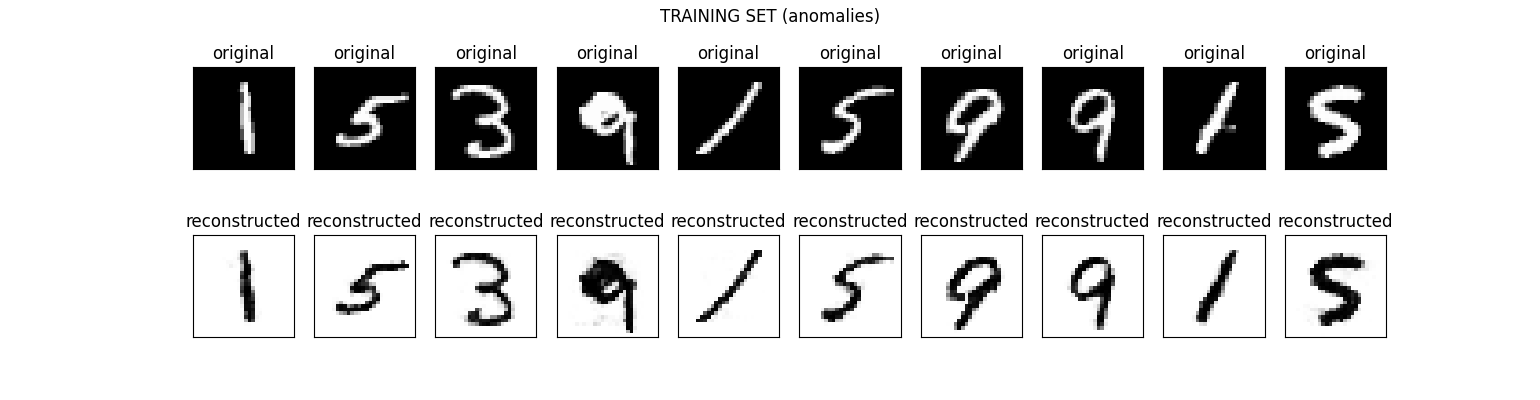} \\
\includegraphics[width=1\textwidth]{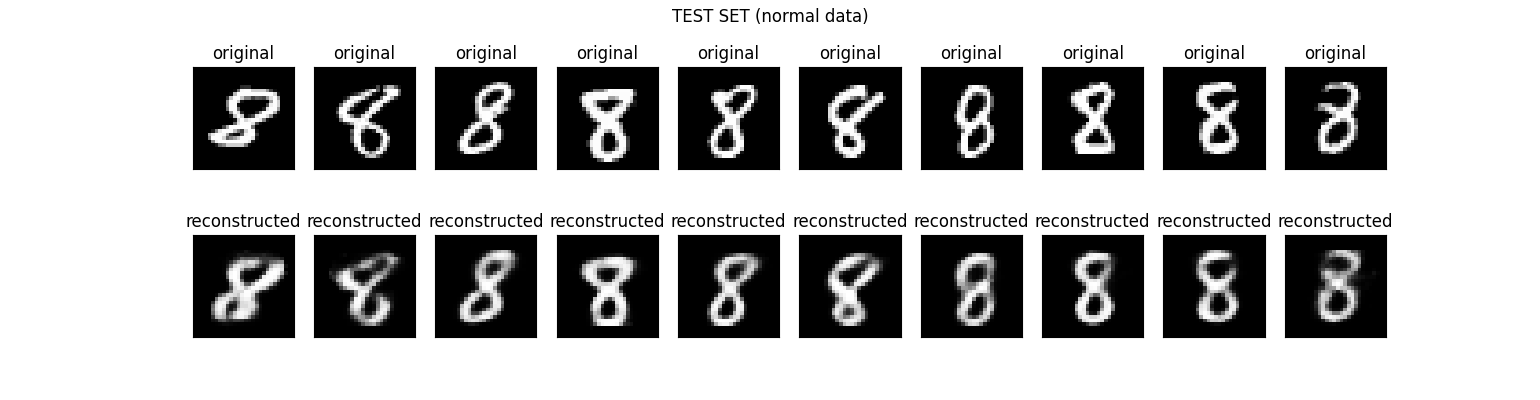} \\
\includegraphics[width=1\textwidth]{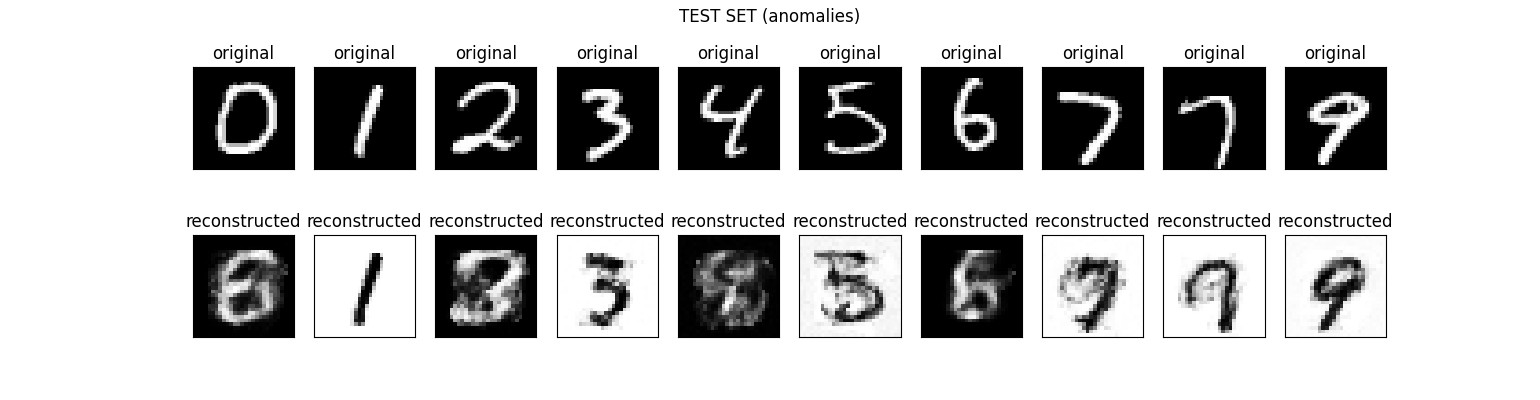} 

\end{minipage}
}
\subfigure[Standard AE.]{
\centering
\begin{minipage}[b]{0.48\textwidth}
\includegraphics[width=1\textwidth]{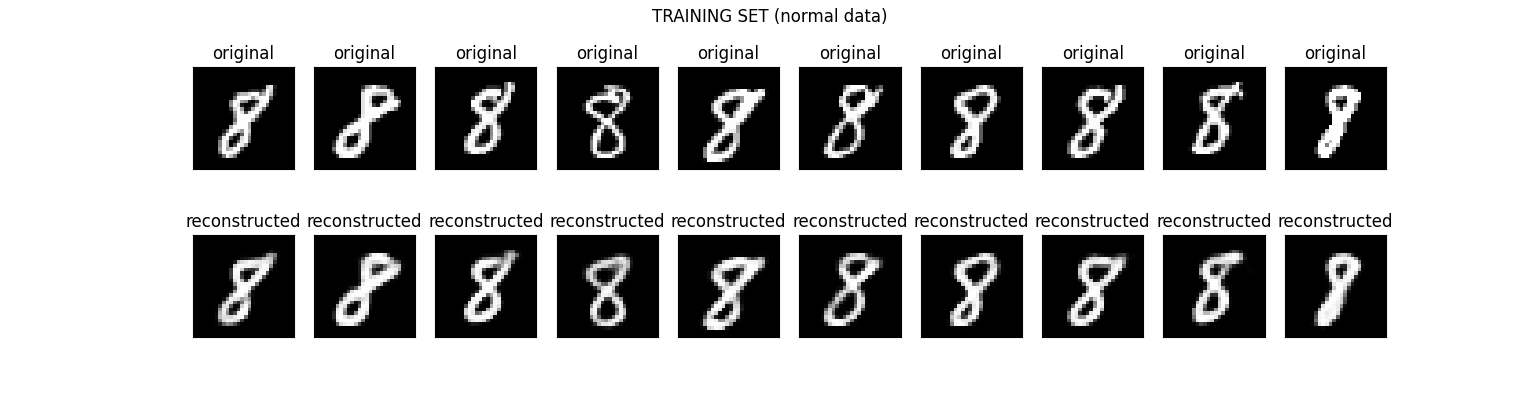} \\
\vspace{1.85cm}\\
\includegraphics[width=1\textwidth]{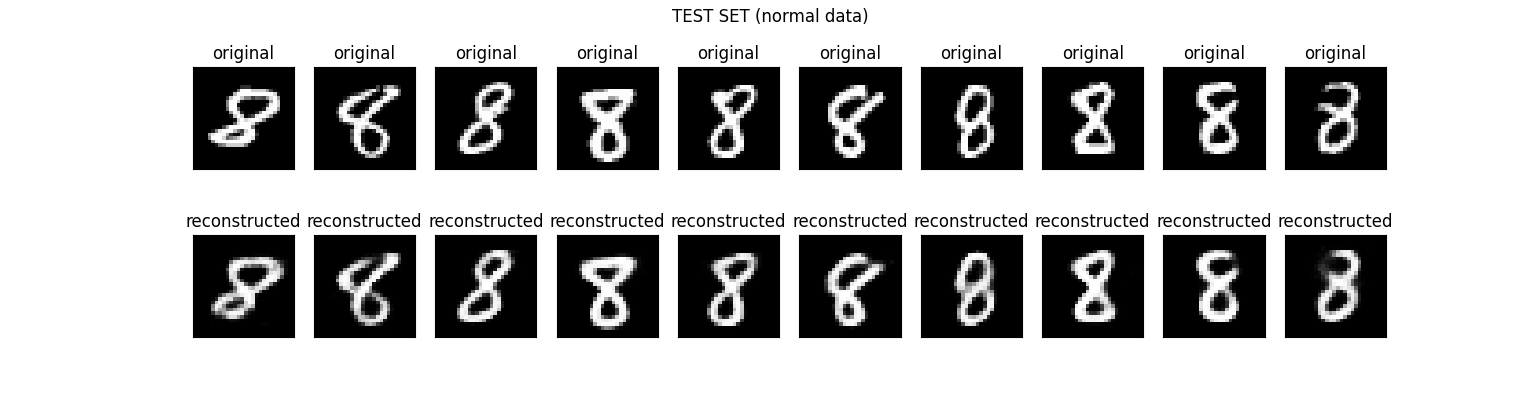} \\
\includegraphics[width=1\textwidth]{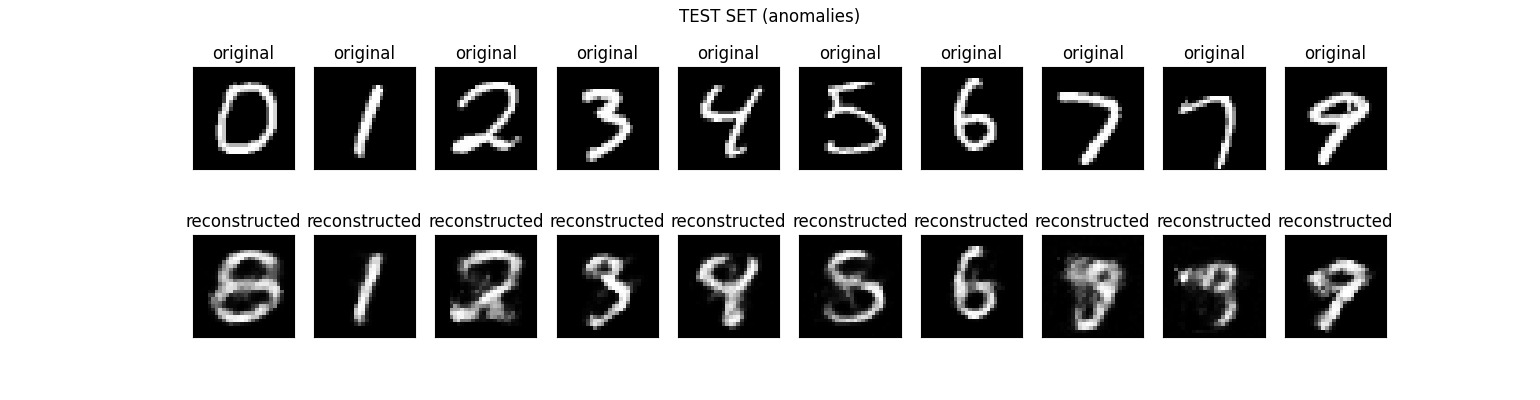} 
\end{minipage}
}
\caption{Original and reconstructed images of training and test set. In this example the class $8$ is normal and the class $1,3,5,9$ are the anomalous classes used in the training.}
\label{fig:example}
\end{figure*}

\subsection{Comparative behavior with standard Autoencoder}

In many real life situations the anomalies that can arise in a evaluation phase may be generated from a distribution different from the anomalies used for the training.

In order to investigate the behaviour of our method in such a situation, we build a set-up in which, for each of the considered datasets, we select for the training phase all the items of a certain class as inliers, and a fixed number of examples from only some of the other classes as anomalies; the test set, on the contrary, is composed by all the classes of the selected dataset. In particular in the test set there will be
\begin{itemize}
    \item examples from the normal class,
    \item anomalies belonging to classes that have been used for the training phase,
    \item anomalies belonging to classes not used for the training, and thus are unknown for the model.
\end{itemize}

In Figure \ref{fig:example}(a) are reported the original and the reconstructed images from the training and the test set obtained in a setting where the class $8$ of the dataset MNIST is considered as normal and the classes $1$, $3$, $5$, and $9$ are the anomalous classes used for the training. There are $100$ labeled anomalies in the training set.
As we expected, in the training set the normal examples are well reconstructed in a similar way as what happens for the standard AE training setting, while the anomalies are reconstructed as the negative of the input images; for what concerns the test set we can see that the normal examples are again well reconstructed, while the anomalies must be distinguished in two different types:
while the classes $1,3,5,9$, that have been used in the training, are reconstructed in a way very similar to the negative image, the remaining classes show a more confused reconstruction. {In particular some unseen classes, such as $7$, are reconstructed in reverse because apparently the Autoencoder judges them more similar to the anomalies used for the training than to the normal class; other classes, such e.g. as $2$ and $4$, present a reconstruction more similar to the original image (the background remains black) but from which it is more difficult to determine the digit that it represents.}

In any case all these reconstructions are very worse than the one obtained in a semi-supervised setting (Figure \ref{fig:example}(b)), and this leads to better performances of the reconstruction error as anomaly score even in this setting where the Autoencoder is trained with only a portion of the anomalies.

\begin{figure*}[t]
    \centering
    \subfigure[]{\includegraphics[width=0.32\textwidth]{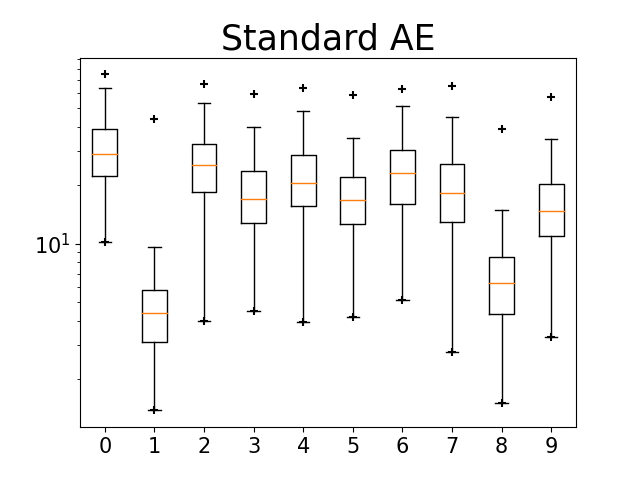}}
    \subfigure[]{\includegraphics[width=0.32\textwidth]{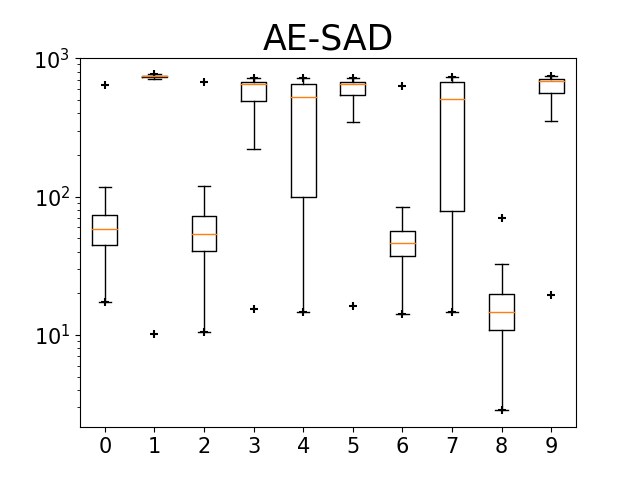}}
    \subfigure[]{\includegraphics[width=0.32\textwidth]{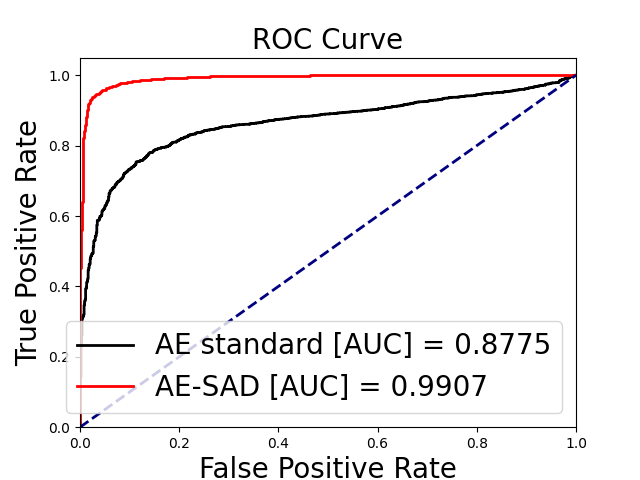}}
    \caption{In this example the class $8$ is normal and the class $1,3,5,9$ are the anomalous classes used in the training.}
    \label{fig:example2}
\end{figure*}

Figure \ref{fig:example2} reports some quantitative results associated with the above experiment. In particular, Figures \ref{fig:example2}(a) and (b) report the boxplot of the test reconstruction error associated with the examples of the classes $0$ to $9$ for the standard AE and for $\ourmethod$. These plots highlight the main effect of our strategy over the classical use of an AE, that is reconstruction errors of anomalous classes are greatly amplified (consider that the y-scale is logarithmic) and the overlapping with the normal class is reduced.
For example, the class $1$, that is essentially indistinguishable from the normal data by the AE, is almost completely separated by our method. Clearly $\ourmethod$ takes advantage of the fact that the class $1$ is an anomalous class seen during training.
However, we can observe that a similar behavior is exhibited also by some classes that are not seen during training as, e.g., class $4$ and $7$.
The reduced overlapping between reconstruction errors
is witnessed by the comparison between AUCs (see \ref{fig:example2}(c)) 
which passes from $0.8775$ in the case of the AE
to $0.9907$ after the adoption of the proposed strategy.

\begin{table}[h!]
    \centering
    \begin{tabular}{cc|ccc}
\multicolumn{2}{c|}{Class} & AE & $\ourmethod$ \\
    \hline
\multirow{4}{*}{\begin{turn}{90}seen\end{turn}} &   1  &    0.2780 & 0.9854 & $+0.7074$ \\
 &   3  &    0.9416 & 0.9960 & $+0.0544$ \\
 &   5  &    0.9365 & 0.9934 & ${+0.0569}$ \\
 &   9  &    0.8863 & 0.9962 & $+0.1099$ \\
\hline
\multirow{5}{*}{\begin{turn}{90}unseen\end{turn}} &   0  &    0.9907 & 0.9924 & ${+0.0017}$ \\
 &   2  &    0.9718 & 0.9774 & ${+0.0056}$ \\
 &   4  &    0.9431 & 0.9902 & ${+0.0471}$ \\
 &   6  &    0.9527 & 0.9698 & $+0.0171$ \\
 &   7  &    0.9275 & 0.9919 & ${+0.0644}$ \\
\hline

    \end{tabular}
    \caption{Comparison between the AUC of AE and $\ourmethod$ on each anomalous class for the example of Figures \ref{fig:example} and \ref{fig:example2}.}
    \label{tab:example}
\end{table}

Table \ref{tab:example}
reports the AUC obtained on a test set consisting in 
unseen examples from the normal and each anomalous class 
of AE and $\ourmethod$, and the increase in AUC
achieved by $\ourmethod$, respectively.
Specifically, the first four rows concern classes seen during 
training, while the remaining five rows classes unseen during training.
As a main result,
we can observe that all the anomalous classes, both seen and unseen, take advantage
of the $\ourmethod$ strategy.
As for the unseen classes, the increase in AUC is always above $+0.05$
and is practically close the maximum achievable, since all the final AUC values
are around $0.99$.
As for the unseen classes, two of them achieve sensible AUC improvements,
namely class $4$ and $7$, 
while for the remaining classes there is a gain, even if less marked.

In conclusion, notably, all the seen anomalous classes, even those which are practically indistinguishable from the normal one by standard AE, become almost completely separated using our method. Moreover, also unseen anomalous may achieve large accuracy improvements either since they are reversely reconstructed or their associated reconstruction error get worse due to a more confused reconstruction.

\section{Experimental results}
\label{sec:exp}
In this section we describe experiments conducted to study the behavior of the proposed method.
In particular, we start by describing the experimental settings (Section \ref{sect:exp_sett}) and, subsequently, we report the experimental evaluation of $\ourmethod$\footnote{Code: \texttt{https://github.com/lucaferragina/AE-SAD}.} driven by three main goals, namely 
\begin{itemize}
    \item studying how parameters affect its behavior,
    \item comparing its performance with existing algorithms,
    \item analyzing its effectiveness in the scenario of polluted data.
\end{itemize}

In detail, Sections are organized as follows.
In Sections \ref{sect:exp_sens}, \ref{sect:exp_epochs} and \ref{sect:exp_f} we describe the sensitivity of the method to the main parameters of the algorithm, namely the number of known anomalies, the regularization term, the number of training epochs and the transformation function $F(\mathbf{x})$. In Section \ref{sect:exp_comp} we compare $\ourmethod$ with existing methods in the context where they are trained on a data set containing normal data and few known anomalies and tested on a data set with normal data and anomalies belonging to both seen and unseen anomalous classes.
In Section \ref{sect:exp_comp_poll} we test $\ourmethod$ in the challenging scenario, often arising in real life applications, where the training set is polluted by mislabeled anomalies. Specifically, the methods are trained with a set containing normal data, data labeled as anomalous, and anomalous data incorrectly labeled as normal.
Furthermore, we report, in Section \ref{sect:exp_a3}, the behavior of the proposed technique when known anomalies belong to a set of classes.

\begin{figure*}[t]
\centering
\includegraphics[width=0.32\textwidth]{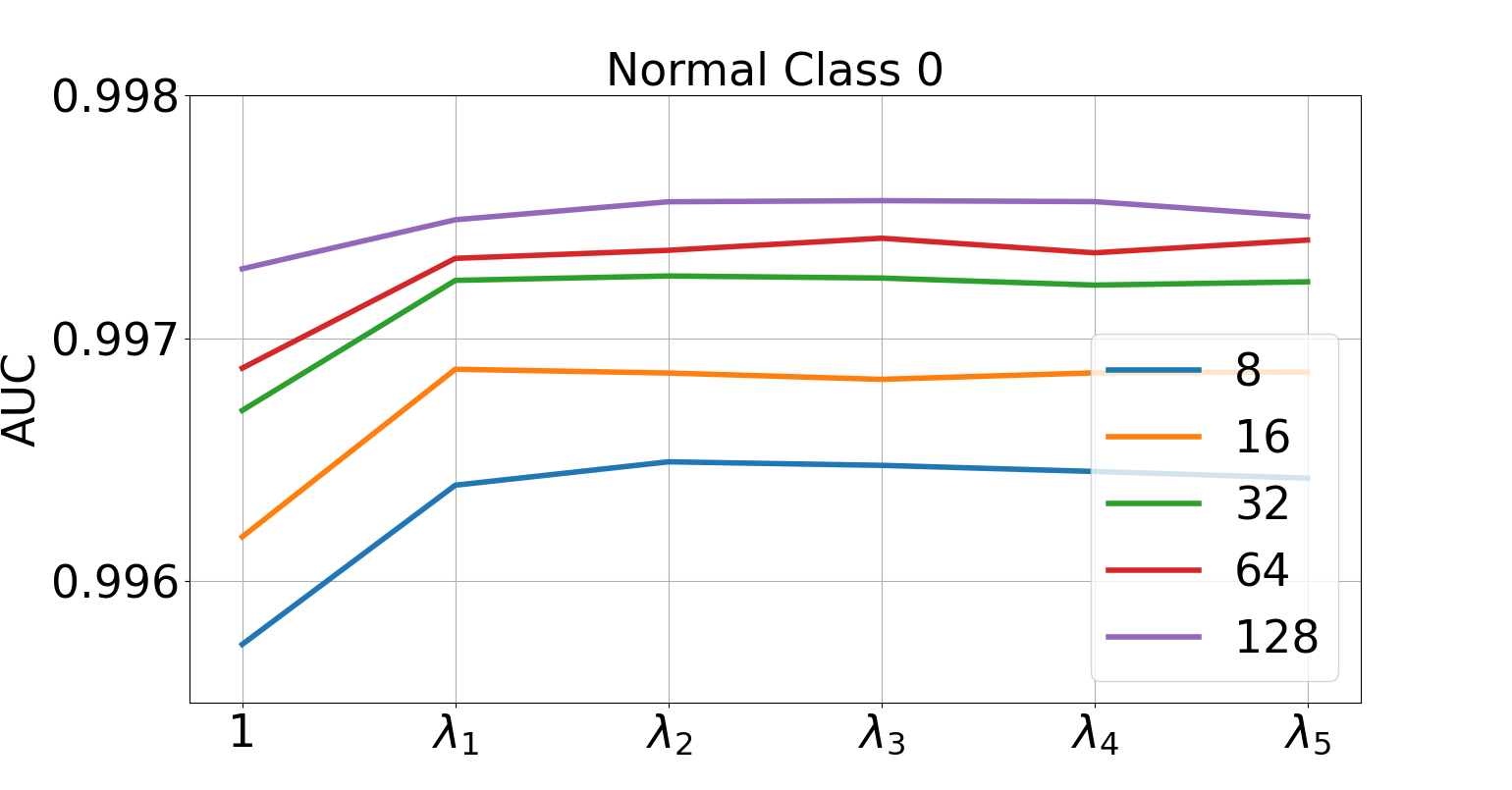} 
\includegraphics[width=0.32\textwidth]{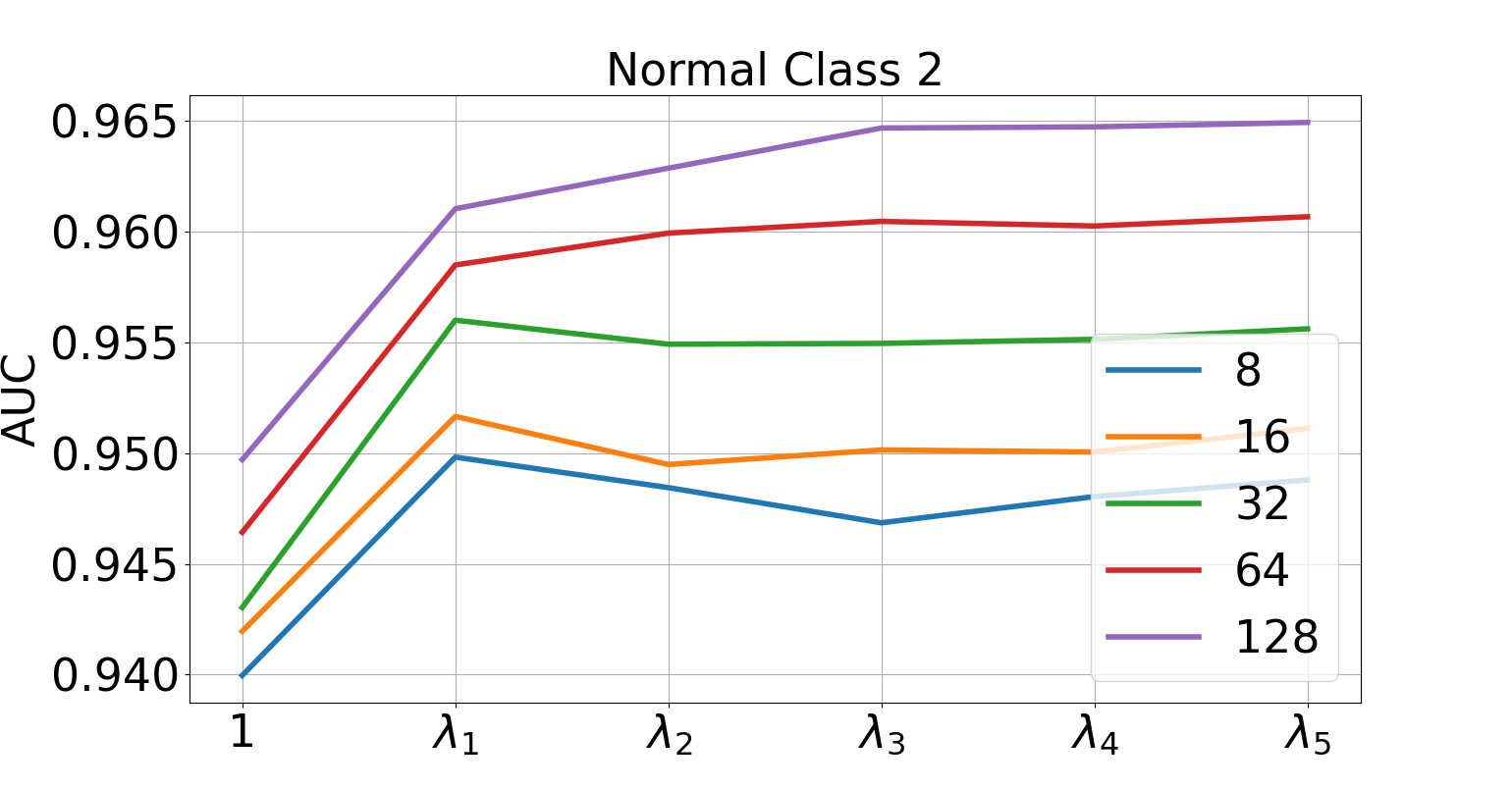} 
\includegraphics[width=0.32\textwidth]{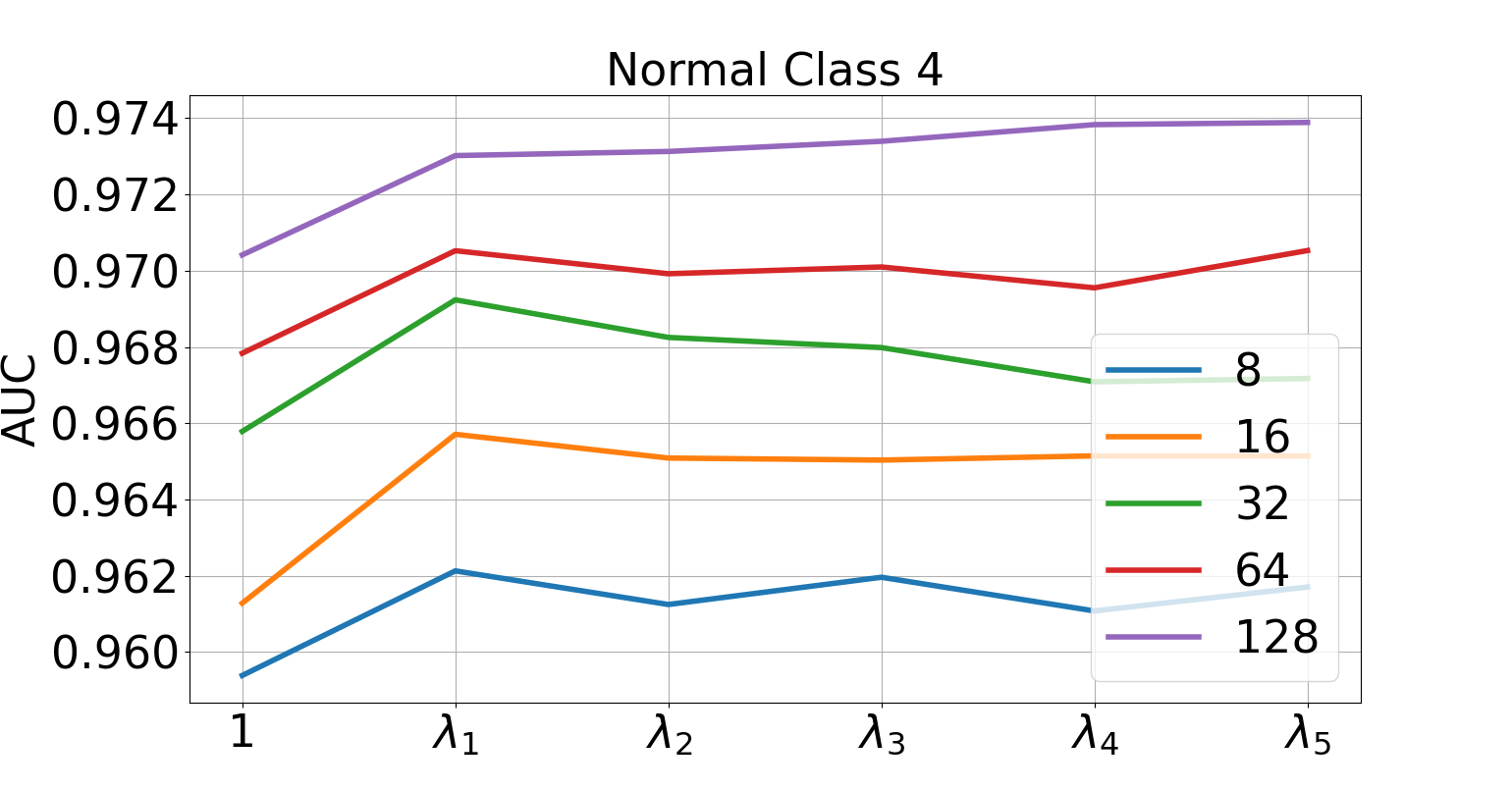} 
\includegraphics[width=0.32\textwidth]{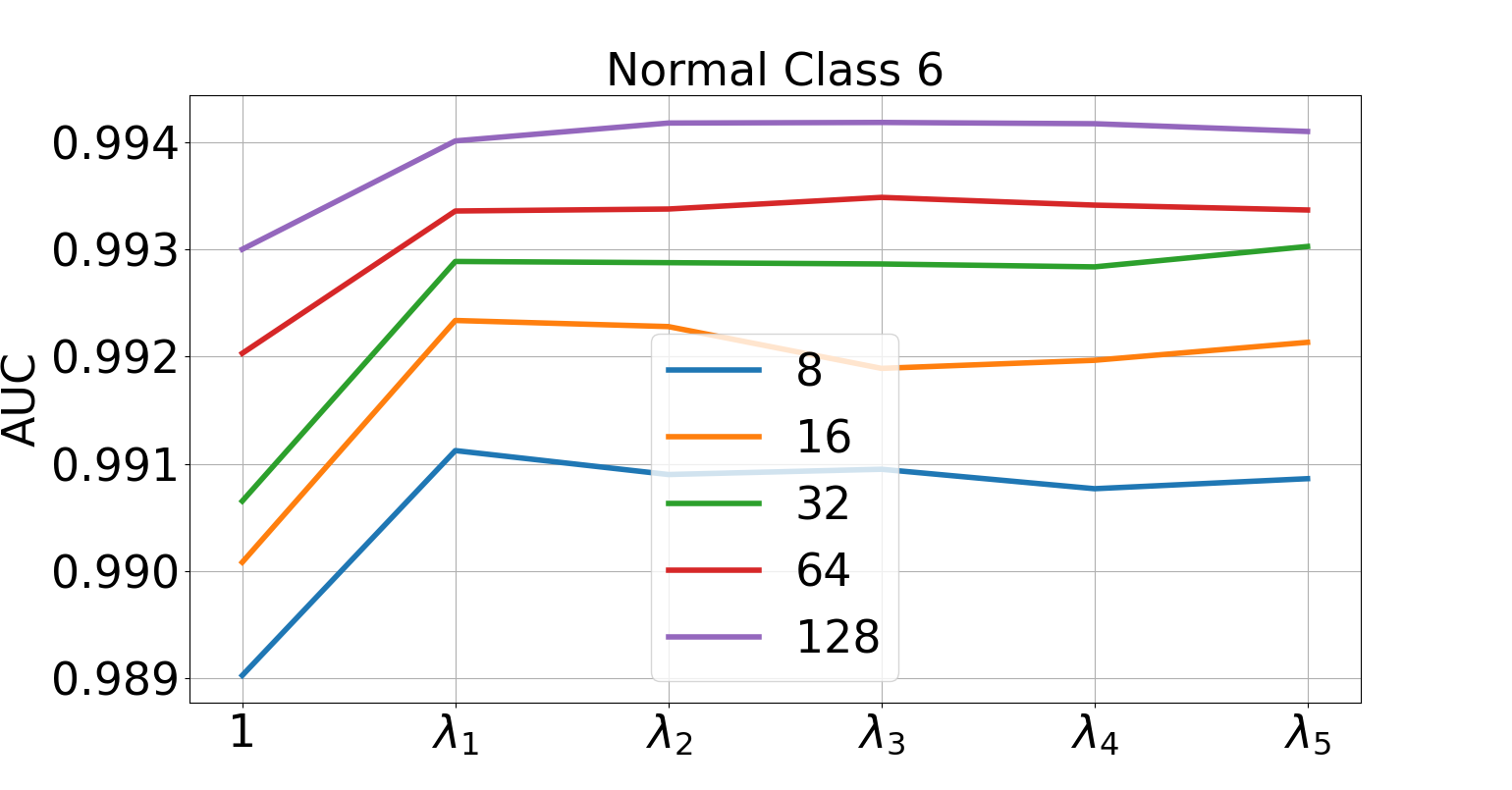} 
\includegraphics[width=0.32\textwidth]{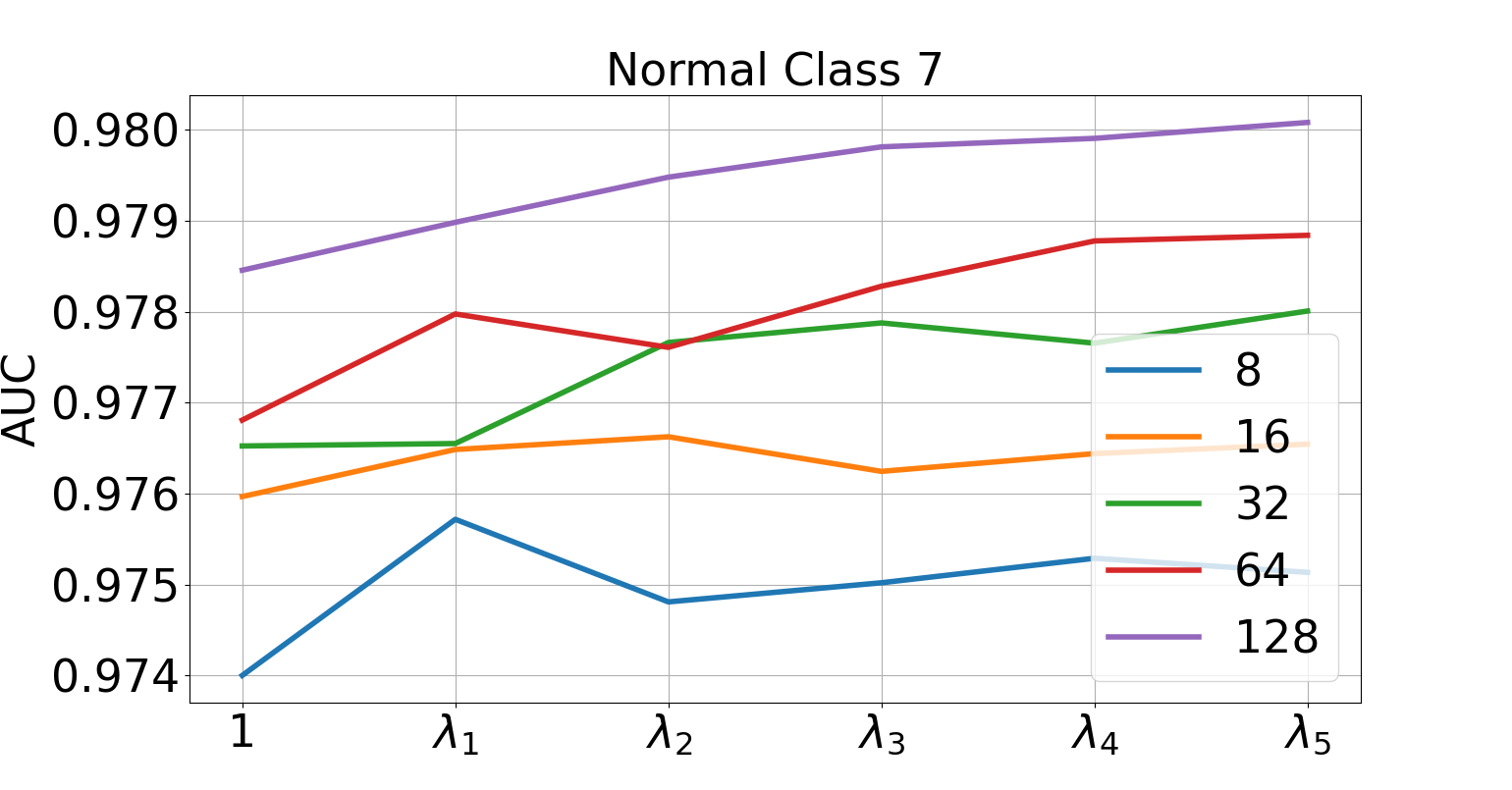} 
\includegraphics[width=0.32\textwidth]{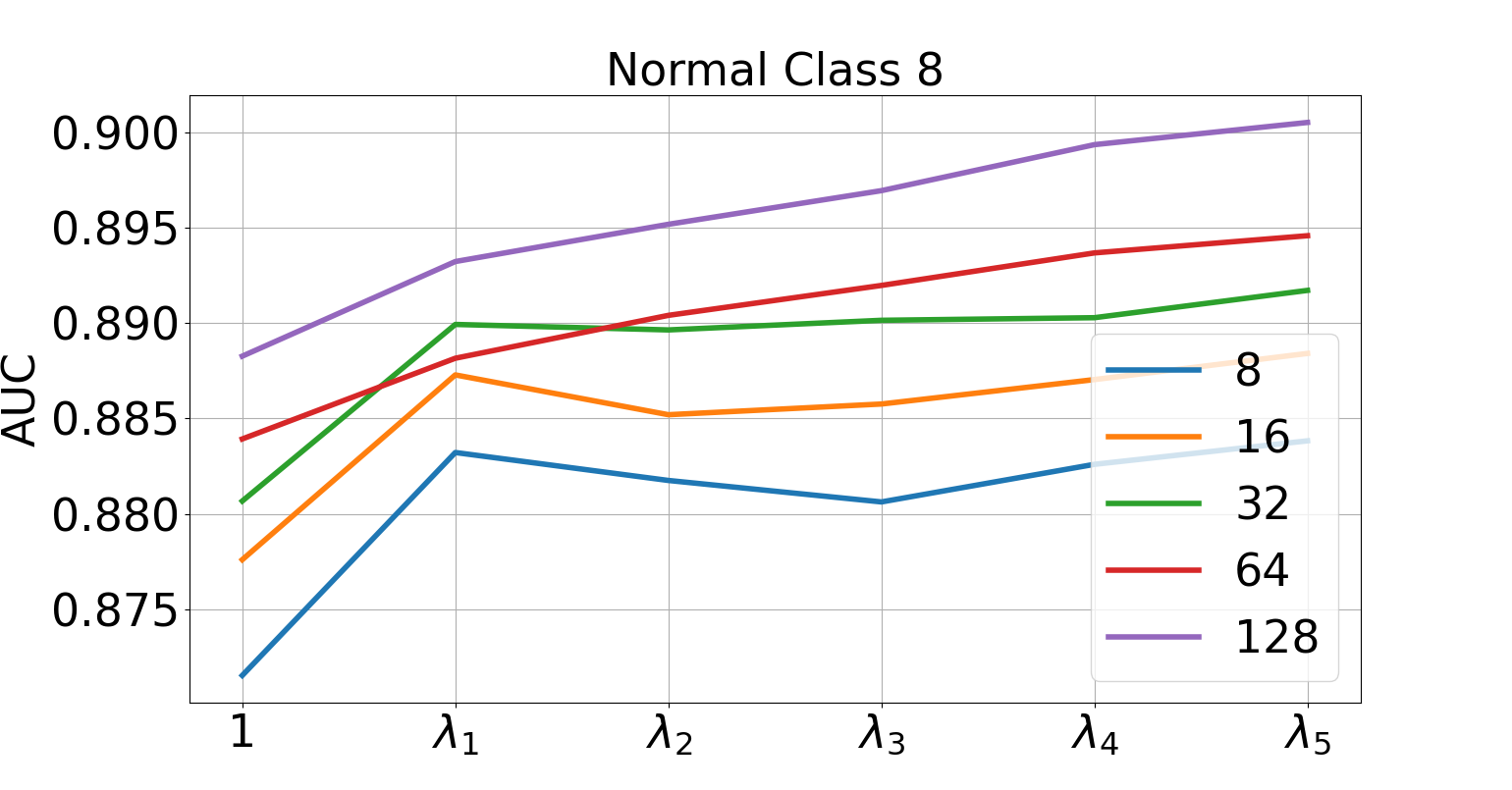} 
\caption{Sensitivity to the regularization parameter $\lambda$.}
\label{fig:auc_LAMBDAS}
\end{figure*}

\begin{figure*}[t]
\centering
\includegraphics[width=0.32\textwidth]{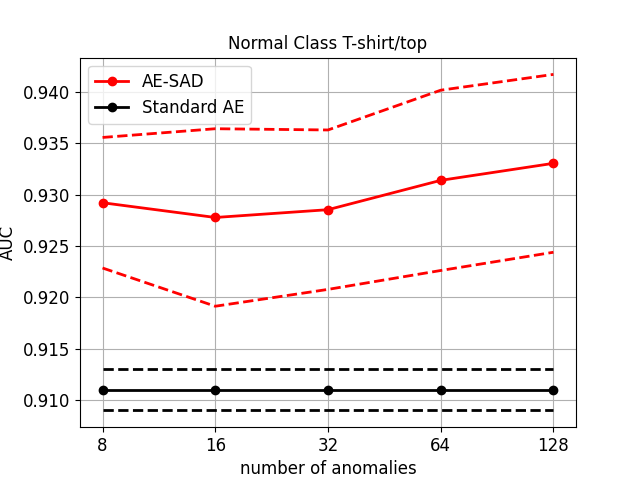} 
\includegraphics[width=0.32\textwidth]{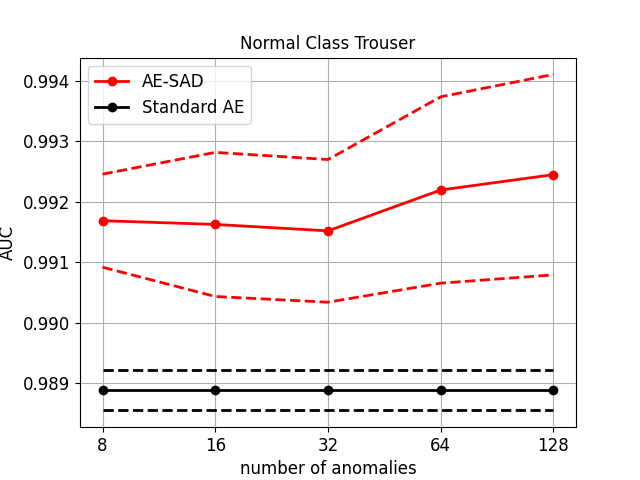} 
\includegraphics[width=0.32\textwidth]{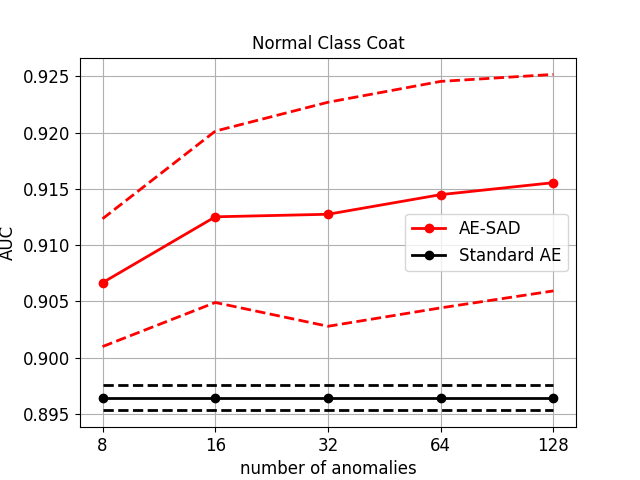} 
\includegraphics[width=0.32\textwidth]{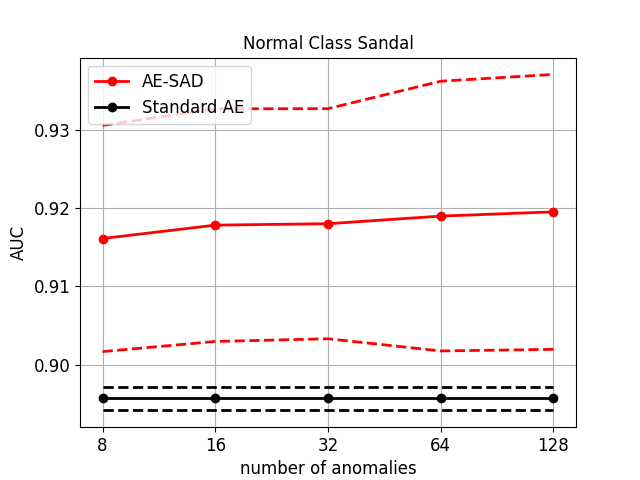} 
\includegraphics[width=0.32\textwidth]{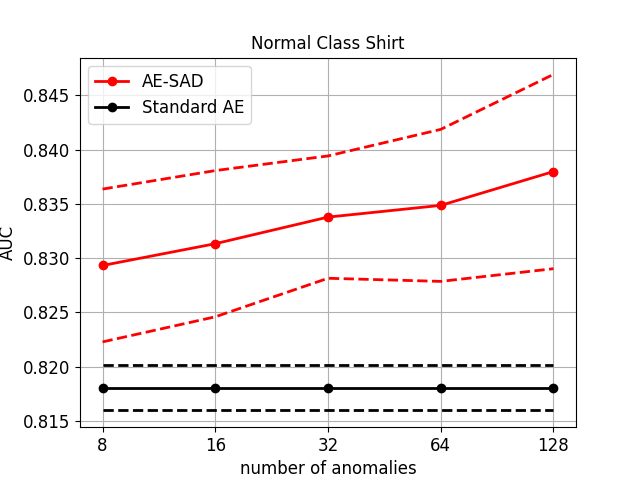} 
\includegraphics[width=0.32\textwidth]{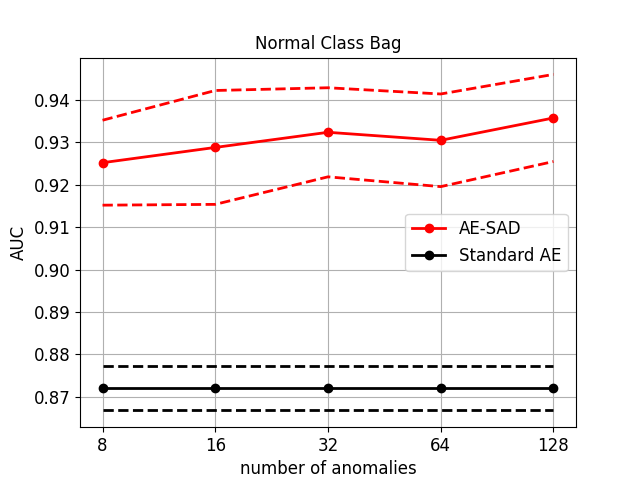}

\caption{Sensitivity to the number of labeled anomalous examples $s$.}
\label{fig:auc_examples}
\end{figure*}

\begin{figure*}[t]
\centering
\includegraphics[width=0.32\textwidth]{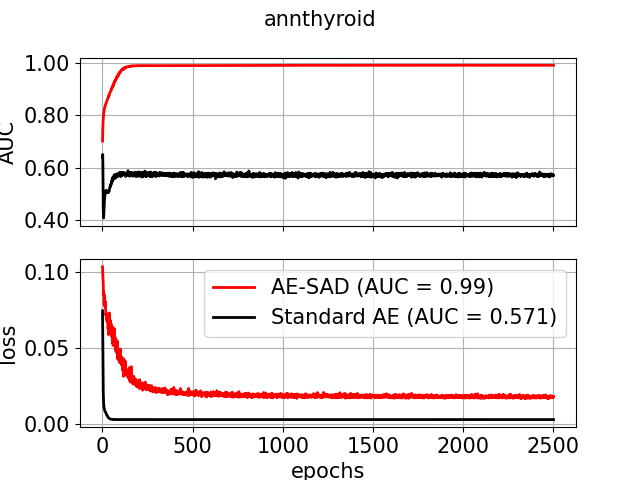} 
\includegraphics[width=0.32\textwidth]{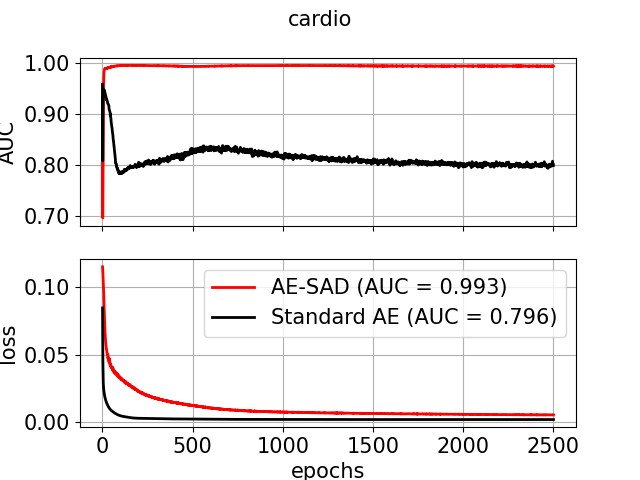} 
\includegraphics[width=0.32\textwidth]{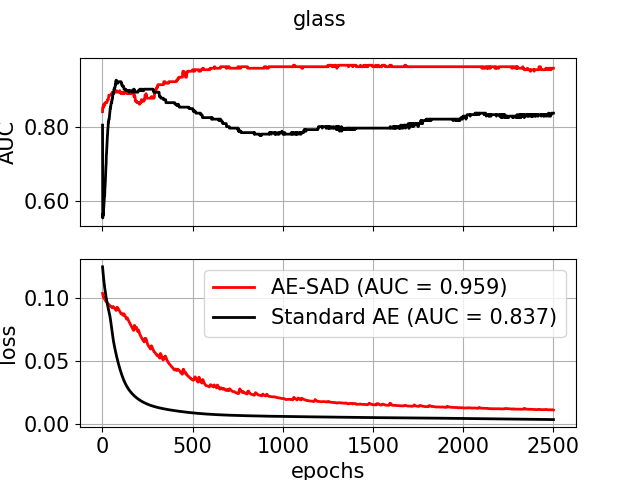} \\
\includegraphics[width=0.32\textwidth]{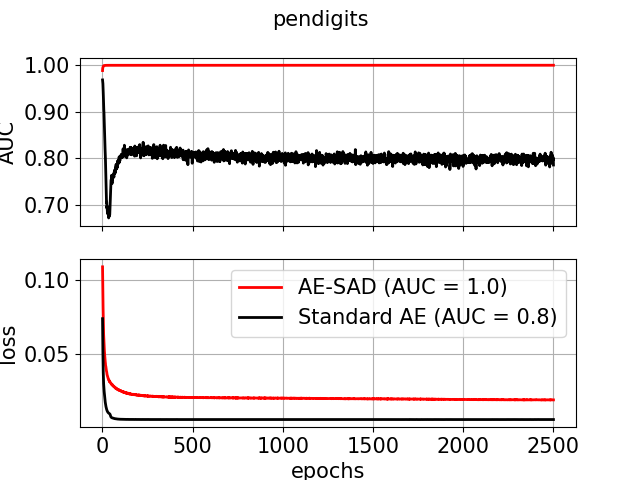}
\includegraphics[width=0.32\textwidth]{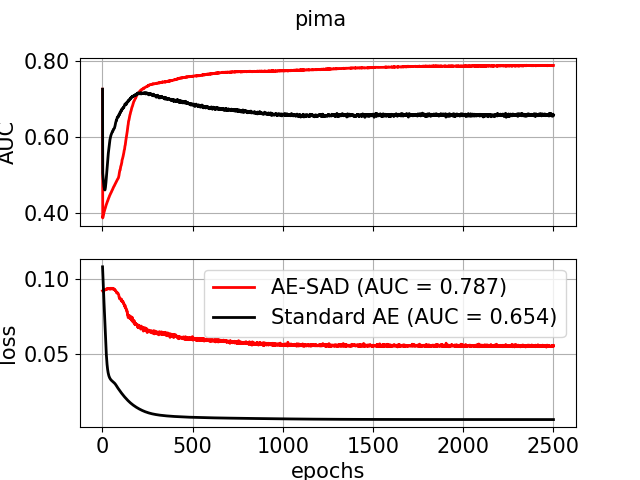}
\includegraphics[width=0.32\textwidth]{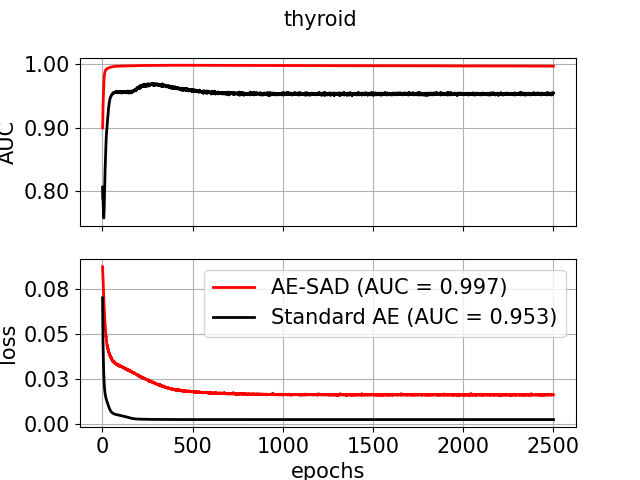}
\caption{Sensitivity to the number of epochs. For both standard Autoencoder and $\ourmethod$ is reported in the legend the AUC value relative to the epoch in which the lowest value of the loss has been observed.}
\label{fig:auc_eps}
\end{figure*}

\subsection{Experimental settings}\label{sect:exp_sett}
In general, given a multi-labeled dataset consisting of a training set and of a test set, we select some of the original classes to form the \textit{normal} class of our problem and some other classes as \textit{known anomalous} classes.
The remaining classes form the \textit{unknown anomalous} classes.
Then, the training set for our method consists of all the training examples of the normal class, labeled as $0$, and of $s$ randomly picked training examples of the anomalies, labeled as $1$. 
Our test set coincides with the original test set and, hence, it comprises all the kinds of anomalies, both known and unknown.
In the rest of the section, we consider different rules for selecting normal and known anomalous classes in order to take into account different scenarios. We will refer to them later as one-vs-one, one-vs-many, one-vs-all, and many-vs-many, depending on the number of original classes used to form both the normal and the known anomalous classes.
Moreover, sometimes we consider also the case in which the \textit{normal data is polluted} by injecting a certain percentage of mislabeled known anomalous examples (these examples are wrongly labeled as $0$).
Because of fluctuations due to the random selection of the $s$ known anomalies, we run the methods on $10$ different versions of the same kind of training sets.
{Results of each experiment are obtained as the mean on the different runs}.

When an unsupervised method is employed to detect the anomalies in the test set, since it does not take advantage of the labels associated with training data, we remove the examples labeled as $1$ from the training set to avoid it incorporates them in the normal behavior. Indeed, in this case the method works as a semi supervised classifier and, hence, it expects to be trained only on normal data.
The considered competitors are the baseline Autoencoder, employing the same architecture of $\ourmethod$ but with the classical reconstruction error loss, and the One-Class SVM (OC-SVM) \cite{ScholkopfPSSW01} as one-class methods, and $A^3$ \cite{a3}, Deep-SAD \cite{Deep-SAD}, Neg-AE \cite{munawar2017limiting}, DRA \cite{ding2022catching} as neural network based methods specific for the semi-supervised setting.

As for the datasets, depending on the experiment,
we consider the tabular real-world anomaly detection specifc datasets belonging to the ODDS library \cite{ODDS}, as well as the MNIST \cite{mnist}, E-MNIST \cite{emnist}, Fashion-MNIST \cite{fmnist}, CIFAR-10 \cite{cifar}, and FRUIT \cite{fruit} image datasets.

As for the architecture of $\ourmethod$, we employ an Autoencoder consisting of two layers for the encoder, a latent space of dimension $32$, and two layers symmetrical to the ones of the encoder for the decoder. We use dense layers on the MNIST and Fashion MNIST datasets and convolutional layers on CIFAR-10. 
{As for the other methods, we employ the parameters suggested in their respective papers.}

\subsection{Sensitivity analysis to the parameters $\lambda$ and $s$}\label{sect:exp_sens}
In this section, we intend to determine the impact of the regularization parameter $\lambda$ on the results of the method and the behavior for different numbers of labeled anomalies.
To this aim, we first consider the MNIST dataset and vary both the number $s$ of labeled anomalies in $\{8,16,32,64,128\}$ and the value of the parameter $\lambda$.

As for the latter parameter, since it weights the contribution to the loss coming from the labeled anomalies, it makes sense to relate it to the number $s$ of anomalies seen during training.
Thus, let $\alpha$ be a value in $[0,1]$. We denote by $\lambda(\alpha)$ the following
value for the regularization parameter
\[ \lambda(\alpha) = 1 + \alpha \left( \frac{n}{s} - 1 \right),  \]
varying between $1$ (for $\alpha=0$) and $n/s$ (for $\alpha=1$).
Since the weight of each is $1$ in our loss, for $\alpha=0$ each single anomalous example weights as a single inlier, while for $\alpha=1$ each single anomalous example weights as $n/s$ inliers and, thus, globally inliers and outliers contribute half of the total loss in terms of weights. For $\alpha>1$ outliers weight globally more than inliers.
In our experiments, we varied $\lambda$ in $\lambda\in\{\lambda_0 = \lambda(0) = 1,\lambda_1=\lambda(\alpha_1),\dots,\lambda_5=\lambda(\alpha_5)\}$, 
where $\alpha_1,\ldots,\alpha_5$ are $5$ log-spaced values between $0.1$ and $2$.

As for the datasets, in this experiment we consider the \textit{one-vs-one} setting.
Consider a dataset containing $m$ classes.
We fix in turn one of the $m$ classes as the normal one and then obtain $m-1$ different training sets by selecting one the remaining classes as the known anomalous.
Since each of the considered datasets consists of $m=10$ classes and since
for each training set we consider $10$ different versions, the total number of training sets is $10m(m-1)=900$.

Figure \ref{fig:auc_LAMBDAS} reports the results of the experiment on MNIST. 
Since we consider $6$ different values for $\lambda$ and $5$ different values for $s$, the total number of runs is in this case $27,\!000$.
Specifically each curve reports the average AUC of $\ourmethod$ when the normal class is held fixed.
All the curves show a similar behavior.
As for the regularization parameter $\lambda$, 
it can be concluded that its selection does not represent a critical choice. For $\lambda>1$ some improvements can be appreciated, but the method appears to be not too much sensitive to the variation of the regularization parameter independently from the number of labeled anomalies.
This behavior can be justified by the fact that, even if anomalies are few in comparison to normal examples and, moreover, even if
anomalies would globally weight less than inliers in the loss function, the reconstruction error of a badly reconstructed anomaly (that is not reversely reconstructed as required by our loss) is potentially much larger than the reconstruction error associated with a single inlier.
Indeed, if the network learned to output the original reconstruction of even a single anomaly, the associated loss contribution would be very large due to the requirement to minimize the distance to the target reverse reconstruction function.
Informally, we can say that {$\ourmethod$ cannot afford to make a mistake in reconstructing anomalies}.
From the above analysis, as for the parameter $\lambda$ in the following, if not otherwise stated, we hold fixed $\lambda$ to the value $\lambda_1 = \lambda(0.1)$.

Figure \ref{fig:auc_examples} shows the sensitivity to the parameter $s$ on Fashion-MNIST. {The AUC of the baseline AE is also reported for comparison. The latter AUC is constant since AE in an unsupervised method and do not uses the labeled anomalies.} These curves confirm the behavior already observed on MNIST.
Summarizing, as for the number of labeled anomalies $s$, $\ourmethod$ performs remarkably well even when only a few anomalies are available. The same reason provided above can be used to justify this additional result.
In the following, if not otherwise stated, we set $s=8$ by default to considered the challenging scenario in which a limited number of labeled anomalies is available.

\subsection{Sensitivity analysis to the number of epochs}\label{sect:exp_epochs}

The number training epochs is an important hyperparameter for all methods that employ neural networks. 
In the anomaly detection context the choice of this hyperparameter might become critical since the value of the loss function of a standard AE is not indicative of the quality as anomaly detector. 
Indeed, while a low MSE value on the training set means that training examples are well reconstructed, this does not necessarily implies that anomalies are better identified.
Differently, with our method we introduce a novel loss and a novel strategy to train an Autoencoder that aims at enlarging the difference between the reconstruction error of the anomalies and the ones of the normal items. 

Thus, in order to compare the behaviour of $\ourmethod$ with standard AEs as the number of epochs grows, we consider the ODDS datasets and we train both the models for $2500$ epochs and at each epoch we compute the AUC on the test set and the value of the loss function on the training set. 
Since the ODDS datasets are not partitioned into a training and a test set by design, we split each of them by randomly selecting the $60\%$ of the normal items to be part of the training set and keeping the rest for the test set.
As for the anomalies, we randomly pick a number of them such that they represent the $5\%$ of the training set, while the remaining are used in the test set.
For $\ourmethod$, we fix $\lambda$ by following the rule obtained in previous section.

The results reported in Figure \ref{fig:auc_eps} show that in general $\ourmethod$ outperforms the standard AE in terms of AUC at each epoch. Moreover the trend of the AUC of our method is much more regular and almost always increasing. This can be justified because after a certain number of epochs the standard AE learns to reconstruct also the anomalies, while by using the loss \eqref{eq:inverted-loss} we always increase the contrast between the reconstruction of the two classes during all the training process.
It can be concluded that, while the number of epochs is a critical hyperparameter for standard AE, this is not the case for $\ourmethod$. In our case it is not hard to select a good value for this hyperparameter, since the more epochs we consider, the higher the expected performances. 

Legends of Figure \ref{fig:auc_eps} report between brackets the AUC value in the epoch in which the lowest value of the loss is reached. As we can see, while for $\ourmethod$ a low value of the loss always implies an AUC value close to the maximum, the same fact does not hold for standard AE. Indeed, for what concerns the latter method, often the maximum AUC value is obtained when the value of the loss is still high. This happens because, the standard AE is able to generalize to classes not included in the training set that could share similarities with anomalies, while the same effect is mitigated in $\ourmethod$ by the nature of the loss.

Finally, from Figure \ref{fig:auc_eps} we can deduce that in general $\ourmethod$ achieves good AUC values even after a relatively small number of epochs, thus the training is not too slow.

\newcommand{\rot}[1]{\begin{turn}{75}#1\end{turn}}
\newcommand{\rta}[2]{\begin{turn}{#1}#2\end{turn}}

\subsection{Sensitivity analysis to the function $F\left(\mathbf{x}\right)$}
\label{sect:exp_f}

In this section we investigate the behavior of $\ourmethod$ with different definitions of the function $F(\mathbf{x})$.

\begin{figure*}[h!]
    \centering
    \includegraphics[width=0.32\textwidth]{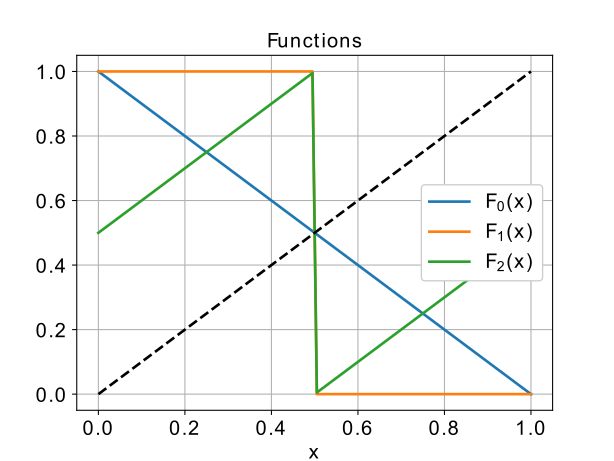}
    \includegraphics[width=0.32\textwidth]{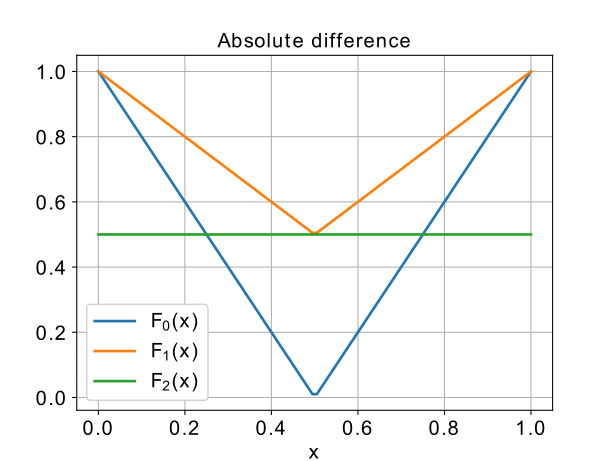}
    \includegraphics[width=0.32\textwidth]{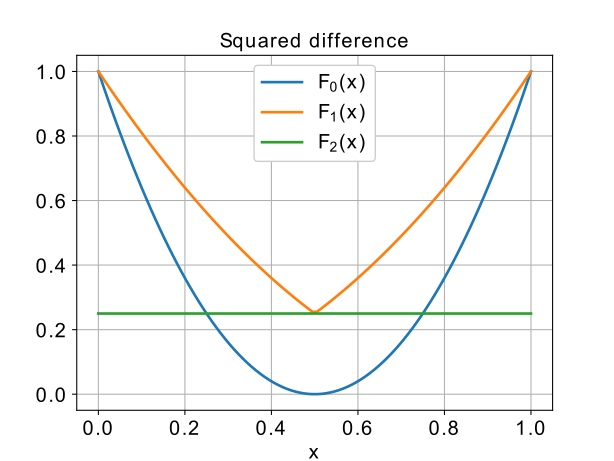}
    \caption{Plots of the functions $F_i(\mathbf{x})$, $\left|F_i(\mathbf{x})-\mathbf{x}\right|$ and $\left(F_i(\mathbf{x})-\mathbf{x}\right)^2$ for $i=0,1,2$.}
    \label{fig:F}
\end{figure*}

In particular, we consider the following functions:
    \begin{itemize}
        \item $F_0(\mathbf{x}) = \mathbf{1}-\mathbf{x}$ which corresponds, in the domain of the images, to the negative image of $\mathbf{x}$.
        \item $F_1(\mathbf{x}) = \mathds{1}_{[0,\frac12]}(\mathbf{x})$ that is the function that maximizes the difference between the input and the output.
        \item $F_2(\mathbf{x}) = (\mathbf{x}+\frac12\mathbf{1})\mathds{1}_{[0,\frac12]}(\mathbf{x})+(\mathbf{x}-\frac12\mathbf{1})\mathds{1}_{(\frac12,1]}(\mathbf{x})$ that has the property that the absolute difference between a point and its output is constantly equal to $\frac12$.
    \end{itemize}

Figure \ref{fig:F} reports the plots of the three functions $F_0$, $F_1$ and $F_2$ (on the left), as well as their absolute (on the center) and squared (on the right) differences with the input. 
These functions are relevant because of the following reasons:
\begin{itemize}
    \item $F_0$ is continuous and its squared difference is differentiable, while it is not the case for $F_1$ and $F_2$;
    \item as for $F_1$, both the absolute and squared differences between the original and the transformed value are maximized compared with the other two definitions;
    \item as for $F_2$, the absolute and squared differences between the original and the transformed value are constantly equal to $\frac12$ and $\frac14$, respectively, thus all the attribute values of an anomalous example provide the same contribution to the anomaly score.
\end{itemize}

\begin{table*}[h!]
        \centering        
        \begin{tabular}{|c|ccc|ccc|ccc|}
        \hline
        \multicolumn{1}{|c|}{Dataset} & \multicolumn{3}{c|}{MNIST} & \multicolumn{3}{c|}{Fashion MNIST} & \multicolumn{3}{c|}{CIFAR-10}\\
        \hline
Class & $F_0(\mathbf{x})$ & $F_1(\mathbf{x})$ & $F_2(\mathbf{x})$ & $F_0(\mathbf{x})$ & $F_1(\mathbf{x})$ & $F_2(\mathbf{x})$ & $F_0(\mathbf{x})$ & $F_1(\mathbf{x})$ & $F_2(\mathbf{x})$\\
\hline 
0 & \bf .998$\pm$.00 & \bf .998$\pm$.00 & .995$\pm$.00 & \bf .952$\pm$.00 & .951$\pm$.01 & .943$\pm$.01 & \bf .700$\pm$.03 & .655$\pm$.08 & .672$\pm$.02 \\
1 & \bf .999$\pm$.00 & \bf .999$\pm$.00 & \bf .999$\pm$.00 & \bf .994$\pm$.00 & .993$\pm$.00 & .991$\pm$.00 & .490$\pm$.05 & .463$\pm$.06 & \bf .490$\pm$.04 \\
2 & .979$\pm$.01 & \bf .983$\pm$.01 & .971$\pm$.01 & .921$\pm$.01 & \bf .923$\pm$.01 & .916$\pm$.01 & .646$\pm$.03 & .647$\pm$.03 & \bf .668$\pm$.01 \\
3 & \bf .982$\pm$.00 & .975$\pm$.01 & .974$\pm$.00 & .954$\pm$.00 & \bf .955$\pm$.00 & .952$\pm$.00 & \bf .534$\pm$.05 & .526$\pm$.04 & .516$\pm$.02 \\
4 & .986$\pm$.00 & \bf .987$\pm$.00 & .980$\pm$.00 & \bf .923$\pm$.00 & .921$\pm$.00 & .922$\pm$.00 & .725$\pm$.02 & .706$\pm$.01 & \bf .729$\pm$.02 \\
5 & \bf .988$\pm$.00 & .987$\pm$.00 & .980$\pm$.00 & \bf .930$\pm$.02 & .924$\pm$.02 & .911$\pm$.01 & .558$\pm$.03 & \bf .589$\pm$.02 & .567$\pm$.04 \\
6 & \bf .998$\pm$.00 & \bf .998$\pm$.00 & .992$\pm$.00 & \bf .854$\pm$.01 & .853$\pm$.01 & .845$\pm$.00 & \bf .752$\pm$.01 & .669$\pm$.07 & .741$\pm$.02 \\
7 & .986$\pm$.00 & \bf .988$\pm$.00 & .983$\pm$.00 & \bf .988$\pm$.00 & .987$\pm$.00 & .986$\pm$.00 & .515$\pm$.03 & \bf .539$\pm$.04 & .524$\pm$.02 \\
8 & .955$\pm$.01 & \bf .965$\pm$.01 & .928$\pm$.01 & \bf .969$\pm$.00 & .968$\pm$.01 & .958$\pm$.01 & \bf .753$\pm$.03 & .687$\pm$.06 & .741$\pm$.04 \\
9 & \bf .982$\pm$.00 & .982$\pm$.00 & .976$\pm$.00 & \bf .983$\pm$.00 & .983$\pm$.00 & .981$\pm$.00 & .482$\pm$.05 & .468$\pm$.09 & \bf .526$\pm$.06 \\
\hline
        \end{tabular}
        \caption{Comparison between three different versions of $\ourmethod$ employing $F_0$, $F_1$ and $F_2$ on MNIST, Fashion-MNIST and CIFAR-10.}
        \label{tab:F}
    \end{table*}

Table \ref{tab:F} reports the results of the comparison between these three functions on MNIST, Fashion MNIST and CIFAR-10.
It can be observed that the function $F_0$,
used as the default $F$ function in Equation \eqref{eq:inverted-loss}, scores the highest number of wins in the considered tests, $F_1$ is the runner up, while $F_2$ seems to present lower performances.
From these results we can draw the conclusion that the property of $F_2$ to vary each value of the same entity has a lower impact on the accuracy of the technique than the property of maximizing the amount of the difference.
As for $F_1$, it shows intermediate performances since, on one hand, it is able to make larger variations than $F_0$ and, in particular, it does not cause variations below $\frac12$; on the other hand, both $F_1$ and $F_2$ present a discontinuity making them harder to be learned than $F_1$, so performances are degraded.

We further note that functions $F_1$ and $F_2$ share the property of mapping any value to a value located at a distance not less than $\frac12$ from the original one. 
Particular settings, such as those in which examples are affected by fog, can be usually satisfactorily managed by using the default function $F_0$ with a normalization pre-processing step. However, if this does not help, $F_1$ and $F_2$ can be suitably used to deal with data distributed around the value
$\frac12$.
Indeed, we simulated a scenario of low contrasting images by uniformly scaling the dataset attributes in the range $[0.4,0.6]$ and considered the $0$ as normal class, obtaining the following results: $AUC(F_0)=0.9766\pm0.0019$, $AUC(F_1)=0.9841\pm0.0054$, and $AUC(F_2)=0.9877\pm0.0022$.

\subsection{Comparison with other methods}\label{sect:exp_comp}
In this section, we compare $\ourmethod$ with the baseline AE, OC-SVM, $A^3$, Deep-SAD, Neg-AE, and DRA. First, we test them on the ODDS datasets splitting them into training and test set in the same way as previous section.
Since the selection of both anomalies and normal examples for training and test set is done by random sampling, we tested all the considered methods on 5 different runs for each dataset and we compute the mean and the standard deviation of the AUC.
The results, reported in the Table \ref{tab:odds} show that our method almost always achieves better performances than competitors, and sometimes this improvements are relevant.
Note that DRA does not appear in Table \ref{tab:odds} since it is specific for images and cannot be applied to tabular data.

\begin{table}[!h]
\setlength{\tabcolsep}{1pt}
    \centering
    \begin{tabular}{|c|cccccc|}
    \multicolumn{7}{c}{ODDS Datasets} \\
\hline Dataset ($d$) & $\ourmethod$ & AE & OC-SVM & $A^3$ & Deep-SAD & Neg-AE\\
\hline
annt. ($6$)    & {\bf.990$\pm$.00} & .692$\pm$.10 & .585$\pm$.00 & .689$\pm$.08 & .662$\pm$.04 & .886$\pm$.06 \\
arrh. ($274$)  & .796$\pm$.05 & .789$\pm$.01 & {\bf.808$\pm$.02} & .692$\pm$.04 & .751$\pm$.04 & .497$\pm$.03 \\
breastw ($9$)       & .992$\pm$.00 & {\bf.995$\pm$.00} & .994$\pm$.00 & {\bf.995$\pm$.00} & .977$\pm$.01 & .989$\pm$.01 \\
cardio ($21$)       & {\bf.994$\pm$.00} & .857$\pm$.08 & .954$\pm$.00 & .904$\pm$.02 & .785$\pm$.07 & .969$\pm$.01 \\
glass ($9$)         & {\bf.974$\pm$.01} & .664$\pm$.11 & .527$\pm$.18 & .639$\pm$.10 & .880$\pm$.08 & .908$\pm$.04 \\
iono. ($33$)   & {\bf.971$\pm$.01} & .910$\pm$.01 & .918$\pm$.01 & .668$\pm$.01 & .953$\pm$.01 & .537$\pm$.07 \\
letter ($32$)       & {\bf.908$\pm$.03} & .802$\pm$.02 & .546$\pm$.02 & .539$\pm$.02 & .768$\pm$.02 & .444$\pm$.03 \\
lympho ($18$)       & .986$\pm$.01 & {\bf.993$\pm$.01} & .940$\pm$.04 & .977$\pm$.04 & .860$\pm$.16 & .965$\pm$.04 \\
mamm. ($6$)   & {\bf.916$\pm$.03} & .874$\pm$.02 & .880$\pm$.02 & .835$\pm$.05 & .759$\pm$.11 & .879$\pm$.07 \\
musk ($166$)        & {\bf1.00$\pm$.00} & {\bf1.00$\pm$.00} & {\bf1.00$\pm$.00} & {\bf1.00$\pm$.00} & {\bf1.00$\pm$.00} & {\bf1.00$\pm$.00} \\
optd. ($64$)    & {\bf1.00$\pm$.00} & .975$\pm$.01 & .718$\pm$.03 & .906$\pm$.03 & .843$\pm$.05 & .997$\pm$.00 \\
pend. ($16$)    & {\bf1.00$\pm$.00} & .747$\pm$.08 & .983$\pm$.01 & .983$\pm$.01 & .964$\pm$.02 & .999$\pm$.00 \\
pima ($8$)          & {\bf.733$\pm$.02} & .594$\pm$.03 & .698$\pm$.01 & .634$\pm$.05 & .657$\pm$.02 & .576$\pm$.10 \\
satellite ($36$)    & {\bf.907$\pm$.02} & .812$\pm$.01 & .728$\pm$.00 & .802$\pm$.00 & .827$\pm$.03 & .773$\pm$.02 \\
sat. ($36$)   & {\bf.999$\pm$.00} & .998$\pm$.00 & .997$\pm$.00 & .982$\pm$.00 & .968$\pm$.01 & .998$\pm$.00 \\
shuttle ($9$)       & .808$\pm$.35 & .914$\pm$.14 & {\bf.993$\pm$.00} & .977$\pm$.01 & .984$\pm$.01 & .979$\pm$.01 \\
speech ($400$)      & .479$\pm$.02 & .463$\pm$.02 & .465$\pm$.02 & .535$\pm$.05 & .478$\pm$.03 & {\bf.622$\pm$.03} \\
thyroid ($6$)       & {\bf.987$\pm$.01} & .922$\pm$.07 & .897$\pm$.04 & .776$\pm$.12 & .885$\pm$.09 & .963$\pm$.06 \\
vertebral ($6$)     & .635$\pm$.14 & .444$\pm$.08 & .522$\pm$.06 & .351$\pm$.03 & .558$\pm$.04 & {\bf.746$\pm$.03} \\
vowels ($12$)       & {\bf.975$\pm$.02} & .904$\pm$.03 & .816$\pm$.05 & .528$\pm$.06 & .937$\pm$.05 & .832$\pm$.11 \\
wbc ($30$)          & .970$\pm$.02 & .964$\pm$.02 & .953$\pm$.02 & .878$\pm$.05 & .891$\pm$.06 & {\bf.983$\pm$.02} \\
wine ($13$)         & .994$\pm$.00 & .829$\pm$.12 & {\bf.998$\pm$.00} & .893$\pm$.08 & .817$\pm$.14 & .982$\pm$.02 \\
 \hline
    \end{tabular}
    \caption{Comparison with competitors in the ODDS datasets.}
    \label{tab:odds}
\end{table}

\begin{table}[!t]
\setlength{\tabcolsep}{4pt}
    \centering
    MNIST\vspace{0.5ex}\\
    \begin{tabular}{|c||c|c|c|c|c|c|c||c|}
    \hline
\it& \rot{$\ourmethod$} & \rot{OC-SVM} & \rot{AE} & \rot{$A^3$} & \rot{Deep-SAD} & \rot{Neg-AE} & \rot{DRA} & \textit{mean}\\
\hline\hline
  $\ourmethod$ & --- & \bf 1.00 & \bf 0.91 & \bf 1.00 & \bf 0.85 & \bf 0.79 & \bf 0.96 & \bf 0.92\\
  \hline
  OC-SVM & 0.00 & --- & 0.00 & \bf 0.86 & 0.05 & 0.01 & \bf 0.72 & 0.27\\
  \hline
  AE & 0.09 & \bf 1.00 & --- & \bf 1.00 & \bf 0.70 & \bf 0.51 & \bf 0.93 & 0.70\\
  \hline
  $A^3$ & 0.00 & 0.14 & 0.00 & --- & 0.02 & 0.00 & 0.49 & 0.11\\
  \hline
  Deep-SAD & 0.15 & \bf 0.95 & 0.30 & \bf 0.98 & --- & 0.24 & \bf 0.91 & 0.59\\   
  \hline
  Neg-AE &  0.21 & \bf 0.99 & 0.49 & \bf 1.00 & \bf 0.76 & --- & \bf 0.79 & 0.71\\
  \hline
  DRA & 0.04 & 0.28 & 0.07 & \bf 0.51 & 0.09 & 0.21 & --- & 0.20 \\
  \hline\hline
  \textit{mean} & \bf 0.08 & 0.73 & 0.30 & 0.89 & 0.41 & 0.29 & 0.80 & ---\\
  \hline
    \end{tabular}

    \vspace{2ex} 
Fashion MNIST\vspace{0.5ex}\\
\begin{tabular}{|c||c|c|c|c|c|c|c||c|}
\hline
& \rot{$\ourmethod$} & \rot{OC-SVM} & \rot{AE} & \rot{$A^3$} & \rot{Deep-SAD} & \rot{Neg-AE} & \rot{DRA} & \textit{mean}\\
    \hline\hline
$\ourmethod$ & --- & \bf 1.00 & \bf 0.98 & \bf 0.98 & \bf 0.64 & \bf 0.51 & \bf 0.96 & \bf 0.85 \\
\hline
OC-SVM & 0.00 & --- & 0.01 & \bf 0.90 & 0.11 & 0.02 & \bf 0.91 & 0.33 \\
\hline
AE & 0.02 & \bf 0.99 & --- & \bf 0.97 & 0.38 & 0.17 & \bf 0.94 & 0.58 \\
\hline
$A^3$ & 0.02 & 0.10 & 0.03 & --- & 0.03 & 0.02 & \bf 0.72 & 0.15 \\
\hline
Deep-SAD & 0.36 & \bf 0.89 & 0.62 & \bf 0.97 & --- & 0.38 & \bf 0.95 & 0.70 \\
\hline
Neg-AE & 0.49 & \bf 0.98 & \bf 0.83 & \bf 0.98 & \bf 0.62 & --- & \bf 0.79 & 0.78 \\
\hline
DRA & 0.04 & 0.09 & 0.06 & 0.28 & 0.05 & 0.21 & --- & 0.12 \\
\hline\hline
\textit{mean} & \bf 0.15 & 0.67 & 0.42 & 0.85 & 0.30 & 0.22 & 0.88 & ---\\
\hline
    \end{tabular}

\vspace{2ex}
CIFAR-10 \vspace{0.5ex}\\
\begin{tabular}{|c||c|c|c|c|c|c|c||c|}
\hline
& \rot{$\ourmethod$} & \rot{OC-SVM} & \rot{AE} & \rot{$A^3$} & \rot{Deep-SAD} & \rot{Neg-AE} & \rot{DRA} & \textit{mean}\\
    \hline\hline
$\ourmethod$ & --- & \bf 0.97 & \bf 1.00  & \bf 0.86 & \bf 0.81 & \bf 0.73 & \bf 0.80 & \bf 0.82 \\
\hline
OC-SVM & 0.03 & --- & 0.48 & 0.39 & 0.29 & 0.26 & \bf 0.67 & 0.35 \\
\hline
AE & 0.00 & \bf 0.52  & --- & \bf 0.54 & 0.42 & 0.26 & \bf 0.63 & 0.39 \\
\hline
$A^3$ & 0.14 & \bf 0.61 & 0.46 & --- & 0.48 & 0.04 & 0.06 & 0.30 \\
\hline
Deep-SAD & 0.19 & \bf 0.71 & \bf 0.58 & \bf 0.52 & --- & 0.05 & 0.08 & 0.35 \\
\hline
Neg-AE & 0.27 & \bf 0.74 & \bf 0.74 & \bf 0.96 & \bf 0.95 & --- & \bf 0.67 & 0.72 \\
\hline
DRA & 0.20 & 0.33 & 0.37 & \bf 0.94 & \bf 0.92 & 0.33 & --- & 0.51 \\
\hline\hline
\textit{mean} & \bf 0.18 & 0.65 & 0.61 & 0.70 & 0.65 & 0.28 & 0.49 & ---\\
\hline
    \end{tabular}

    \caption{Probability of winning for each pair of methods in the one-vs-one experimental setting.}
    \label{tab:2classes_competitors}
\end{table}

\begin{table}[!t]
\setlength{\tabcolsep}{3pt}
\scriptsize
\centering
MNIST \vspace{0.5ex}\\
\begin{tabular}{|c|cccccccccc|c|}
    \hline
\diagbox[width=4em]{N}{A} & 0 & 1 & 2 & 3 & 4 & 5 & 6 & 7 & 8 & 9 & $\mu$  \\
\hline
0 & ~$-$ & 1.00 & 1.00 & 1.00 & 0.97 & 1.00 & 1.00 & 0.95 & 1.00 & 0.98 & 0.99 \\
1 & 0.95 & ~$-$ & 0.90 & 0.95 & 1.00 & 0.98 & 0.95 & 1.00 & 0.97 & 1.00 & 0.97 \\
2 & 0.77 & 1.00 & ~$-$ & 0.83 & 0.87 & 0.87 & 0.72 & 0.85 & 0.93 & 0.88 & 0.86 \\
3 & 0.85 & 0.98 & 0.97 & ~$-$ & 0.90 & 0.97 & 0.97 & 0.97 & 0.98 & 0.97 & 0.95 \\
4 & 0.70 & 0.97 & 1.00 & 0.95 & ~$-$ & 0.78 & 0.63 & 0.92 & 0.93 & 0.83 & 0.86 \\
5 & 0.85 & 1.00 & 0.97 & 0.98 & 0.90 & ~$-$ & 0.95 & 0.97 & 1.00 & 0.98 & 0.96 \\
6 & 0.78 & 1.00 & 0.90 & 0.93 & 0.80 & 0.93 & ~$-$ & 0.85 & 0.95 & 0.93 & 0.90 \\
7 & 0.88 & 1.00 & 0.92 & 0.88 & 0.92 & 0.93 & 0.88 & ~$-$ & 0.97 & 0.98 & 0.93 \\
8 & 0.63 & 0.92 & 0.82 & 0.75 & 0.75 & 0.82 & 0.73 & 0.72 & ~$-$ & 0.83 & 0.77 \\
9 & 0.83 & 0.97 & 1.00 & 0.97 & 0.95 & 0.97 & 0.93 & 1.00 & 0.92 & ~$-$ & 0.95 \\
\hline
$\mu$ & 0.81 & 0.98 & 0.94 & 0.92 & 0.89 & 0.92 & 0.86 & 0.91 & 0.96 & 0.93 & 0.91 \\
\hline
    \end{tabular}
    \vspace{2ex}\\
Fashion MNIST\vspace{0.5ex}\\
\begin{tabular}{|c|cccccccccc|c|}
    \hline
\diagbox[width=4em]{N}{A} & 0 & 1 & 2 & 3 & 4 & 5 & 6 & 7 & 8 & 9 & $\mu$  \\
\hline
0 & ~$-$ & 0.90 & 0.97 & 0.92 & 0.97 & 0.93 & 0.95 & 0.75 & 0.90 & 0.83 & 0.90 \\
1 & 0.95 & ~$-$ & 0.93 & 0.90 & 0.93 & 0.85 & 0.98 & 0.95 & 0.97 & 0.88 & 0.93 \\
2 & 1.00 & 0.85 & ~$-$ & 0.98 & 0.82 & 0.80 & 0.93 & 0.80 & 0.85 & 0.80 & 0.87 \\
3 & 0.90 & 0.83 & 0.88 & ~$-$ & 0.88 & 0.82 & 0.93 & 0.72 & 0.83 & 0.72 & 0.84 \\
4 & 0.93 & 0.85 & 0.77 & 0.97 & ~$-$ & 0.73 & 0.83 & 0.60 & 0.77 & 0.73 & 0.80 \\
5 & 0.75 & 0.70 & 0.75 & 0.78 & 0.70 & ~$-$ & 0.70 & 0.92 & 0.72 & 0.80 & 0.76 \\
6 & 0.98 & 0.92 & 0.85 & 0.98 & 0.93 & 0.85 & ~$-$ & 0.78 & 1.00 & 0.90 & 0.91 \\
7 & 0.88 & 0.82 & 0.85 & 0.83 & 0.90 & 0.93 & 0.88 & ~$-$ & 0.83 & 0.88 & 0.87 \\
8 & 0.77 & 0.62 & 0.77 & 0.75 & 0.83 & 0.63 & 0.78 & 0.68 & ~$-$ & 0.65 & 0.72 \\
9 & 0.68 & 0.77 & 0.65 & 0.77 & 0.67 & 0.78 & 0.78 & 0.98 & 0.68 & ~$-$ & 0.75 \\
\hline
$\mu$ & 0.87 & 0.81 & 0.82 & 0.88 & 0.85 & 0.81 & 0.86 & 0.80 & 0.84 & 0.80 & 0.83 \\
\hline
    \end{tabular}
    \vspace{2ex}\\
    CIFAR-10\vspace{0.5ex}\\
\begin{tabular}{|c|cccccccccc|c|}
    \hline
\diagbox[width=4em]{N}{A} & 0 & 1 & 2 & 3 & 4 & 5 & 6 & 7 & 8 & 9 & $\mu$  \\
\hline
0 & ~$-$ & 0.83 & 0.93 & 0.92 & 0.98 & 0.92 & 0.98 & 0.95 & 0.77 & 0.88 & 0.91\\
1 & 0.82 & ~$-$ & 0.80 & 0.78 & 0.92 & 0.85 & 0.83 & 0.75 & 0.78 & 0.65 & 0.80\\
2 & 0.95 & 1.00 & ~$-$ & 0.98 & 0.98 & 0.93 & 0.95 & 0.93 & 1.00 & 0.97 & 0.97\\
3 & 0.83 & 0.95 & 0.88 & ~$-$ & 0.90 & 0.87 & 0.87 & 0.87 & 0.93 & 0.98 & 0.90\\
4 & 0.98 & 0.98 & 0.97 & 0.95 & ~$-$ & 0.95 & 1.00 & 0.98 & 1.00 & 0.97 & 0.98\\
5 & 0.88 & 0.87 & 0.73 & 0.75 & 0.88 & ~$-$ & 0.62 & 0.65 & 0.95 & 0.85 & 0.80\\
6 & 1.00 & 0.98 & 0.98 & 0.98 & 1.00 & 0.98 & ~$-$ & 1.00 & 0.98 & 0.98 & 0.99\\
7 & 0.88 & 0.75 & 0.87 & 0.93 & 0.87 & 0.68 & 0.78 & ~$-$ & 0.97 & 0.75 & 0.83\\
8 & 0.83 & 0.85 & 0.95 & 0.85 & 0.98 & 0.95 & 0.97 & 0.93 & ~$-$ & 0.87 & 0.91\\
9 & 0.65 & 0.55 & 0.88 & 0.70 & 0.85 & 0.75 & 0.70 & 0.68 & 0.68 & ~$-$ & 0.72\\
\hline
$\mu$ & 0.87 & 0.86 & 0.89 & 0.87 & 0.93 & 0.88 & 0.86 & 0.86 & 0.90 & 0.88 & 0.88\\
\hline
    \end{tabular}
    \caption{Probability of winning of $\ourmethod$ for each pair of normal and anomalous classes in the one-vs-one setting.}    \label{tab:2classes_comparison_eval}
\end{table}

\begin{figure}[!t]
\centering
\includegraphics[width=0.24\textwidth]{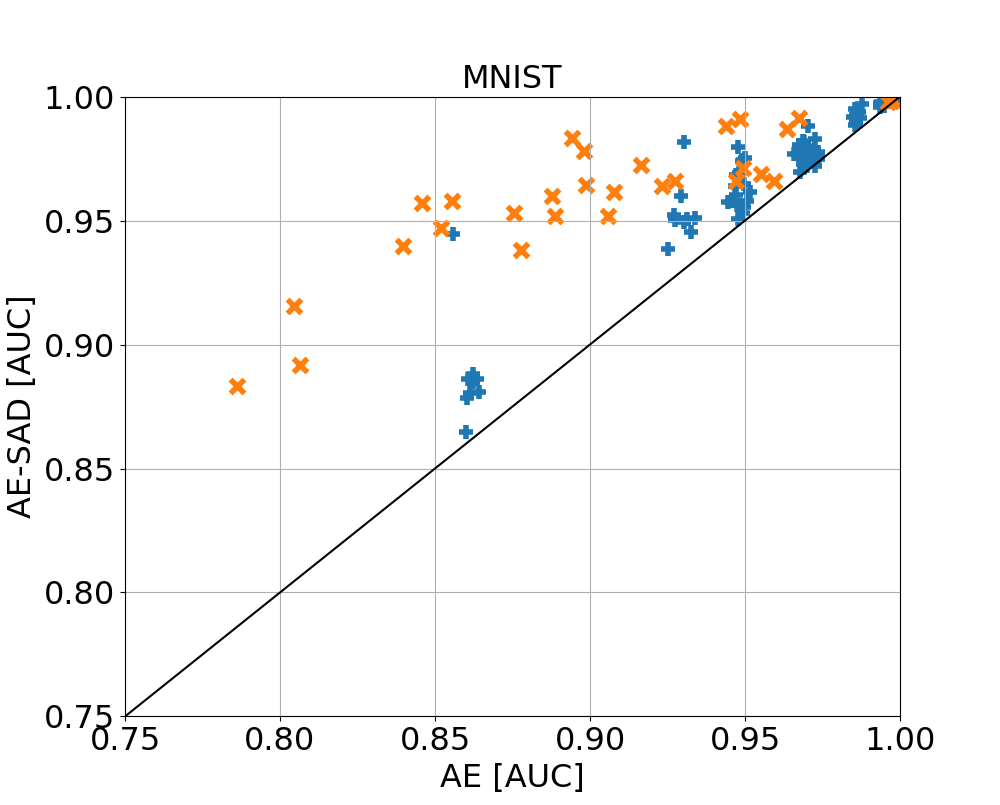} 
\includegraphics[width=0.24\textwidth]{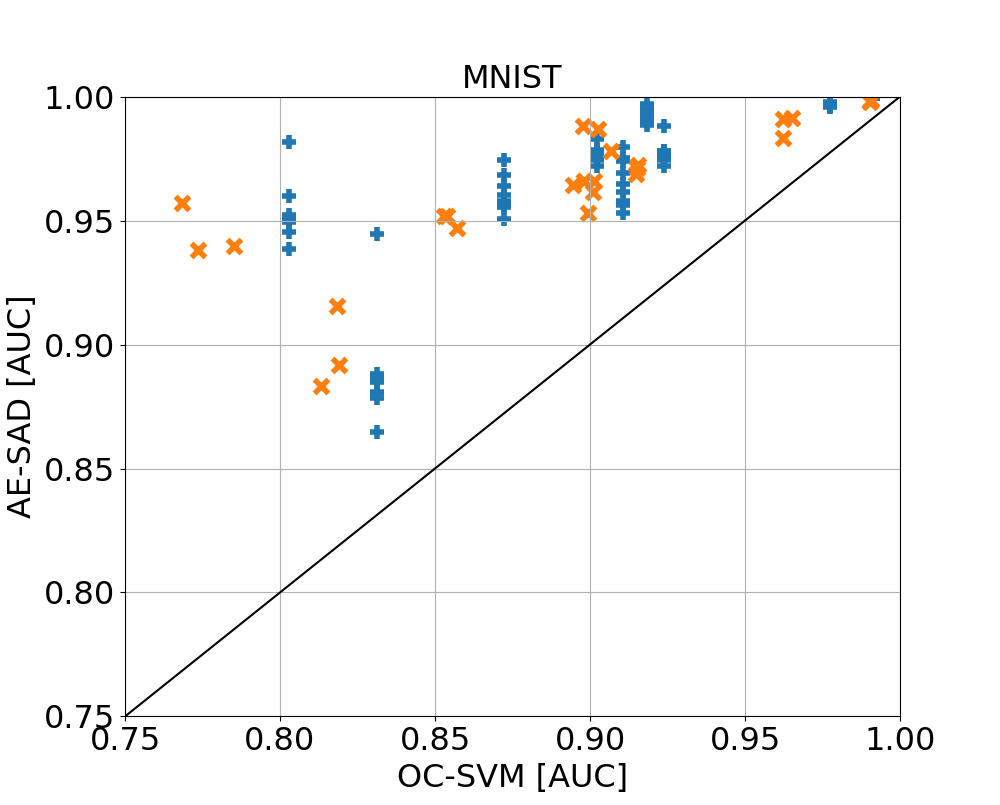} 
\\
\includegraphics[width=0.24\textwidth]{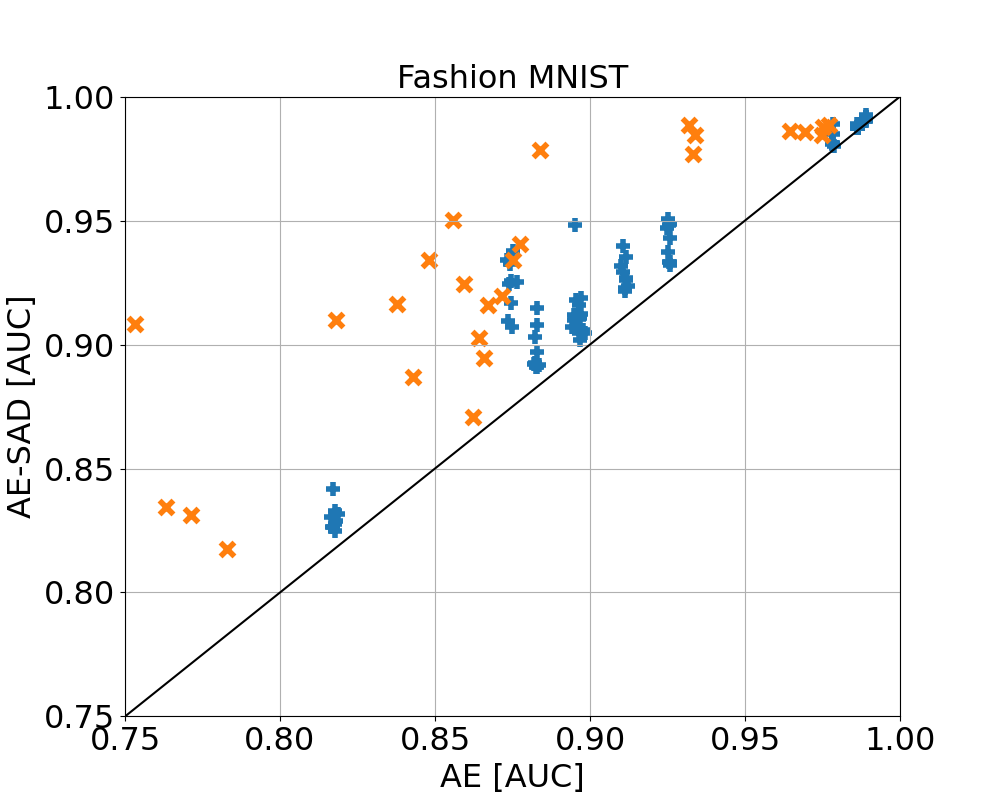} 
\includegraphics[width=0.24\textwidth]{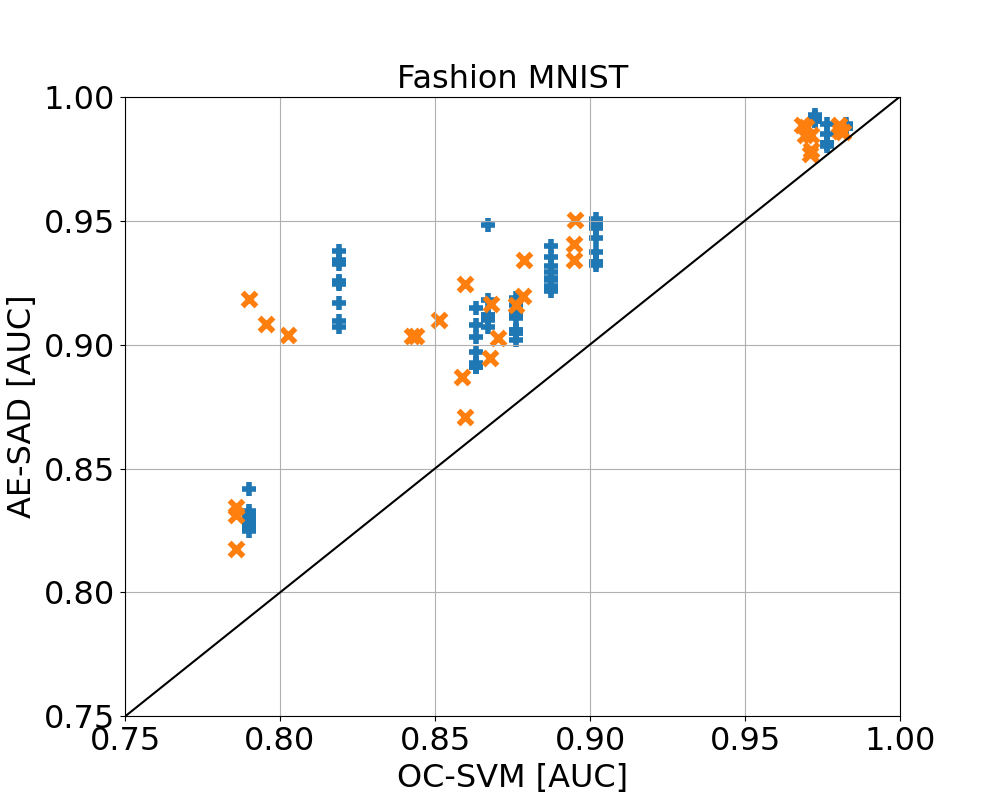} 
\\
\includegraphics[width=0.24\textwidth]{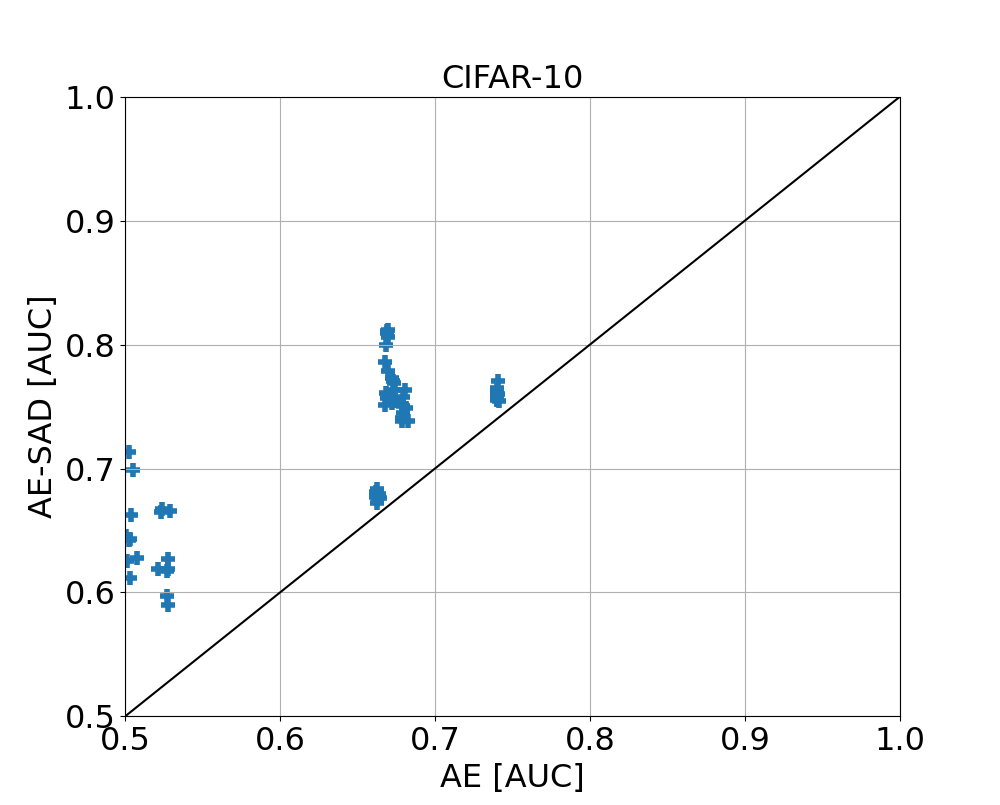} 
\includegraphics[width=0.24\textwidth]{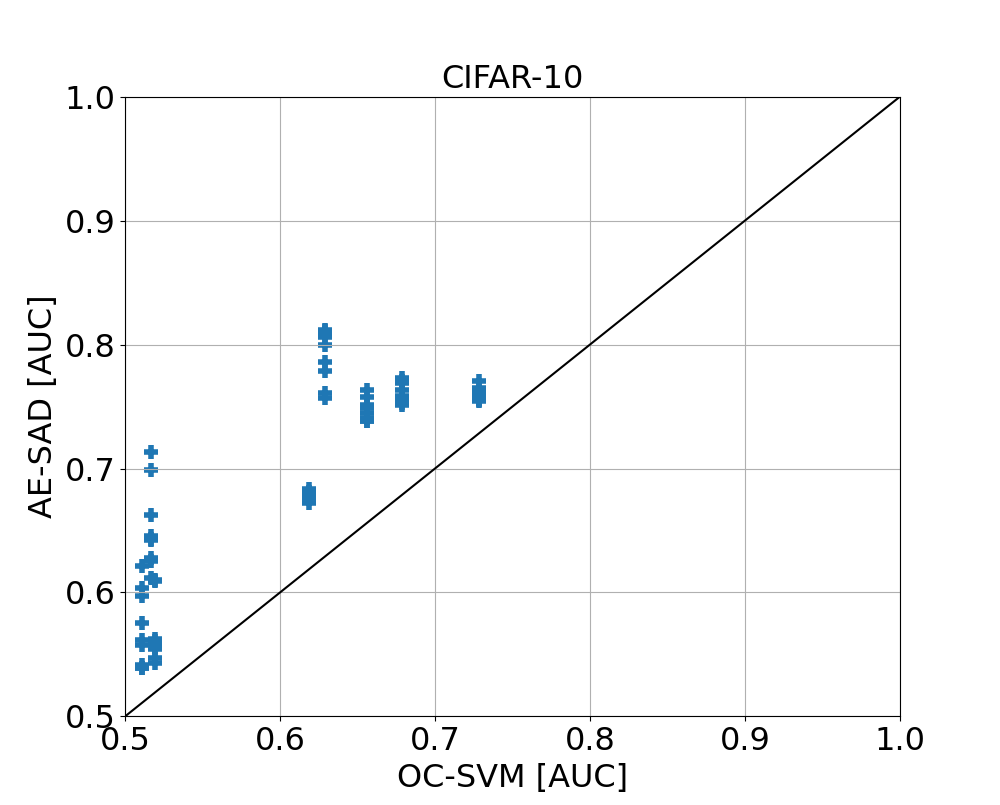} 
\caption{Comparison of the AUCs of $\ourmethod$ and competitors on both clean (blue $+$-marks) and polluted (red $\times$-marks) data. Group 1.}
\label{fig:scatter1}
\end{figure}

\begin{figure}[!t]
\centering
\includegraphics[width=0.24\textwidth]{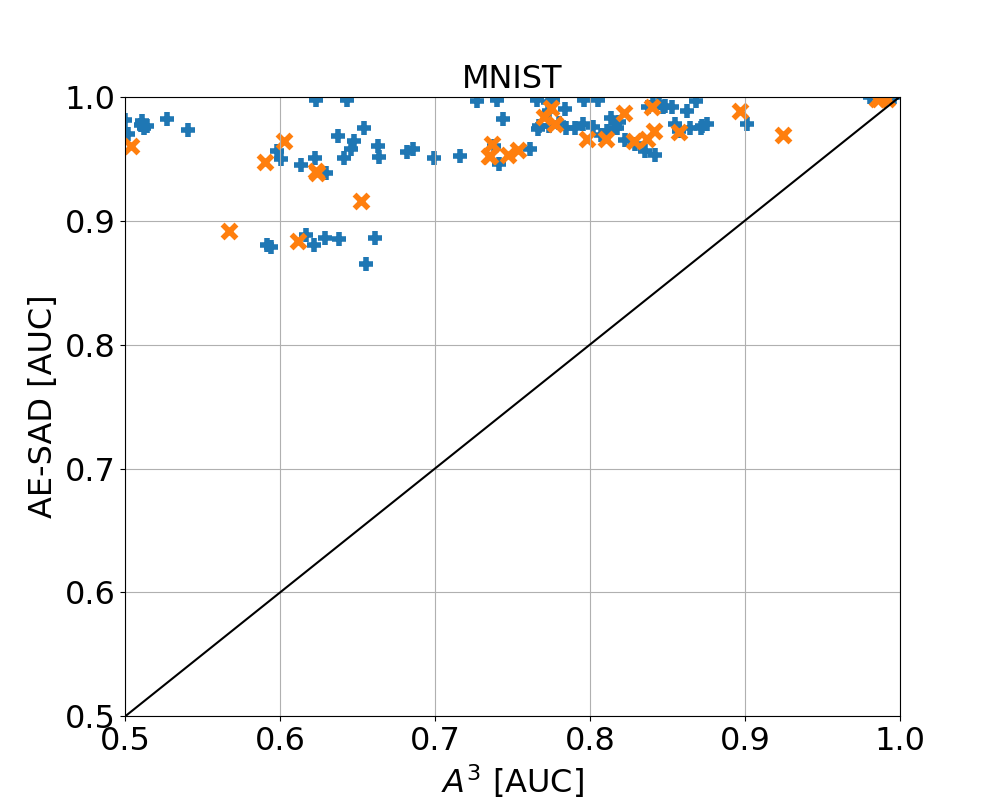} 
\includegraphics[width=0.24\textwidth]{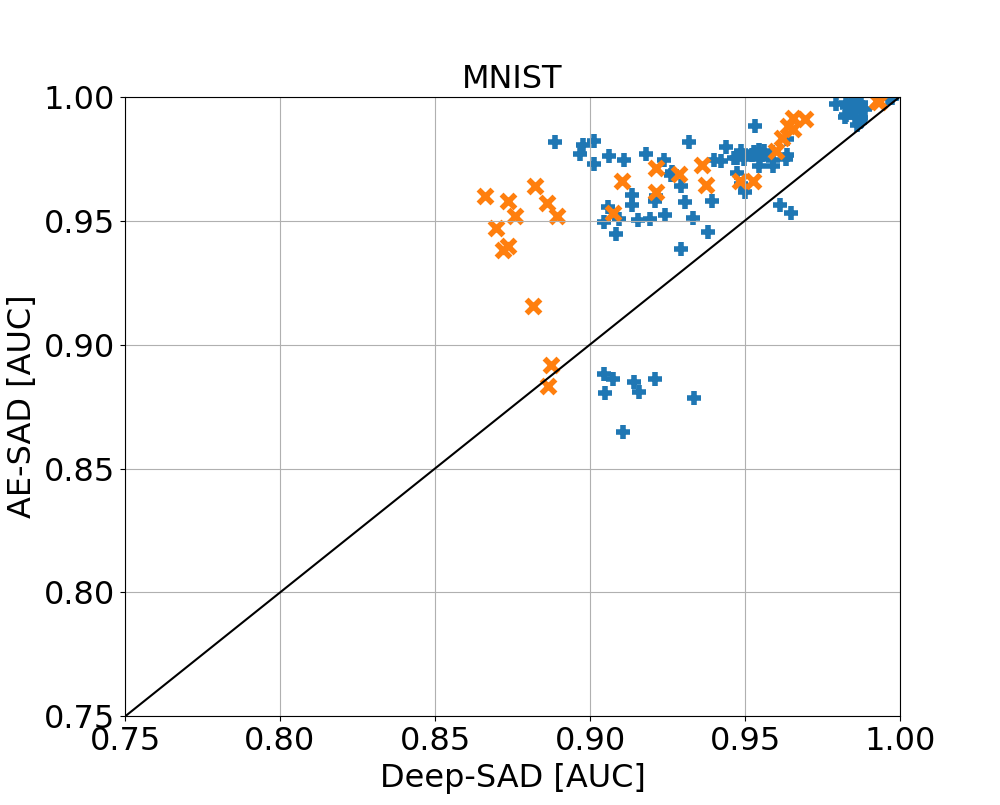} 
\\
\includegraphics[width=0.24\textwidth]{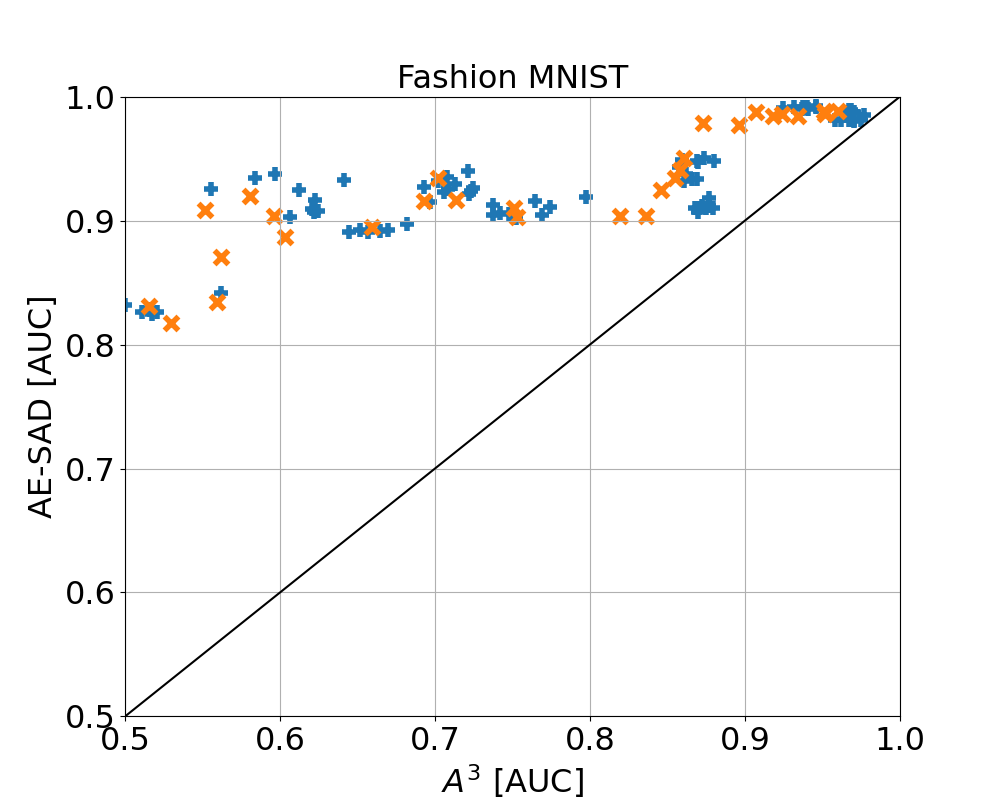} 
\includegraphics[width=0.24\textwidth]{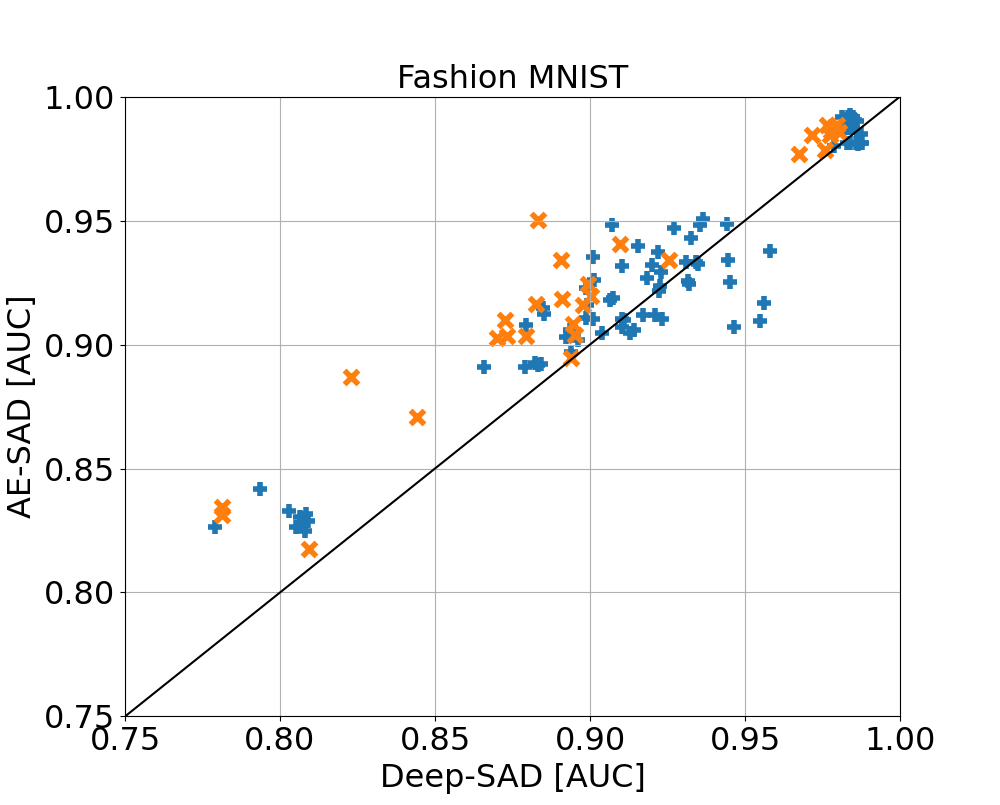} 
\\
\includegraphics[width=0.24\textwidth]{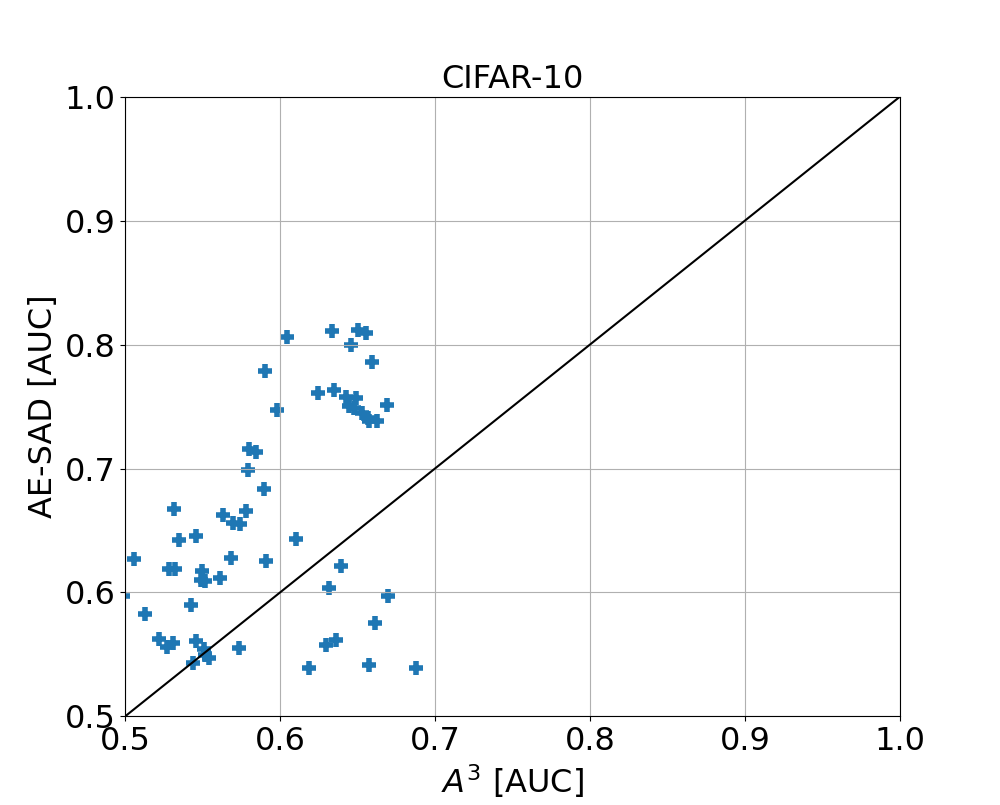} 
\includegraphics[width=0.24\textwidth]{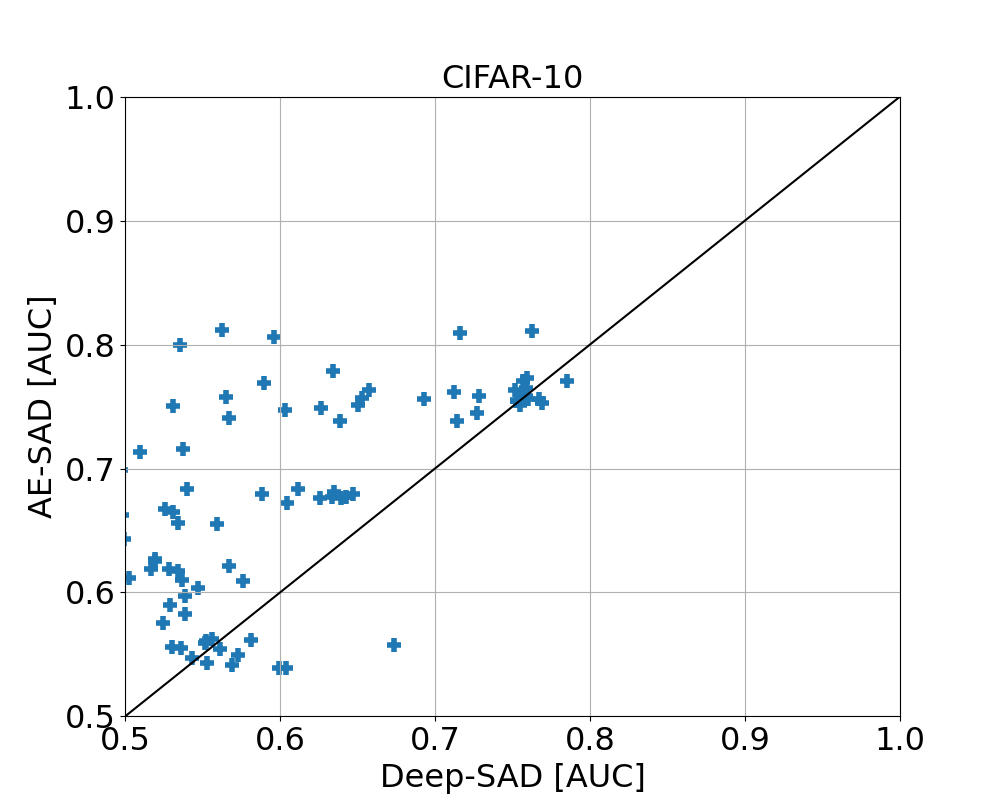} 
\caption{Comparison of the AUCs of $\ourmethod$ and competitors on both clean (blue $+$-marks) and polluted (red $\times$-marks) data. Group 2.}
\label{fig:scatter2}
\end{figure}

\begin{figure}[!t]
\centering
\includegraphics[width=0.24\textwidth]{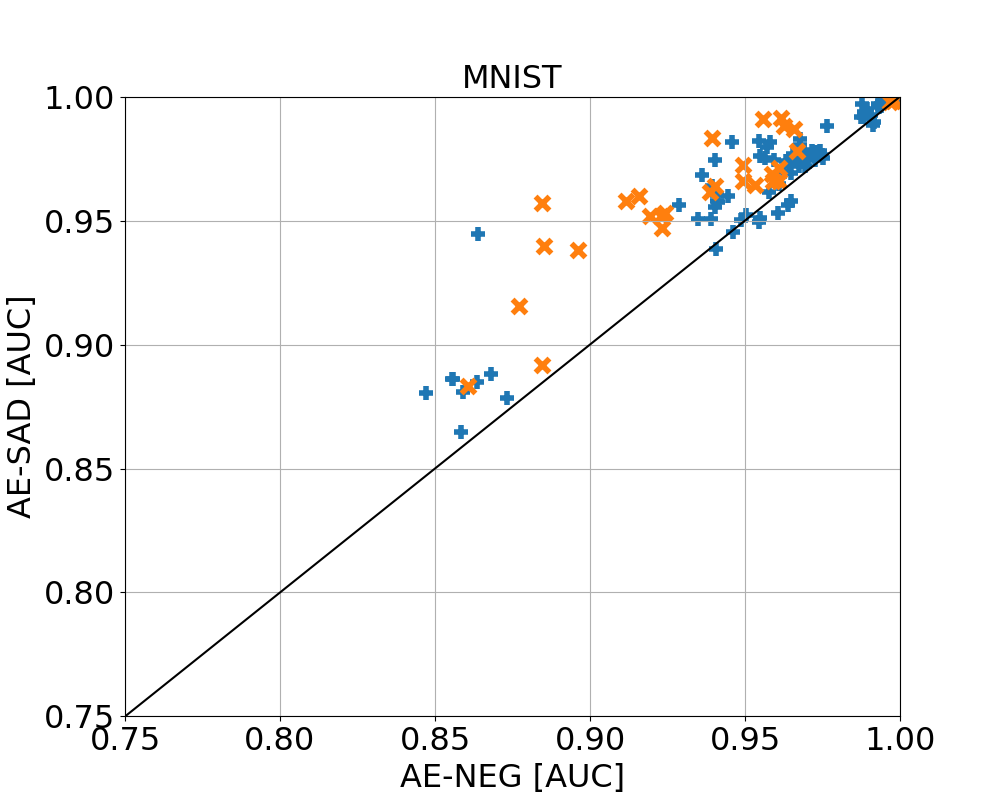} 
\includegraphics[width=0.24\textwidth]{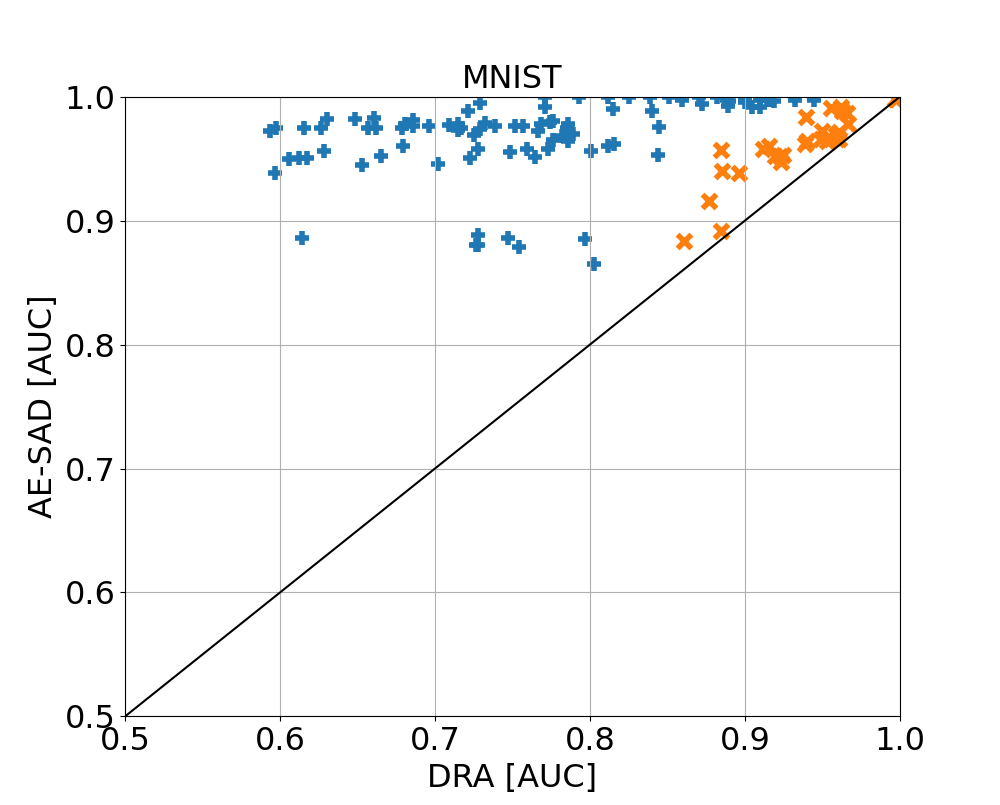} 
\\
\includegraphics[width=0.24\textwidth]{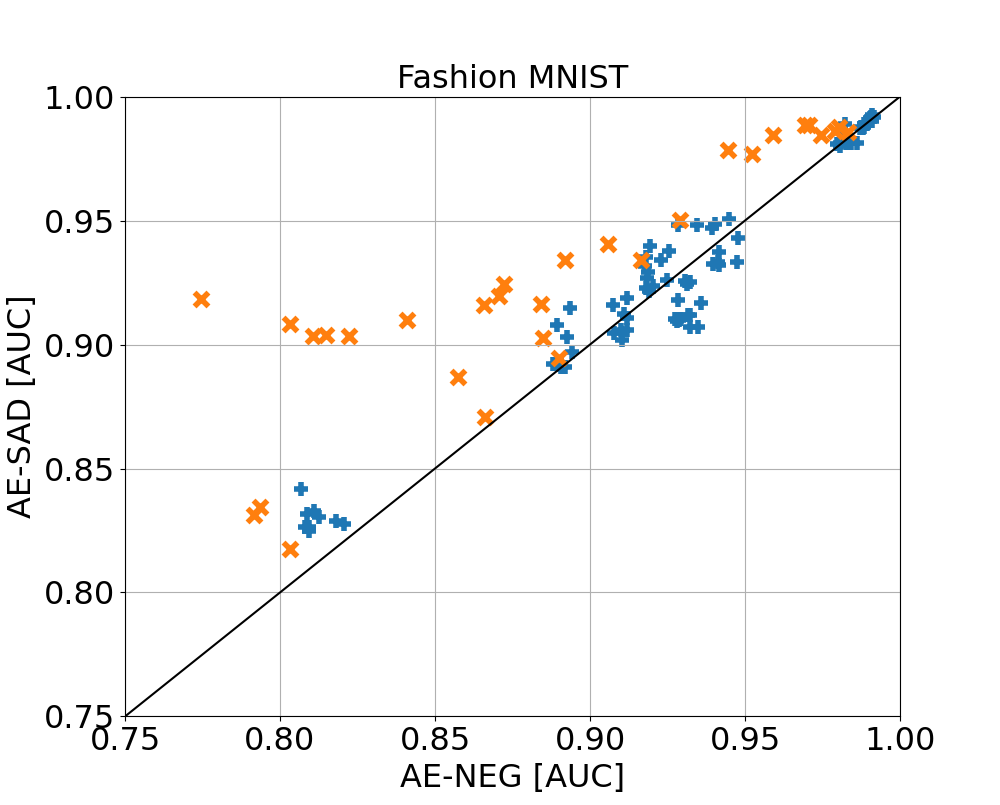} 
\includegraphics[width=0.24\textwidth]{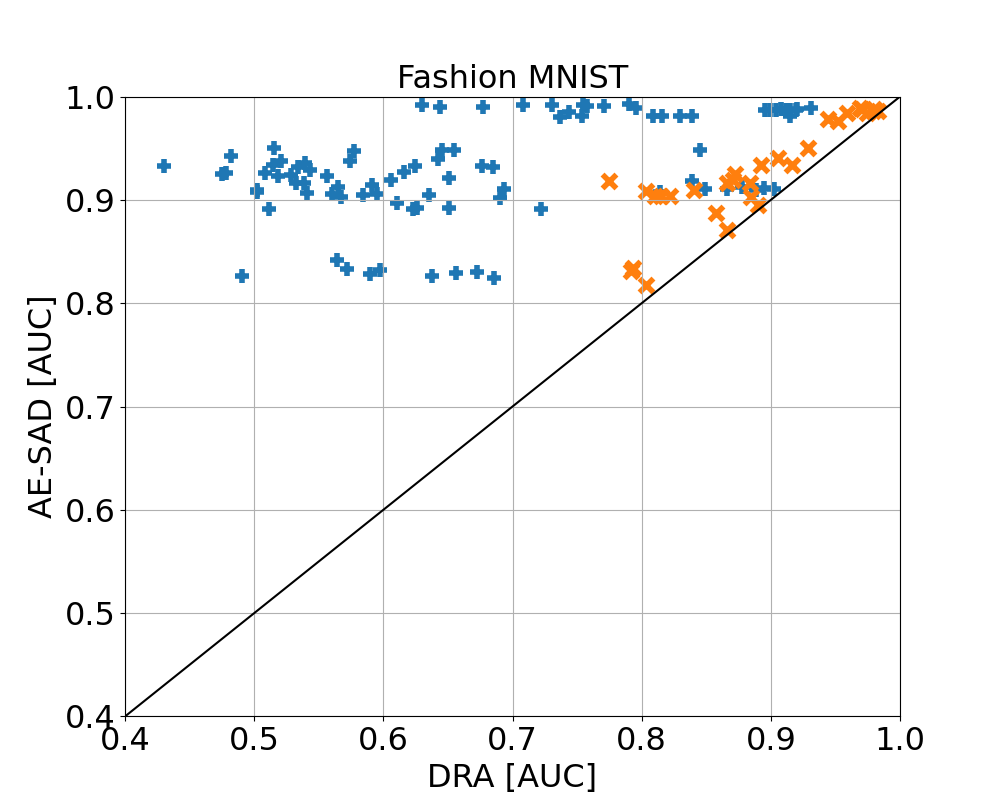} 
\\
\includegraphics[width=0.24\textwidth]{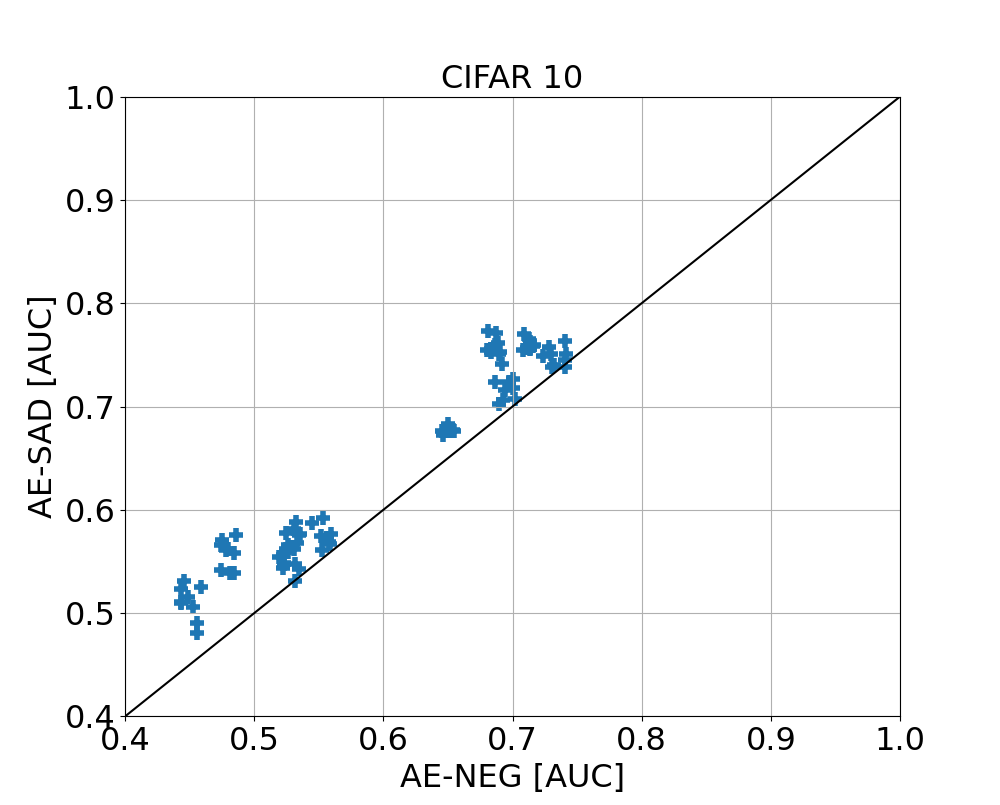} 
\includegraphics[width=0.24\textwidth]{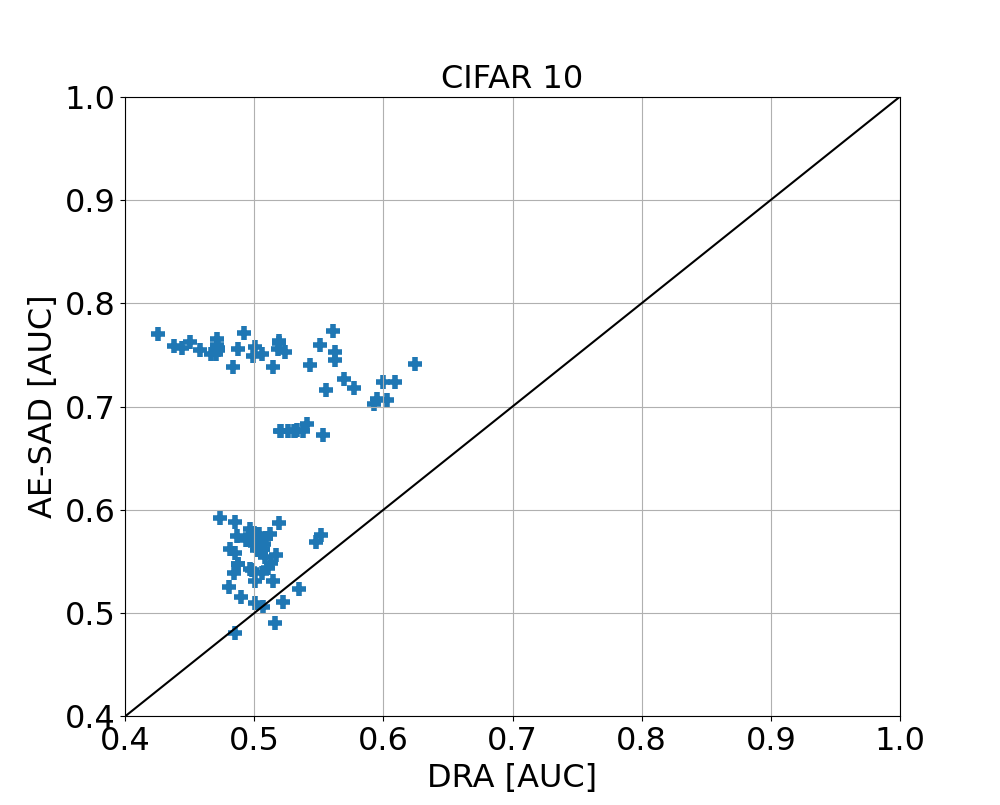} 
\caption{Comparison of the AUCs of $\ourmethod$ and competitors on both clean (blue $+$-marks) and polluted (red $\times$-marks) data. Group 3.}
\label{fig:scatter3}
\end{figure}

We also consider for the comparison with the competitors the MNIST, Fashion-MNIST and CIFAR-10 datasets, for which we implement the one-vs-one setting already illustrated in Section \ref{sect:exp_sens}.
This setting simulates the scenario in which the detector has limited knowledge of the expected anomalies and, as such, it is a notable scenarios to deal with.

Table \ref{tab:2classes_competitors} reports the relative number of wins for each pair of methods on the MNIST, Fashion MNIST, and CIFAR-10 datasets. 
Specifically, each entry of the table is the probability that the method located on the corresponding row scores an AUC greater than that of the method located on the corresponding column.
In the considered setting $\ourmethod$ shows a larger probability of winning over all the competitors on all the datasets. 
Neg-AE is the runner-up in terms of number of wins, and also Deep-SAD achieve sufficient performances, while $A^3$ and DRA do not perform well in this setting despite being semi-supervised methods.
In Section \ref{sect:exp_a3}, we compare $\ourmethod$ and $A^3$ on a setting where competitor performance are shown in \cite{a3} to be strong across various configurations.

Table \ref{tab:2classes_comparison_eval} details the result for each pair of normal and known anomalous classes, by reporting the relative number of wins of $\ourmethod$ on the competitors. 
Specifically, each entry of the table is the probability that the AUC scored by $\ourmethod$ is greater than the AUC scored by another method. For almost all the combinations $\ourmethod$ is preferable to a randomly selected method of the pool.

To provide details also on the comparison of the methods on the each single dataset,
Figures \ref{fig:scatter1}, \ref{fig:scatter2}, \ref{fig:scatter3} reports the scatter plot of the AUC of $\ourmethod$ (on the $y$-axis) vs the AUC of each competitor (on the $x$-axis) associated with each of the above experiments.
Specifically, blue $+$-marks are relative to the experiments described in this section, while red $\times$-marks are relative experiments reported in the next section concerning polluted normal data. 

\subsection{Behaviour varying the image resolution}
In order to investigate how $\ourmethod$ performances vary with images of different size, we consider the FRUIT dataset \cite{fruit}. It is composed of high resolution images of 15 types of fruit, each of which represents a distinct class. Each class consists of at least two thousands examples (on the average about $3000$ items per class).
We consider a \emph{one-vs-all} strategy, in which the training set is built by considering one class as normal and by randomly sampling 50 anomalies from the other classes.
Table \ref{tab:fruit} details the comparison between $\ourmethod$ applied to the images scaled at different sizes (namely $16\times24$, $32\times48$, $64\times96$, and $128\times192$ have been considered) and the competitors applied to the images with size $128\times192$.
We can observe that high resolution images do not cause performance degradation. On the contrary, as reported in Table \ref{tab:fruit}, $\ourmethod$  benefits of larger resolution inputs, since the AUC  is increasing with the image size. 
Moreover, $\ourmethod$ outperforms all the competitors even when it deals with high resolution images.

\begin{table*}[h!]
        \centering      
        \begin{tabular}{|c|c|c|c|c|cccccc|}
        \hline
Methods & \multicolumn{4}{c|}{$\ourmethod$} & AE & OC-SVM & $A^3$ & Deep-SAD & Neg-AE & DRA\\
\hline
Class & $16\times24$ & $32\times48$ & $64\times96$ & $128\times192$ & \multicolumn{6}{c|}{$128\times192$} \\
\hline 
Apple        & .648$\pm$.03 & .810$\pm$.03 & .916$\pm$.02 & \bf .939$\pm$.01 & .473$\pm$.01 & .309$\pm$.00 & .629$\pm$.04 & .516$\pm$.04 & .468$\pm$.01 & .747$\pm$.08\\
Banana       & .877$\pm$.01 & .899$\pm$.01 & .956$\pm$.01 & \bf .984$\pm$.01 & .611$\pm$.02 & .405$\pm$.00 & .640$\pm$.17 & .491$\pm$.05 & .554$\pm$.08 & .660$\pm$.16\\
Carambola    & .869$\pm$.03 & .940$\pm$.04 & \bf .979$\pm$.02 & \bf .979$\pm$.02 & .907$\pm$.01 & .360$\pm$.00 & .478$\pm$.07 & .462$\pm$.04 & .653$\pm$.05 & .926$\pm$.03\\
Guava        & .912$\pm$.04 & .979$\pm$.01 & .984$\pm$.01 & \bf .997$\pm$.00 & .953$\pm$.01 & .934$\pm$.00 & .726$\pm$.07 & .784$\pm$.05 & .894$\pm$.07 & .722$\pm$.12\\
Kiwi         & .988$\pm$.00 & .992$\pm$.00 & \bf .996$\pm$.00 & .995$\pm$.00 & .955$\pm$.00 & .942$\pm$.00 & .805$\pm$.07 & .783$\pm$.04 & .970$\pm$.01 & .858$\pm$.20\\
Mango        & .891$\pm$.03 & .889$\pm$.06 & .936$\pm$.01 & \bf .949$\pm$.02 & .579$\pm$.02 & .572$\pm$.00 & .813$\pm$.07 & .647$\pm$.05 & .543$\pm$.02 & .540$\pm$.08\\
Muskmelon    & .965$\pm$.01 & .985$\pm$.01 & \bf .994$\pm$.00 & \bf .994$\pm$.00 & .893$\pm$.00 & .776$\pm$.00 & .851$\pm$.02 & .733$\pm$.04 & .563$\pm$.09 & .948$\pm$.02\\
Orange       & .989$\pm$.00 & .997$\pm$.00 & \bf .999$\pm$.00 & .998$\pm$.00 & .771$\pm$.01 & .493$\pm$.00 & .707$\pm$.10 & .703$\pm$.04 & .641$\pm$.02 & .783$\pm$.15\\
Peach        & .973$\pm$.00 & .988$\pm$.00 & .992$\pm$.00 & \bf .997$\pm$.00 & .963$\pm$.00 & .901$\pm$.00 & .757$\pm$.04 & .802$\pm$.02 & .846$\pm$.06 & .837$\pm$.12\\
Pear         & .935$\pm$.03 & .935$\pm$.02 & .966$\pm$.03 & \bf .974$\pm$.02 & .700$\pm$.01 & .631$\pm$.00 & .721$\pm$.07 & .679$\pm$.03 & .643$\pm$.05 & .756$\pm$.08\\
Persimmon    & .978$\pm$.01 & .985$\pm$.01 & .997$\pm$.01 & \bf 1.00$\pm$.00 & .563$\pm$.01 & .585$\pm$.00 & .905$\pm$.04 & .596$\pm$.05 & .494$\pm$.01 & .961$\pm$.02\\
Pitaya       & .970$\pm$.02 & .991$\pm$.00 & .994$\pm$.01 & \bf .998$\pm$.00 & .425$\pm$.01 & .522$\pm$.00 & .896$\pm$.04 & .667$\pm$.01 & .487$\pm$.07 & .734$\pm$.13\\
Plum         & .998$\pm$.00 & \bf 1.00$\pm$.00 & \bf 1.00$\pm$.00 & \bf 1.00$\pm$.00 & .988$\pm$.00 & .826$\pm$.00 & .901$\pm$.02 & .820$\pm$.04 & .792$\pm$.13 & .872$\pm$.24\\
Pomegranate  & .964$\pm$.01 & .982$\pm$.01 & .988$\pm$.01 & \bf .994$\pm$.00 & .359$\pm$.01 & .612$\pm$.00 & .926$\pm$.02 & .699$\pm$.05 & .428$\pm$.01 & .822$\pm$.10\\
Tomatoes     & .944$\pm$.01 & .976$\pm$.01 & .990$\pm$.00 & \bf .995$\pm$.00 & .327$\pm$.01 & .251$\pm$.00 & .805$\pm$.10 & .580$\pm$.05 & .437$\pm$.08 & .830$\pm$.05\\
\hline
        \end{tabular}
          \caption{$\ourmethod$ applied to the FRUIT dataset scaled at different dimensions and compared with comparison on the size $128\times192$.}
        \label{tab:fruit}
\end{table*}

\subsection{Normal data polluted}\label{sect:exp_comp_poll}

\begin{figure}[t!]
\centering
\includegraphics[width=0.48\textwidth]{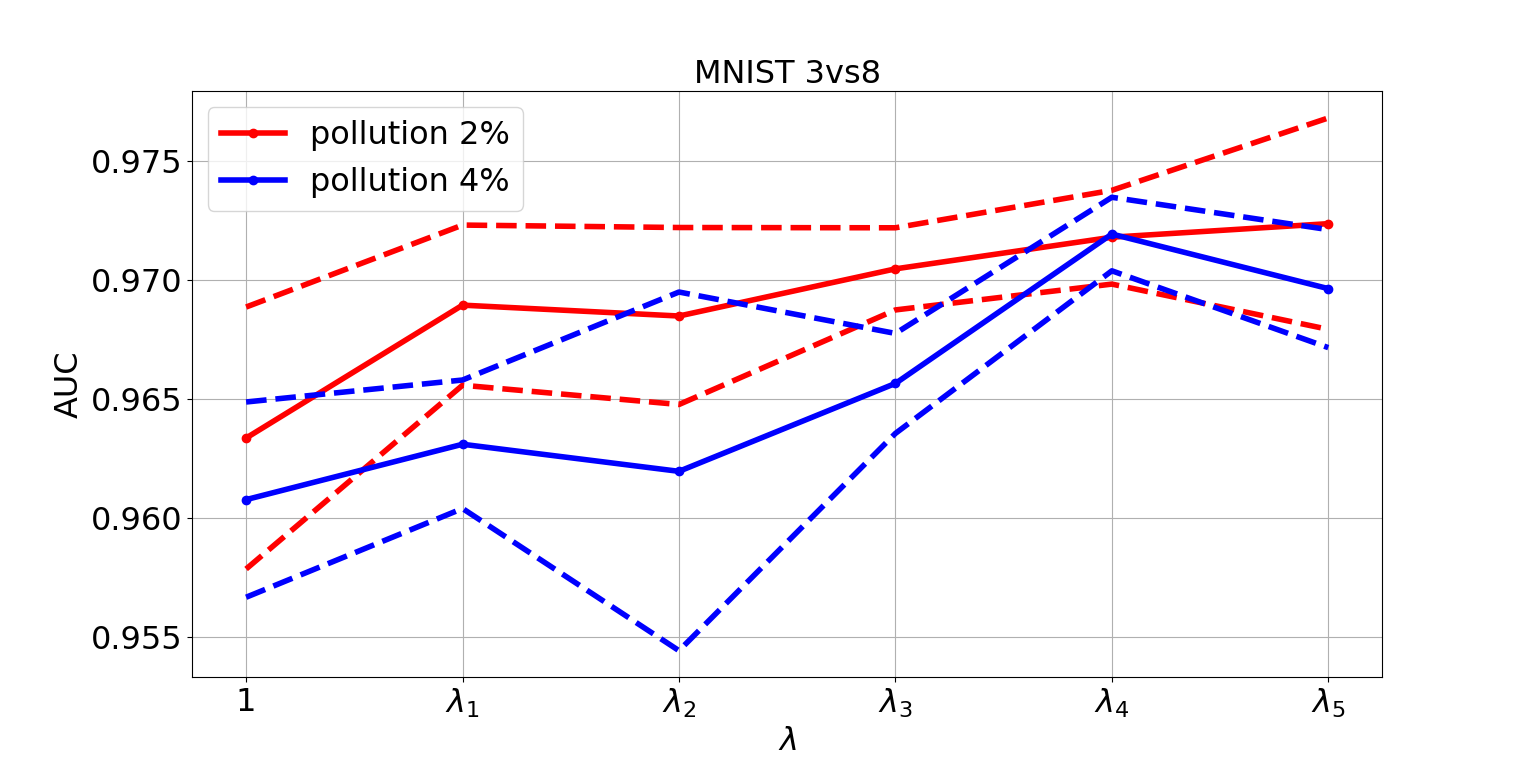} \\
\includegraphics[width=0.48\textwidth]{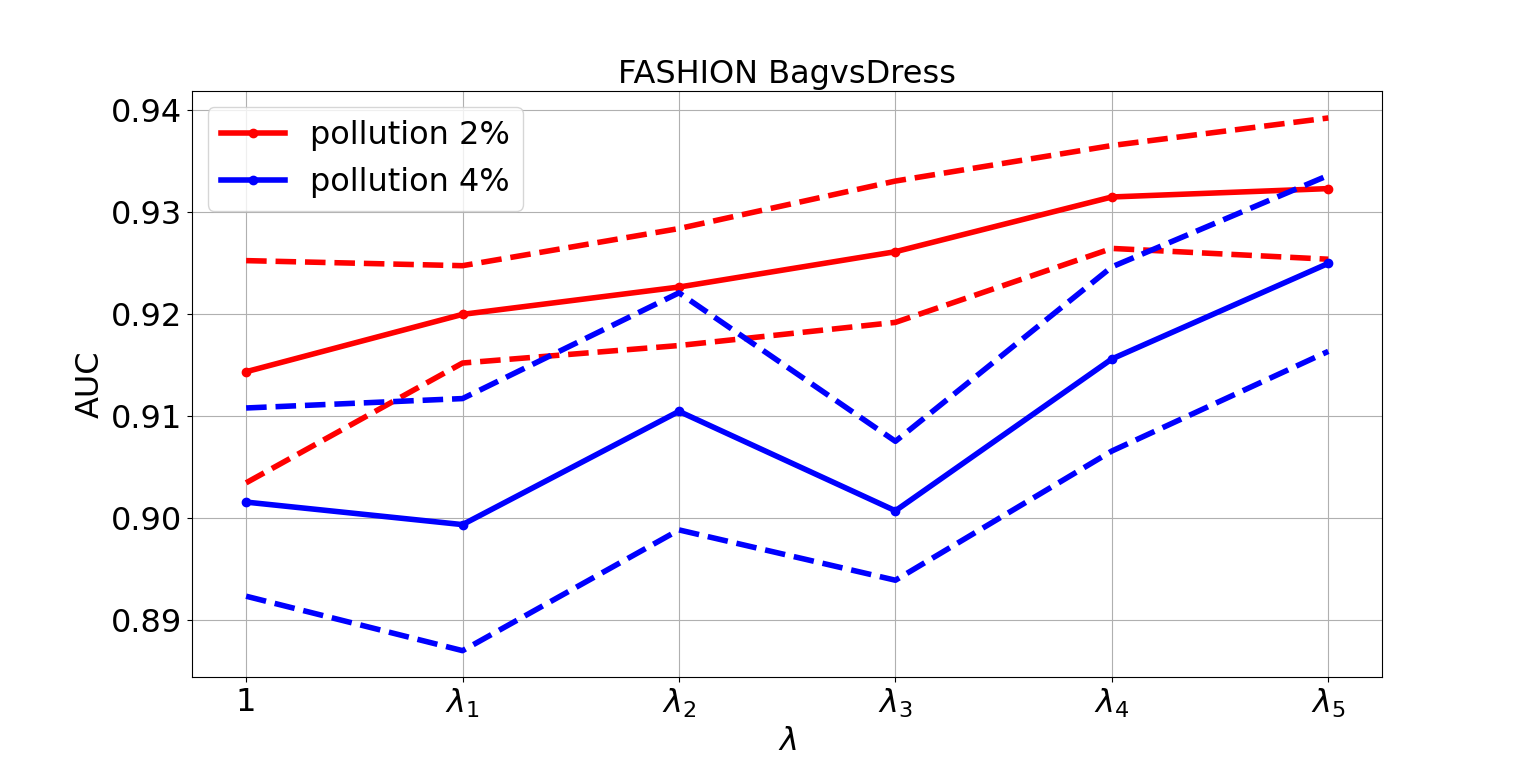} 
\caption{Sensitivity analysis of $\lambda$ for different levels of pollution.}
\label{fig:pollution_lambda}
\end{figure}

\begin{table}[t!]
\setlength{\tabcolsep}{2pt}
    \centering
    \begin{tabular}{|c|ccccc|}
    \multicolumn{6}{c}{MNIST 3vs[4,5,6]}\\
    \hline
\multirow{2}{*}{Method} &  \multicolumn{5}{c|}{Pollution} \\
\cline{2-6}
 & $0.25\%$ & $0.5\%$ & $2.5\%$ & $5\%$ & $12.5\%$ \\
\hline
$\ourmethod$   &  \bf .968$\pm$.01 & \bf .966$\pm$.00 & \bf .957$\pm$.01 & \bf .961$\pm$.01 & \bf .955$\pm$.00 \\
Standard AE & .944$\pm$.01 & .942$\pm$.00 & .919$\pm$.01 & .904$\pm$.01 & .864$\pm$.01\\
OC-SVM & .871$\pm$.00 & .870$\pm$.00 & .863$\pm$.00 & .854$\pm$.00 & .821$\pm$.00 \\
$A^3$ & .622$\pm$.04 & .675$\pm$.03 & .725$\pm$.10 & .653$\pm$.08 & .603$\pm$.08\\
Deep-SAD & .958$\pm$.10 & .957$\pm$.00 & .948$\pm$.00 & .944$\pm$.01 & .911$\pm$.03\\
AE-Neg & .940$\pm$.00 & .942$\pm$.00 & .933$\pm$.00 & .926$\pm$.00 & .910$\pm$.10 \\
DRA & .941$\pm$.05 & .908$\pm$.07 & .909$\pm$.01 & .893$\pm$.05 & .785$\pm$.16 \\
\hline
\multicolumn{6}{c}{~}\\
\multicolumn{6}{c}{Fashion MNIST 3vs[7,8,9]}\\
    \hline
\multirow{2}{*}{Method} &  \multicolumn{5}{c|}{Pollution} \\
\cline{2-6}
 & $0.25\%$ & $0.5\%$ & $2.5\%$ & $5\%$ & $12.5\%$ \\
\hline
$\ourmethod$   & \bf .946$\pm$.00 & \bf .944$\pm$.00 & \bf .945$\pm$.00 & \bf .942$\pm$.00 & \bf .939$\pm$.00\\
Standard AE & .917$\pm$.00 & .908$\pm$.00 & .890$\pm$.00 & .880$\pm$.00 & .858$\pm$.00\\
OC-SVM & .902$\pm$.00 & .901$\pm$.00 & .898$\pm$.00 & .895$\pm$.00 & .880$\pm$.00\\
$A^3$ & .877$\pm$.01 & .875$\pm$.01 & .861$\pm$.01 & .852$\pm$.01 & .838$\pm$.01\\
Deep-SAD & .925$\pm$.01 & .913$\pm$.01 & .904$\pm$.02 & .905$\pm$.01 & .901$\pm$.01\\
AE-Neg & .941$\pm$.00 & .939$\pm$.00 & .932$\pm$.01 & .921$\pm$.00 & .887$\pm$.02\\
DRA & .774$\pm$.09 & .681$\pm$.05 & .631$\pm$.11 & .576$\pm$.11 & .456$\pm$.14 \\
\hline
    \end{tabular}
    \caption{Comparison with competitors for various percentage levels of pollution.}
    \label{tab:pollutioncompex}
\end{table}

\begin{table}[!t]
\setlength{\tabcolsep}{0.5pt}
    \centering
    \begin{tabular}{|c|c|c|c|c|c|c|c|}
    \multicolumn{8}{c}{MNIST} \\
\hline 
Class & $\ourmethod$ & AE & OC-SVM & $A^3$ & Deep-SAD & Neg-AE & DRA\\
\hline
0 & \textbf{.980$\pm$.01} & .885$\pm$.01 & \it .911$\pm$.00 & .770$\pm$.02 & .910$\pm$.04 & .903$\pm$.01 & .609$\pm$.27\\
1 & \textbf{.996$\pm$.00} & \it .996$\pm$.00 & .986$\pm$.00 & .983$\pm$.04 & .983$\pm$.01 & .996$\pm$.00 & .367$\pm$.28\\
2 & \textbf{.913$\pm$.02} & .767$\pm$.02 & .729$\pm$.00 & .676$\pm$.02 & \it .869$\pm$.03 & .827$\pm$.01 & .464$\pm$.16\\
3 & \textbf{.929$\pm$.01} & .821$\pm$.01 & .820$\pm$.00 & .730$\pm$.01 & .861$\pm$.04 & \it .885$\pm$.02 & .540$\pm$.15\\
4 & \textbf{.940$\pm$.01} & .854$\pm$.01 & .876$\pm$.00 & .794$\pm$.07 & .880$\pm$.04 & \it .910$\pm$.01 & .557$\pm$.14\\
5 & \textbf{.915$\pm$.01} & .807$\pm$.02 & .697$\pm$.00 & .484$\pm$.11 & .827$\pm$.06 & \it .876$\pm$.01 & .616$\pm$.12\\
6 & \textbf{.958$\pm$.01} & .877$\pm$.02 & .863$\pm$.00 & .882$\pm$.08 & .925$\pm$.03 & \it .927$\pm$.01 & .632$\pm$.13\\
7 & \textbf{.956$\pm$.02} & .907$\pm$.00 & .895$\pm$.00 & .879$\pm$.06 & .900$\pm$.03 & \it .935$\pm$.01 & .606$\pm$.20\\
8 & \it .843$\pm$.02 & .750$\pm$.02 & .788$\pm$.00 & .663$\pm$.07 & \textbf{.866$\pm$.03} & .824$\pm$.02 & .598$\pm$.18\\
9 & \textbf{.942$\pm$.02} & .897$\pm$.01 & .884$\pm$.00 & .829$\pm$.07 & .922$\pm$.01 & \it .941$\pm$.01 & .573$\pm$.19\\
\hline
\multicolumn{8}{c}{~}\\
\multicolumn{8}{c}{Fashion MNIST}\\
\hline 
Class & $\ourmethod$ & AE & OC-SVM & $A^3$ & Deep-SAD & Neg-AE & DRA\\
\hline
0 & \textbf{.911$\pm$.07} & .781$\pm$.01 & \it .859$\pm$.00 & .692$\pm$.05 & .831$\pm$.06 & .838$\pm$.01 & .490$\pm$.11\\
1 & \textbf{.980$\pm$.01} & .958$\pm$.00 & .957$\pm$.00 & .949$\pm$.01 & .957$\pm$.03 & \it .962$\pm$.00 & .425$\pm$.17\\
2 & \textbf{.877$\pm$.04} & .729$\pm$.02 & \it .843$\pm$.00 & .576$\pm$.10 & .801$\pm$.07 & .813$\pm$.01 & .397$\pm$.15\\
3 & \textbf{.920$\pm$.03} & .807$\pm$.01 & .879$\pm$.00 & .857$\pm$.02 & \it .904$\pm$.02 & .893$\pm$.01 & .425$\pm$.13\\
4 & \it .847$\pm$.04 & .770$\pm$.04 & \textbf{.853$\pm$.00} & .762$\pm$.06 & .843$\pm$.05 & .845$\pm$.01 & .524$\pm$.13\\
5 & \textbf{.908$\pm$.02} & .633$\pm$.02 & .807$\pm$.00 & .841$\pm$.04 & \it .864$\pm$.01 & .755$\pm$.03 & .375$\pm$.19\\
6 & \textbf{.796$\pm$.04} & .701$\pm$.04 & \it .778$\pm$.00 & .452$\pm$.08 & .760$\pm$.03 & .761$\pm$.01 & .456$\pm$.11\\
7 & \textbf{.982$\pm$.00} & .927$\pm$.01 & \it .975$\pm$.00 & .970$\pm$.00 & .931$\pm$.06 & .957$\pm$.01 & .408$\pm$.14\\
8 & \textbf{.940$\pm$.02} & .541$\pm$.01 & .755$\pm$.00 & .607$\pm$.05 & \it .878$\pm$.05 & .665$\pm$.02 & .442$\pm$.08\\
9 & \textbf{.971$\pm$.01} & .807$\pm$.02 & .952$\pm$.00 & \it .966$\pm$.02 & .937$\pm$.07 & .890$\pm$.02 & .415$\pm$.17\\
 \hline
    \end{tabular}
    \caption{Comparison with competitors in the \textit{one-vs-all} polluted setting.}
    \label{tab:pollution_onevsall}
\end{table}

\begin{table*}[!t]
\setlength{\tabcolsep}{1.2pt}
    \centering
    \begin{tabular}{|cc|ccccccc|ccccccc|}
\hline
\multicolumn{2}{|c|}{Dataset} & \multicolumn{7}{c|}{MNIST} & \multicolumn{7}{c|}{Fashion MNIST} \\
\hline N & A & $\ourmethod$ & AE & OC-SVM & $A^3$ & Deep-SAD & Neg-AE & DRA & $\ourmethod$ & AE & OC-SVM & $A^3$ & Deep-SAD & Neg-AE & DRA\\
\hline
\multirow{3}{*}{0} & [1,2,3] & \textbf{.983$\pm$.01} & .896$\pm$.00 & .962$\pm$.00 & .770$\pm$.08 & .962$\pm$.01 & .939$\pm$.01 & .531$\pm$.28 & \textbf{.934$\pm$.00} & .849$\pm$.00 & .879$\pm$.00 & .702$\pm$.04 & .891$\pm$.02 & .892$\pm$.01 & .445$\pm$.14\\
                   & [4,5,6] & \textbf{.991$\pm$.00} & .968$\pm$.00 & .965$\pm$.00 & .840$\pm$.08 & .965$\pm$.01 & .962$\pm$.01 & .478$\pm$.19 & \textbf{.920$\pm$.00} & .870$\pm$.00 & .878$\pm$.00 & .581$\pm$.05 & .900$\pm$.00 & .871$\pm$.02 & .409$\pm$.18\\
                   & [7,8,9] & \textbf{.991$\pm$.00} & .948$\pm$.00 & .962$\pm$.00 & .775$\pm$.06 & .969$\pm$.00 & .956$\pm$.01 & .418$\pm$.23 & \textbf{.916$\pm$.00} & .867$\pm$.00 & .876$\pm$.00 & .693$\pm$.01 & .898$\pm$.01 & .866$\pm$.01 & .482$\pm$.15\\
\hline                                                                                                                                 
\multirow{3}{*}{1} & [0,2,3] & \textbf{.998$\pm$.00} & .998$\pm$.00 & .991$\pm$.00 & .986$\pm$.01 & .993$\pm$.00 & .998$\pm$.00 & .373$\pm$.11 & \textbf{.988$\pm$.00} & .975$\pm$.00 & .970$\pm$.00 & .907$\pm$.04 & .978$\pm$.00 & .980$\pm$.00 & .462$\pm$.31\\
                   & [4,5,6] & \textbf{.998$\pm$.00} & .997$\pm$.00 & .990$\pm$.00 & .992$\pm$.00 & .993$\pm$.00 & .998$\pm$.00 & .651$\pm$.31 & \textbf{.984$\pm$.00} & .974$\pm$.00 & .969$\pm$.00 & .934$\pm$.01 & .977$\pm$.00 & .974$\pm$.00 & .341$\pm$.13\\
                   & [7,8,9] & \textbf{.998$\pm$.00} & .996$\pm$.00 & .990$\pm$.00 & .985$\pm$.02 & .993$\pm$.00 & .997$\pm$.00 & .577$\pm$.25 & \textbf{.988$\pm$.00} & .978$\pm$.00 & .968$\pm$.00 & .951$\pm$.01 & .980$\pm$.00 & .971$\pm$.00 & .463$\pm$.18\\
\hline                                                                                                                                                                                                                                                       
\multirow{3}{*}{2} & [0,1,3] & \textbf{.940$\pm$.01} & .841$\pm$.00 & .785$\pm$.00 & .623$\pm$.09 & .873$\pm$.01 & .885$\pm$.02 & .667$\pm$.09 & \textbf{.910$\pm$.00} & .861$\pm$.00 & .851$\pm$.00 & .751$\pm$.02 & .873$\pm$.01 & .841$\pm$.01 & .375$\pm$.20\\
                   & [4,5,6] & \textbf{.938$\pm$.00} & .879$\pm$.01 & .774$\pm$.00 & .624$\pm$.09 & .872$\pm$.03 & .896$\pm$.01 & .455$\pm$.13 & \textbf{.871$\pm$.01} & .819$\pm$.00 & .860$\pm$.00 & .562$\pm$.04 & .844$\pm$.02 & .866$\pm$.01 & .422$\pm$.09\\
                   & [7,8,9] & \textbf{.957$\pm$.01} & .844$\pm$.01 & .768$\pm$.00 & .753$\pm$.10 & .886$\pm$.02 & .885$\pm$.01 & .665$\pm$.10 & \textbf{.887$\pm$.00} & .844$\pm$.00 & .859$\pm$.00 & .603$\pm$.02 & .823$\pm$.04 & .858$\pm$.01 & .606$\pm$.13\\
\hline                                                                                                                                 
\multirow{3}{*}{3} & [0,1,2] & \textbf{.947$\pm$.00} & .851$\pm$.00 & .857$\pm$.00 & .591$\pm$.07 & .869$\pm$.05 & .923$\pm$.01 & .581$\pm$.09 & \textbf{.950$\pm$.00} & .855$\pm$.00 & .895$\pm$.00 & .861$\pm$.01 & .883$\pm$.07 & .929$\pm$.01 & .460$\pm$.20 \\
                   & [4,5,6] & \textbf{.952$\pm$.00} & .905$\pm$.01 & .854$\pm$.00 & .736$\pm$.05 & .889$\pm$.02 & .919$\pm$.01 & .443$\pm$.18 & \textbf{.934$\pm$.00} & .876$\pm$.00 & .895$\pm$.00 & .855$\pm$.01 & .909$\pm$.01 & .916$\pm$.00 & .446$\pm$.16 \\
                   & [7,8,9] & \textbf{.952$\pm$.01} & .891$\pm$.01 & .853$\pm$.00 & .735$\pm$.04 & .876$\pm$.04 & .923$\pm$.01 & .539$\pm$.10 & \textbf{.941$\pm$.01} & .879$\pm$.00 & .895$\pm$.00 & .858$\pm$.01 & .926$\pm$.02 & .906$\pm$.01 & .366$\pm$.11 \\
\hline                                                                                                                                                                                                                                                          
\multirow{3}{*}{4} & [0,1,2] & \textbf{.953$\pm$.01} & .875$\pm$.00 & .899$\pm$.00 & .747$\pm$.05 & .907$\pm$.01 & .924$\pm$.01 & .593$\pm$.11 & \textbf{.916$\pm$.00} & .837$\pm$.00 & .868$\pm$.00 & .714$\pm$.05 & .882$\pm$.01 & .884$\pm$.01 & .380$\pm$.16 \\
                   & [3,5,6] & \textbf{.966$\pm$.00} & .926$\pm$.00 & .901$\pm$.00 & .798$\pm$.07 & .910$\pm$.02 & .949$\pm$.00 & .679$\pm$.10 & \textbf{.895$\pm$.01} & .865$\pm$.00 & .868$\pm$.00 & .660$\pm$.05 & .870$\pm$.01 & .890$\pm$.00 & .530$\pm$.13 \\
                   & [7,8,9] & \textbf{.961$\pm$.01} & .906$\pm$.00 & .901$\pm$.00 & .737$\pm$.06 & .921$\pm$.03 & .939$\pm$.01 & .700$\pm$.16 & \textbf{.903$\pm$.00} & .864$\pm$.00 & .870$\pm$.00 & .753$\pm$.02 & .894$\pm$.03 & .885$\pm$.01 & .299$\pm$.09 \\
\hline                                                                                                                                                                                                                                                          
\multirow{3}{*}{5} & [0,1,2] & \textbf{.958$\pm$.01} & .856$\pm$.00 & .731$\pm$.00 & .482$\pm$.09 & .873$\pm$.01 & .912$\pm$.02 & .552$\pm$.20 & \textbf{.903$\pm$.00} & .715$\pm$.01 & .844$\pm$.00 & .836$\pm$.01 & .873$\pm$.01 & .811$\pm$.01 & .394$\pm$.11 \\
                   & [3,4,6] & \textbf{.964$\pm$.00} & .916$\pm$.01 & .725$\pm$.00 & .603$\pm$.04 & .882$\pm$.02 & .940$\pm$.01 & .589$\pm$.13 & \textbf{.903$\pm$.00} & .746$\pm$.00 & .843$\pm$.00 & .819$\pm$.03 & .879$\pm$.01 & .822$\pm$.03 & .413$\pm$.26 \\
                   & [7,8,9] & \textbf{.960$\pm$.00} & .889$\pm$.00 & .722$\pm$.00 & .504$\pm$.05 & .866$\pm$.01 & .916$\pm$.02 & .622$\pm$.16 & \textbf{.925$\pm$.01} & .858$\pm$.00 & .860$\pm$.00 & .846$\pm$.05 & .899$\pm$.01 & .872$\pm$.00 & .403$\pm$.18 \\
\hline                                                                                                                                                                                                                                                          
\multirow{3}{*}{6} & [0,1,2] & \textbf{.978$\pm$.00} & .899$\pm$.00 & .907$\pm$.00 & .777$\pm$.10 & .960$\pm$.01 & .967$\pm$.01 & .465$\pm$.24 & \textbf{.831$\pm$.01} & .771$\pm$.00 & .786$\pm$.00 & .516$\pm$.03 & .781$\pm$.04 & .792$\pm$.01 & .596$\pm$.21 \\
                   & [3,4,5] & \textbf{.987$\pm$.00} & .962$\pm$.00 & .903$\pm$.00 & .822$\pm$.09 & .965$\pm$.01 & .966$\pm$.01 & .601$\pm$.21 & \textbf{.818$\pm$.01} & .783$\pm$.00 & .786$\pm$.00 & .530$\pm$.04 & .781$\pm$.01 & .803$\pm$.01 & .408$\pm$.10 \\
                   & [7,8,9] & \textbf{.988$\pm$.00} & .943$\pm$.00 & .898$\pm$.00 & .897$\pm$.05 & .963$\pm$.00 & .963$\pm$.00 & .653$\pm$.07 & \textbf{.834$\pm$.00} & .763$\pm$.00 & .786$\pm$.00 & .560$\pm$.03 & .809$\pm$.01 & .793$\pm$.01 & .502$\pm$.14 \\
\hline                                                                                                                                                                                                                                                          
\multirow{3}{*}{7} & [0,1,2] & \textbf{.972$\pm$.00} & .916$\pm$.00 & .916$\pm$.00 & .841$\pm$.05 & .936$\pm$.01 & .949$\pm$.00 & .454$\pm$.13 & \textbf{.989$\pm$.00} & .934$\pm$.01 & .980$\pm$.00 & .960$\pm$.01 & .977$\pm$.00 & .969$\pm$.00 & .432$\pm$.22 \\
                   & [3,4,5] & \textbf{.969$\pm$.00} & .956$\pm$.00 & .915$\pm$.00 & .924$\pm$.02 & .929$\pm$.01 & .959$\pm$.00 & .551$\pm$.07 & \textbf{.986$\pm$.00} & .966$\pm$.00 & .981$\pm$.00 & .924$\pm$.03 & .978$\pm$.00 & .979$\pm$.00 & .643$\pm$.23 \\
                   & [6,8,9] & \textbf{.971$\pm$.00} & .949$\pm$.00 & .915$\pm$.00 & .857$\pm$.05 & .921$\pm$.02 & .961$\pm$.00 & .671$\pm$.10 & \textbf{.986$\pm$.00} & .970$\pm$.00 & .981$\pm$.00 & .951$\pm$.02 & .980$\pm$.00 & .983$\pm$.00 & .340$\pm$.22 \\
\hline                                                                                                                                                                                                                                                          
\multirow{3}{*}{8} & [0,1,2] & \textbf{.915$\pm$.01} & .804$\pm$.01 & .819$\pm$.00 & .652$\pm$.04 & .882$\pm$.02 & .877$\pm$.00 & .453$\pm$.10 & \textbf{.918$\pm$.00} & .701$\pm$.01 & .790$\pm$.00 & .490$\pm$.05 & .891$\pm$.03 & .774$\pm$.03 & .421$\pm$.15 \\
                   & [3,4,5] & \textbf{.892$\pm$.01} & .802$\pm$.01 & .819$\pm$.00 & .567$\pm$.08 & .887$\pm$.02 & .885$\pm$.01 & .340$\pm$.15 & \textbf{.908$\pm$.02} & .753$\pm$.01 & .795$\pm$.00 & .552$\pm$.06 & .895$\pm$.02 & .803$\pm$.02 & .441$\pm$.09 \\
                   & [6,7,9] & .883$\pm$.01 & .782$\pm$.01 & .813$\pm$.00 & .612$\pm$.94 & \textbf{.886$\pm$.01} & .861$\pm$.01 & .608$\pm$.07 & \textbf{.904$\pm$.02} & .737$\pm$.00 & .803$\pm$.00 & .596$\pm$.02 & .895$\pm$.02 & .815$\pm$.01 & .498$\pm$.11 \\
\hline                                                                                                                                                                                                                                                          
\multirow{3}{*}{9} & [0,1,2] & \textbf{.964$\pm$.00} & .899$\pm$.00 & .895$\pm$.00 & .829$\pm$.03 & .937$\pm$.01 & .953$\pm$.00 & .536$\pm$.13 & \textbf{.978$\pm$.00} & .886$\pm$.01 & .971$\pm$.00 & .873$\pm$.05 & .972$\pm$.01 & .945$\pm$.01 & .604$\pm$.08 \\
                   & [3,4,5] & \textbf{.966$\pm$.00} & .959$\pm$.00 & .898$\pm$.00 & .837$\pm$.03 & .953$\pm$.01 & .961$\pm$.00 & .423$\pm$.17 & \textbf{.977$\pm$.00} & .934$\pm$.00 & .971$\pm$.00 & .896$\pm$.03 & .967$\pm$.01 & .952$\pm$.00 & .563$\pm$.18 \\
                   & [6,7,8] & \textbf{.966$\pm$.00} & .948$\pm$.00 & .898$\pm$.00 & .810$\pm$.07 & .948$\pm$.01 & .959$\pm$.00 & .597$\pm$.10 & \textbf{.984$\pm$.00} & .934$\pm$.00 & .971$\pm$.00 & .918$\pm$.03 & .976$\pm$.01 & .959$\pm$.00 & .413$\pm$.21 \\
 \hline
\end{tabular}
    \caption{Comparison with competitors in the \textit{one-vs-many} polluted setting.}
    \label{tab:pollution_one-vs-group}
\end{table*}

In real scenarios it may happen that the training set is polluted by anomalies that are mislabeled and thus appear to the model as normal items. 
{E.g. consider the cases in which the normal data is collected by exploiting semi-automatic or not completely reliable procedures and/or limited resources are available for cleaning the possibly huge normal data collection, and an human analyst points out to the system a small number of selected anomalies in order to improve its performances.}
In this section we investigate the above scenario.
We pollute the datasets by injecting a certain percentage of mislabeled known anomalous examples (these examples are wrongly labeled as 0).
The \textit{percentage level of pollution} quantifies the number of mislabeled known anomalies as a percentage of the number of true inliers in the dataset. That is to say, a $5\%$ level of pollution means that in the training set there are $0.05n$ mislabeled known anomalies (i.e., labeled $0$) and $n$ inliers (also labeled $0$).

Because of the shape of the loss \eqref{eq:inverted-loss} the Autoencoder is trained to well reconstruct the mislabeled anomalies, it is important to select a value of the hyper-parameter $\lambda$ that forces the model to give more relevance to the (bad) reconstruction of the anomalies labeled $1$.
In Figure \ref{fig:pollution_lambda} we show the results obtained for different values of $\lambda$. 
The number $s$ of correctly labeled known anomalies is set to $s=100$.
The red lines refer to the $2\%$ percentage level of pollution, while the blue lines to the $4\%$ of pollution.
We can observe that increasing values of $\lambda$ tend to mitigate the effect of the pollution and to obtain better results. Despite this, the choice of a good value for $\lambda$ is not critical, since all the values of $\lambda$ considered guarantee good performances, though the largest ones are generally associated with the best scores.
In the sequel of this subsection we set $\lambda$ to about $n/s$ which corresponds approximately to $\lambda_4$.

In Table \ref{tab:pollutioncompex} we compare our method with other anomaly detection algorithms for various percentages of pollution. 
{The one-vs-three setting is considered here to increase the diversity of the pollution.}
The number of correctly labeled known anomalies is $s=150$ ($50$ for each of the three known anomalous classes).
$\ourmethod$ improves on the baseline AE and performs better also in this setting.

In order to make the results more robust, we next build up two systematic experiments. 
In each one, we add to the inliers in the training set $100$ randomly selected elements for each anomalous class. Moreover, we assume that only $2$ examples per anomalous class are correctly labeled $1$.

The first experiment concerns the \textit{one-vs-all scenario} in which we consider all the examples of a selected class as inliers and all the other classes as known anomalous ones.
These datasets are extremely polluted, being their pollution percentage about $15\%$, and anomalies have large diversity.
The results, reported in Table \ref{tab:pollution_onevsall}, show that for each class our method achieves good performances being almost always the best method.
Even in those few cases in which it is not the best method, it is always the second ranked with an AUC value very close to the one of the first. 

The second experiment considers the \textit{one-vs-many} scenario in which we have three known anomalous classes. Specifically, to make experiments easily reproducible we fix in turn the normal class and partition the remaining nine classes into three groups consisting of three consecutive classes. The pollution percentage in this case is about the $5\%$.
Table \ref{tab:pollution_one-vs-group} reports the result of this experiment, which confirm the generalization ability of $\ourmethod$ in the polluted scenario.
In Figures \ref{fig:scatter1}, \ref{fig:scatter2}, \ref{fig:scatter3}, red $\times$-marks compare the AUCs on polluted data of experiments in Tables \ref{tab:pollution_onevsall} and \ref{tab:pollution_one-vs-group}.

\subsection{Behavior in the $A^3$ experimental setting}\label{sect:exp_a3}

In this section, in order to compare $\ourmethod$ and $A^3$ on a setting where competitor performance are shown in \cite{a3} to be strong across various configurations, we consider precisely the one proposed by the $A^3$ authors where they show how their method well surpasses the unsupervised baseline methods, and on most experiments also the semi-supervised baseline methods.
In these experiments both normal examples and anomalies belong to a set of classes, thus can be characterized as a \textit{many-vs-many} scenario, and consider the there employed image data, namely $\text{MNIST}$ and $\text{E-MNIST}$.
The composition of the datasets, as specified in \cite{a3}, are reported in Table \ref{tab:a3} and the results have been obtained by using the same experimental setting.
In this experiments, authors \cite{a3} compare their method, namely $A^3$, with standard AE, Isolation Forest \cite{LiuTZ12}, DAGMM \cite{DAGMM} and DevNet \cite{Pang2021}.
Since $A^3$ overcomes other competitors, we report only $A^3$ results and refer the interested readers to Table 3 of \cite{a3} for the performances of other methods.

It can be noted that performances of $A^3$ and $\ourmethod$ are comparable when the set of anomalies in the test set belong to the same classes of the known anomalies, while $\ourmethod$ outperforms $A^3$ in all datasets and experiments when the test set contains anomalies in classes not represented in the training set.
This confirms the ability of $\ourmethod$ in detecting anomalies belonging to completely unknown classes, an important requirement of anomaly detection systems.

\begin{table}
\setlength{\tabcolsep}{5pt}
\centering
\begin{tabular}{|c|c|c|c|c|c|} 
\hline
\multirow{2}{*}{{Data }} & \multirow{2}{*}{{Normal }} & {Train}   & {Test}                                                                                   & $\ourmethod$         & $A^3$                 \\
                                &                                   & {Anomaly} & {Anomaly}                                                                                & {AUC}         & {AUC}          \\ 
\hline
\multirow{4}{*}{MNIST}                           & $0, \dots, 5$                     & 6, 7             & 6, 7                                                                                            & .98$\pm$.00          & \textbf{.99$\pm$.00}  \\ 
\cline{2-6}
                           & $4, \dots, 9$                     & 0, 1             & 0,1                                                                                             & \textbf{1.0$\pm$.00} & \textbf{1.0$\pm$.00}  \\ 
\cline{2-6}
                           & $0, \dots, 5$                     & 6, 7             & 6, 7, 8, 9                                                                                      & \textbf{.92$\pm$.01} & .88$\pm$.03           \\ 
\cline{2-6}
                           & $4, \dots, 9$                     & 0, 1             & 0, 1, 2, 3                                                                                      & \textbf{.96$\pm$.01} & .92$\pm$.02           \\ 
\hline
\multirow{4}{*}{E-MNIST}                         & $0, \dots, 5$                     & $A, \dots, E$    & $A, \dots, E$                                                                                   & \textbf{.99$\pm$.00} & \textbf{.99$\pm$.01}  \\ 
\cline{2-6}
                         & $0, \dots, 5$                     & $A, \dots, E$    & \begin{tabular}[c]{@{}c@{}}$A, \dots, E,$\\$V, \dots, Z$\end{tabular} & \textbf{.98$\pm$.00} & .96$\pm$.01           \\ 
\cline{2-6}
                         & $0, \dots, 5$                     & $V, \dots, Z$    & $V, \dots, Z$                                                                                   & \textbf{.99$\pm$.00} & \textbf{.99$\pm$.00}  \\ 
\cline{2-6}
                         & $0, \dots, 5$                     & $V, \dots, Z$    & \begin{tabular}[c]{@{}c@{}}$A, \dots, E,$\\$V, \dots, Z$\end{tabular} & \textbf{.98$\pm$.00} & .95$\pm$.02           \\
\hline
\end{tabular}
\caption{Comparison with $A^3$ in the
\textit{many-vs-many} setting \cite{a3}.}
\label{tab:a3}
\end{table}

\section{Conclusion}
\label{sec:concl}

In this work, we deal 
with the semi-supervised anomaly detection problem in which a limited number of anomalous examples is available and introduce the $\ourmethod$ algorithm, 
based on a new loss function that exploits the information of the labelled anomalies and allows Autoencoders to learn to reconstruct them poorly. This strategy increases the contrast between the reconstruction error of normal examples and that associated with both known and unknown anomalies, thus enhancing 
anomaly
detection 
performances.
We focus on classical Autoencoders and show that our proposed strategy is able to reach state-of-the-art performances even when built on very simple architectures.
We also prove that $\ourmethod$ is very effective in some relevant anomaly detection scenarios, namely the one in which anomalies in the test set are different from the ones in the training set and the one in which the training set is polluted by some mislabeled examples.
As a future work we intend to inject the here presented strategy in other reconstruction-based architectures that permit to organize their latent space to guarantee \textit{continuity} and \textit{completeness}, such as VAEs and GANs, to possibly obtain even improved detection performances.

\section*{Acknowledgments}
We acknowledge the support  of the PNRR project FAIR - Future AI Research (PE00000013), Spoke 9 - Green-aware AI, under the NRRP MUR program funded by the NextGenerationEU.

\ifCLASSOPTIONcaptionsoff
  \newpage
\fi

\vspace{-1.1cm}

\begin{IEEEbiography}[{\includegraphics[width=1in,height=1.25in,clip,keepaspectratio]{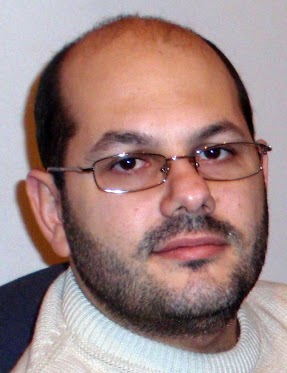}}]{Fabrizio Angiulli} is a full professor of computer science at DIMES, University of Calabria, Italy. His research interests include data mining, machine learning, and artificial intelligence, with a focus on anomaly detection, large and high-dimensional data analysis, and explainable learning. He has authored more than one hundred papers appearing in premier journals and conferences. He regularly serves on the program committee of several conferences and, as an associate editor of AI Communications.
\end{IEEEbiography}
\vspace{-1.1cm}
\begin{IEEEbiography}[{\includegraphics[width=1in,height=1.25in,clip,keepaspectratio]{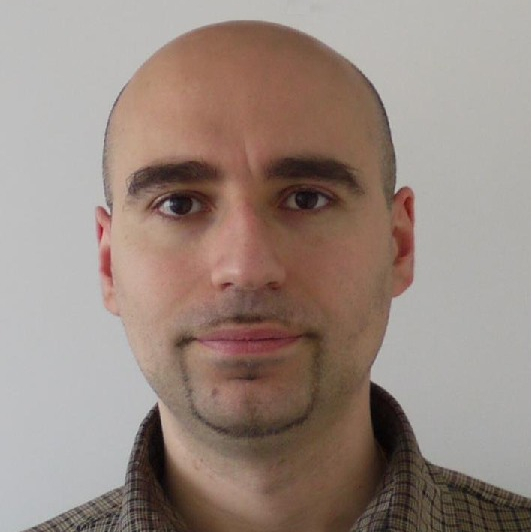}}]{Fabio Fassetti}
received the PhD degree in system engineering and computer science in 2008 from the University of Calabria, Italy. He currently is an associate professor of computer engineering at DIMES Department, University of Calabria, Italy. His research interests include bioinformatics, machine learning, data mining, artificial intelligence, knowledge representation and reasoning.
\end{IEEEbiography}
\vspace{-1.1cm}
\begin{IEEEbiography}[{\includegraphics[width=1in,height=1.25in,clip,keepaspectratio]{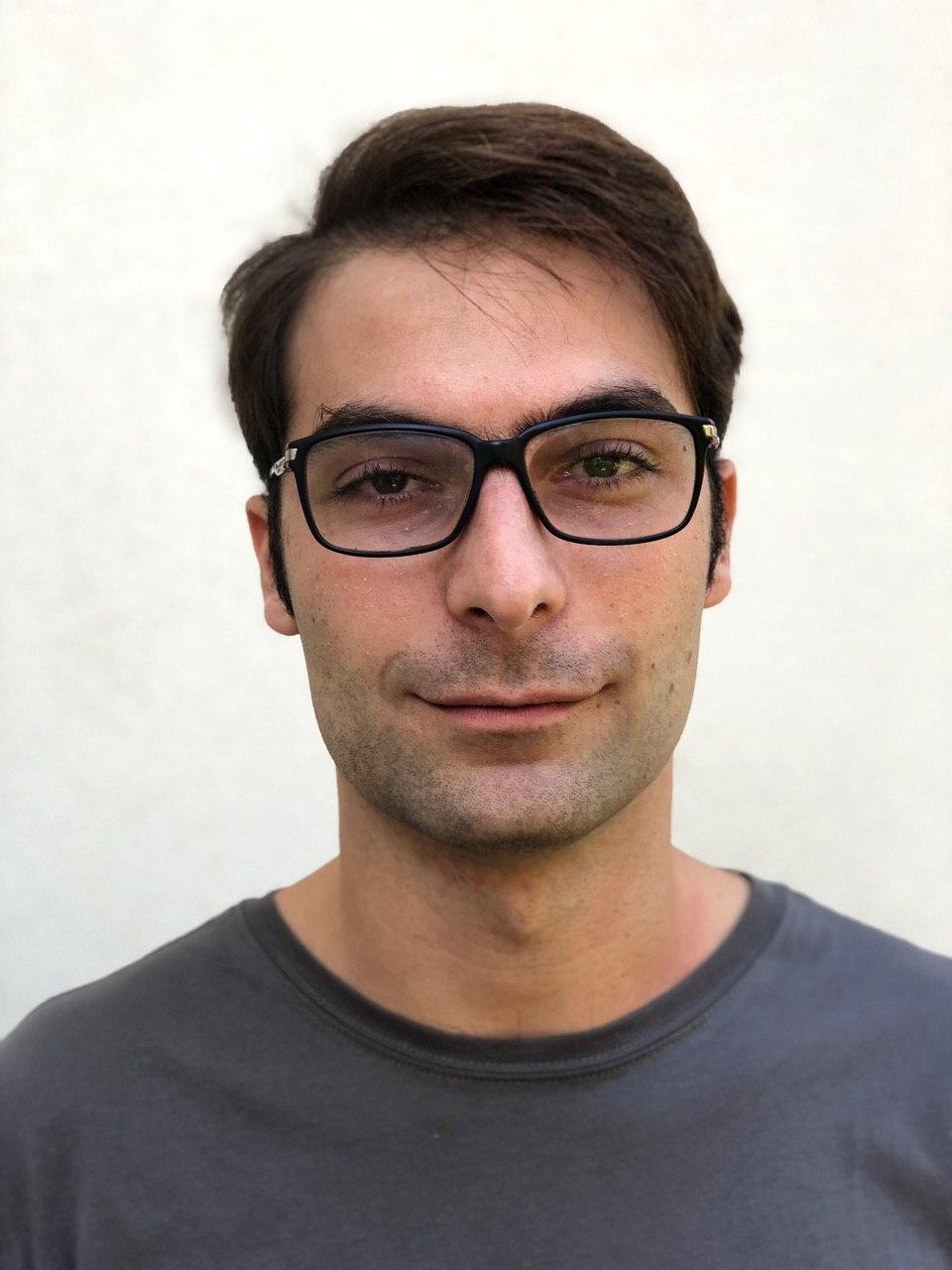}}]{Luca Ferragina}
received the PhD in Information and Communication Technology from the University of Calabria, Italy. He currently is assistant professor at DIMES Department, University of Calabria, Italy. The main topics of his research are about Anomaly Detection and Deep Learning.
\end{IEEEbiography}

\end{document}